%% file: main.tex
\documentclass{article}
\usepackage[a4paper, total={7in, 10in}]{geometry}

\input{newcmnds}

\begin{document}
        \title{CROCS: A Two-Stage Clustering Framework for Behaviour-Centric Consumer Segmentation with Smart Meter Data}
        
	\author{Luke W. Yerbury, G. C. Livingston Jr, Ricardo J.G.B. Campello,\\Mark Goldsworthy, Lachlan O'Neil}
	\date{}

        \maketitle

	\input{Abstract}

	\input{Nomenclature}

        \renewcommand{\nomname}{\normalsize Nomenclature}
        \begin{footnotesize}
        \begin{framed}
            \vspace{-0.5cm}
            \printnomenclature
            \vspace{-0.4cm}
        \end{framed}
        \end{footnotesize}
 
	\input{Introduction}
        \input{RelatedWorks}

        \input{TheFramework}
        
        \input{Methodology}

	\input{Results}

	\input{Application}

	\input{Discussion}

	\input{Conclusion}
		
        \small
	\bibliographystyle{unsrt}
	\bibliography{references}

\end{document}

%% file: newcmnds.tex
\usepackage{amsmath}
\usepackage{amsfonts}
\usepackage{amssymb}
\usepackage{siunitx}
\usepackage{bbding}
\usepackage{bm}
\usepackage[export]{adjustbox}
\usepackage{enumitem}
\usepackage{mathtools}
\usepackage{booktabs}

\usepackage[dvipsnames, table]{xcolor}
\usepackage{subcaption}
\usepackage{floatrow}
\usepackage[framemethod=tikz]{mdframed}

\usepackage{array}
\usepackage{calc} 
\usepackage{makecell} 
\usepackage{colortbl}

\usepackage{rotating}
\usepackage{pdflscape}
\usepackage{afterpage}

\usepackage{pgfplots}
\pgfplotsset{compat=1.18}

\usepackage{nomencl}
\makenomenclature
\usepackage{framed}
\usepackage{multicol}
\renewcommand{\nompreamble}{\begin{multicols}{2}}
\renewcommand{\nompostamble}{\end{multicols}}
\renewcommand\nompreamble{\begin{multicols}{2}}
\renewcommand\nompostamble{\end{multicols}}

\usepackage[numbers, compress]{natbib}
\usepackage[colorlinks = true, pdfstartview = FitV, linkcolor = blue, citecolor = blue, urlcolor = blue, pdfencoding=auto, psdextra]{hyperref}
\usepackage{cleveref}

\usepackage{caption}
\usepackage{soul}
\soulregister\cite7
\soulregister\Cref7
\soulregister\texttt7
\sethlcolor{ForestGreen!20}

\crefname{section}{Section}{Sections}
\crefname{appendix}{Appendix}{Appendices}

\newcommand*\justify{%
  \fontdimen2\font=0.4em
  \fontdimen3\font=0.2em
  \fontdimen4\font=0.1em
  \fontdimen7\font=0.1em
  \hyphenchar\font=`\-
}

\renewcommand{\texttt}[1]{%
  \begingroup
  \ttfamily
  \begingroup\lccode`~=`/\lowercase{\endgroup\def~}{/\discretionary{}{}{}}%
  \begingroup\lccode`~=`[\lowercase{\endgroup\def~}{[\discretionary{}{}{}}%
  \begingroup\lccode`~=`.\lowercase{\endgroup\def~}{.\discretionary{}{}{}}%
  \catcode`/=\active\catcode`[=\active\catcode`.=\active
  \justify\scantokens{#1\noexpand}%
  \endgroup
}

\usepackage{tikz}
\usetikzlibrary{matrix,calc,arrows.meta,shapes,arrows,positioning,shadows,trees,mindmap,backgrounds,fit}

\definecolor{tikz_Blue}{HTML}{1f77b4}
\definecolor{tikz_Red}{HTML}{d62728}
\definecolor{tikz_Green}{HTML}{2ca02c}
\definecolor{tikz_Orange}{HTML}{ff7f0e}

\newdimen\figrasterwd
\figrasterwd\textwidth

\newcommand{\Rom}[1]{\uppercase\expandafter{\romannumeral #1\relax}}

\newcommand {\xx}  { {\bm{x}} }

\definecolor{MyForestGreen}{RGB}{34, 139, 34}
\definecolor{MyRedOrange}{RGB}{255, 69, 0}

\newcommand{\n}{\textemdash}

\definecolor{TableGray}{RGB}{240, 240, 240}
\definecolor{TableWhite}{RGB}{253, 253, 253}

\newcommand*\bigO[1]{\mathcal O\left( #1\right)}

\usepackage{listings}

%% file: Abstract.tex
\begin{abstract}

With grid operators confronting rising uncertainty from renewable integration and a broader push toward electrification, Demand-Side Management (DSM) --- particularly Demand Response (DR) --- has attracted significant attention as a cost-effective mechanism for balancing modern electricity systems. Unprecedented volumes of consumption data from a continuing global deployment of smart meters enable consumer segmentation based on real usage behaviours, promising to inform the design of more effective DSM and DR programs. However, existing clustering-based segmentation methods insufficiently reflect the behavioural diversity of consumers, often relying on rigid temporal alignment, and faltering in the presence of anomalies, missing data, or large-scale deployments. 

To address these challenges, we propose a novel two-stage clustering framework --- Clustered Representations Optimising Consumer Segmentation (CROCS). In the first stage, each consumer’s daily load profiles are clustered independently to form a Representative Load Set (RLS), providing a compact summary of their typical diurnal consumption behaviours. In the second stage, consumers are clustered using the Weighted Sum of Minimum Distances (WSMD), a novel set-to-set measure that compares RLSs by accounting for both the prevalence and similarity of those behaviours. Finally, community detection on the WSMD-induced graph reveals higher-order prototypes that embody the shared diurnal behaviours defining consumer groups, enhancing the interpretability of the resulting clusters.

Extensive experiments on both synthetic and real Australian smart meter datasets demonstrate that CROCS captures intra-consumer variability, uncovers both synchronous and asynchronous behavioural similarities, and remains robust to anomalies and missing data, while scaling efficiently through natural parallelisation. These results highlight CROCS as a robust and extensible framework for advancing behaviour-centric consumer segmentation, strengthening the analytical foundations of DSM and DR strategies, and offering insights at both consumer and system levels that extend to a wide range of energy applications.

\end{abstract}

%% file: Nomenclature.tex
\nomenclature[Vp]{DSM}{Demand-Side Management}
\nomenclature[Vp]{DR}{Demand Response}
\nomenclature[Vp]{DLP}{Daily Load Profile} 
\nomenclature[Vp]{RLP}{Representative Load Profile}
\nomenclature[Vp]{ED}{Euclidean Distance} 
\nomenclature[Vp]{DTW}{Dynamic Time Warping}
\nomenclature[Vp]{CROCS}{Clustered Representations\\Optimising Consumer Segmentation}
\nomenclature[Vp]{RLS}{Representative Load Set}
\nomenclature[Vp]{WSMD}{Weighted Sum of Minimum Distances}
\nomenclature[Vp]{RRLS}{Refined Representative Load Set}
\nomenclature[Vp]{JSD}{Jensen-Shannon Divergence}
\nomenclature[Vp]{TPM}{Transition Probability Matrix}
\nomenclature[Vp]{ARI}{Adjusted Rand Index}
\nomenclature[Vp]{AMI}{Adjusted Mutual Information}
\nomenclature[Vp]{PSI}{Pair Sets Index}
\nomenclature[Vp]{DCP}{Dominant Consumption Profile }
\nomenclature[Vp]{GPF}{Global Pattern Frequency}
\nomenclature[Vp]{HAC-Wa}{Hierarchical Agglomerative Clustering\\with Ward's linkage}
\nomenclature[Vp]{KMd}{$k$-medoids}
\nomenclature[Vp]{AG}{Ausgrid solar home electricity dataset}
\nomenclature[Vp]{SGSC}{Smart-Grid Smart-City dataset}
\nomenclature[Vp]{EVI}{External Validity Index}

%% file: Introduction.tex
\section{Introduction}
\label{Sec:Introduction}

Smart metering infrastructure has proliferated globally over recent decades, with a 2021 survey finding installation programs across 47 countries targeting nearly 1.5 billion households \cite{Sovacool2021GlobalTransitions}. This investment is driven by the technology's capacity to digitise and demystify energy consumption in a transitioning energy market, enabling better informed decision-making across the energy ecosystem. In some countries, these roll-outs are approaching or have already reached maturity --- for instance, the Australian Energy Market Commission requires universal smart meter deployment across the National Electricity Market by 2030 \cite{AustralianEnergyMarketCommission2024AcceleratingDetermination}, meanwhile Italy is proceeding with a second-generation roll-out \cite{Stagnaro2024}. 

The resulting data deluge arrives at a critical juncture, as modern electrical systems are undergoing profound transformations, from firm to variable energy sources, synchronous to inverter-based generation, centralised to decentralised architectures, and passive to active consumers \cite{AustralianEnergyMarketOperator2020}. The increasing penetration of renewable and distributed generation introduces substantial variability and unpredictability on the supply side. Simultaneously, the electrification of transportation \cite{Yuan2021} and heating systems \cite{Thomaen2021} is projected to dramatically increase demand \cite{Bobmann2015,Castillo2022}. Maintaining supply-demand balance within this evolving landscape presents a complex challenge for system operators.

Among the available strategies for managing this complexity, including fast-acting supply and energy storage, Demand-Side Management (DSM) --- particularly Demand Response (DR) --- has attracted significant recent attention as a cost-effective and scalable solution \cite{Antonopoulos2020,Yilmaz2019,Michalakopoulos2024APrograms,Meng2023,Qiu2023PersonalizedMarket,Parrish2020AResponse,Yan2018}. DR enables consumers to actively participate in grid management by reducing or shifting their electricity usage in response to external signals, such as dynamic pricing or direct control incentives \cite{Yan2018,Parrish2020AResponse}. These programs not only promote grid reliability, but also contribute to environmental sustainability by synchronising energy consumption with renewable energy availability, all while providing economic benefits to participating consumers. However, effectively implementing DR programs requires a nuanced understanding of consumer behaviour. 

The process of collecting similar users into distinct groups, known as consumer segmentation, is key to achieving this understanding. Historically, segmentation relied on contractual data, commercial classifications, and electrical parameters rather than real consumption behaviours \cite{Chicco2001,Chicco2003}. However, studies have shown that these and other demographic attributes are poor predictors of actual electricity usage, with households that appear similar on paper often exhibiting vastly different consumption patterns \cite{Yan2018,Rajabi2017}. 

Recent approaches overwhelmingly rely on \textit{clustering} of smart meter time series data to form behaviourally meaningful consumer segments, where similarity is based on dynamic demand patterns rather than static traits \cite{Wang2019,Rajabi2017}. Clustering is a ubiquitous unsupervised learning technique aimed at partitioning objects from datasets into groups, such that objects within groups share a greater notion of similarity or homogeneity than objects in different groups \cite{Aggarwal2014,Gan2007DataApplications}. Beyond DR, consumer clustering supports a range of other applications, including load forecasting \cite{Kim2023Time-seriesData,Auder2018,Li2016Short-TermBehavior,Quilumba2015b}, identification of energy theft \cite{Mishra2025,Qi2022,DeSouza2020}, planning of energy efficiency and net metering schemes \cite{Wen2023,Ahir2022}, and anomaly detection \cite{Fenza2019}. 

Despite numerous methodologies being proposed for consumer clustering in the literature \cite{Michalakopoulos2024APrograms,Meng2023,Qiu2023PersonalizedMarket,Alonso2022ClusteringGrids,Kaur2022BehaviorApproach,Banales2021,AlKhafaf2021AConsumers,Lee2020LoadStudy,Alonso2020HierarchicalAutocovariances,Sun2020,Motlagh2019,Yilmaz2019,Sandels2019,Tureczek2018,Chen2017,Rajabi2017,Wang2016b,Chicco2012a,Tsekouras2007Two-stageCustomers}, current methods are constrained by one or more of the following fundamental limitations: (i) inadequate representation of consumer behavioural diversity; (ii) oversimplified characterisations of clusters; (iii) reliance on temporally anchored comparisons; (iv) sensitivity to behavioural anomalies; (v) requiring time-synchronised input data; (vi) unsatisfactory handling of missing data; (vii) sensitivity to regular discontinuities; and (viii) limited scalability to real-world populations. These shortcomings diminish the effectiveness of consumer segmentation and consequently reduce the potential returns from substantial investments in smart metering infrastructure. 

No previous approach simultaneously addresses all of these limitations, representing a significant gap in the literature. Bridging this gap requires rethinking how consumption behaviours are represented, compared, and aggregated within the clustering process. In this paper, we operationalise this perspective through the proposed Clustered Representations Optimising Consumer Segmentation (CROCS) framework, a novel two-stage methodology that more faithfully represents and compares consumer load behaviour, enabling interpretable, scalable and robust clustering for real-world datasets. Specifically, this paper makes the following contributions:

\begin{itemize}
    \item We comprehensively review existing consumer clustering methodologies and their limitations, including a structured taxonomy of two-stage clustering methods that contextualises the gap addressed by CROCS.
    \item We propose CROCS, a novel two-stage consumer clustering framework that, to the best of our knowledge, is the first to simultaneously address all eight limitations identified in existing approaches.
    \item We introduce the novel Representative Load Set (RLS), a flexible set-based consumer representation learned in the first stage of CROCS by independently clustering each consumer's Daily Load Profiles (DLP), thereby capturing the diversity of their recurring consumption behaviours.
    \item We introduce the Weighted Sum of Minimum Distances (WSMD), a novel set-to-set dissimilarity measure that drives the second stage of consumer clustering in CROCS, comparing consumer RLSs by accounting for both the prevalence and similarity of their constituent behavioural patterns.
    \item We develop a graph-based community detection procedure for extracting Refined Representative Load Sets (RRLS), uniquely enhancing the interpretability of consumer clusters by providing multi-profile summaries of their shared diurnal behaviours.
    \item We validate the effectiveness of CROCS through comprehensive experimental analyses using both synthetic and real-world smart meter datasets, interrogating key design choices, comparing performance against existing methodologies, and demonstrating practical utility through an application to real Australian data.
\end{itemize}

The remainder of the paper is organised as follows. \Cref{Sec:LiteratureReview} reviews existing consumer clustering methodologies and their limitations in detail, including the specific taxonomy of two-stage methods. \Cref{Sec:TheFramework} describes the proposed framework, including mathematical formulations and implementation details of each stage. \Cref{Sec:Methodology,Sec:Results} present the experimental methodology and results respectively.  \Cref{Sec:Application}  demonstrates the practical utility of CROCS through an application to real Australian smart meter data. \Cref{Sec:Discussion} discusses implications of the experimental results, provides practical guidance for implementing CROCS, and explores future research directions. Finally, \Cref{Sec:Conclusion} concludes the paper.

%% file: RelatedWorks.tex
\section{Literature Review}
\label{Sec:LiteratureReview}

This section reviews existing consumer clustering methodologies that utilise smart meter time series data, outlining the key limitations this work seeks to address. We begin by introducing some nomenclature for decomposing time series clustering approaches into their core components, providing a basis for contrasting existing methods. We then survey the consumer clustering literature, introducing a structured taxonomy of multi-stage approaches to situate CROCS within the broader methodological landscape. Finally, we discuss the eight fundamental limitations that collectively motivate the proposed framework.

\subsection{The Components of a Time Series Clustering Approach} \label{Sec:LiteratureReview-ClusteringComponents}

Popular traditional clustering algorithms such as $k$-means or Hierarchical Agglomerative Clustering (HAC) were proposed to operate on objects described by orderless vectors of attributes assumed to exist in standard Euclidean space \cite{Aggarwal2014}. However in the age of smart meters, consumers produce time series data, where the sequential ordering of observations contains critical temporal information that should also be accounted for during the clustering process. It has been suggested that most time series clustering approaches are composed of five fundamental \textit{components}: a data normalisation procedure, a data representation method, a distance measure, a clustering algorithm, and a prototype definition \cite{Yerbury2024b,Yerbury2024,Aghabozorgi2015}. Studies proposing consumer clustering approaches overwhelmingly suggest alternative data representations, as this component in particular facilitates compatibility between the complex temporal consumption data and the wealth of traditional clustering algorithms that are well-understood and readily deployable.

\textit{Representation method} is a catch-all term for any feature extraction method or transformation of the \textit{raw} time series which is intended to improve clustering outcomes. An effective representation should simultaneously emphasise characteristics most relevant for segmentation and reduce dimensionality, thereby moderating the influence of noise and increasing the computational efficiency of the clustering approach \cite{Aghabozorgi2015, Wang2013b}. Without appropriate representations, energy consumer clustering with long smart meter time series would be subject to the ``curse of dimensionality'' that renders comparisons between high-dimensional objects virtually meaningless \cite{Aggarwal2001}.

\textit{Distance measures} quantify the similarity between pairs of data objects, and are required for the vast majority of clustering algorithms. Typically representation methods specify a subset of compatible distances. For example, while the Euclidean Distance (ED) \cite{Aggarwal2014} can be applied across a swathe of different data types and representations, it is not compatible with the string-based Symbolic Aggregate Approximation \cite{Lin2007ExperiencingSeries}. 

A \textit{clustering algorithm} is a procedure which utilises a data representation and distance measure to produce a partition between a set of objects, where objects placed in the same group are more similar than objects separated into different groups. The obtained partitions are typically hard or crisp, indicating that each object belongs to a single cluster, though partial cluster membership can be obtained by calling upon fuzzy or probabilistic clustering algorithms \cite{Aggarwal2014}. 

Depending on the representation, a \textit{normalisation procedure} can be applied before and/or after the representation has been prepared from the data. Normalisation of data prior to clustering has long been deemed necessary for generating meaningful clusters \cite{Keogh2002, Rakthanmanon2012}. For time series data, normalisation is often applied independently to the individual series, as this serves to direct attention toward the shapes of the time series, rather than their amplitudes when clustering. 

Finally, various \textit{prototype definitions} have been proposed to represent or summarise the constituents of a group of objects. They are commonly utilised post-clustering for visualisations, summary purposes or as exemplars for downstream applications. They are not explicitly required for most clustering approaches, but are involved within subroutines of some clustering algorithms (e.g. $k$-means, $k$-medians and $k$-medoids \cite{Aggarwal2014}). 

It should be noted that some clustering approaches do not atomise completely into all five of these components. For instance, clustering time series with Hidden Markov Models and Expectation-Maximisation \cite{Aggarwal2014} does not make explicit use of any specific distance measure. Furthermore, any normalisation or prototype definition could be incorporated into this clustering approach.

\subsection{Related Works} \label{Sec:LiteratureReview-RelatedWorks}

\Cref{Tab:OtherCustomerClusteringStudies} presents a collection of consumer clustering approaches that have been proposed in the literature, detailing their suggested representation methods, distance measures and clustering algorithms. In the following discussion, we will focus on the representations that have been proposed, as this is where the most novelty and variation between studies appears. As can be seen in \Cref{Tab:OtherCustomerClusteringStudies}, most studies ultimately employ standard distance measures and clustering algorithms --- predominantly $k$-means with ED.

\input{Table---OtherConsumerClusteringStudies}

One of the most popular representations is the Representative Load Profile (RLP), which is obtained by taking the point-wise average of all DLPs in the period of interest \cite{Michalakopoulos2024APrograms,Qiu2023PersonalizedMarket,Yilmaz2019,Chicco2012a,Flath2012}. This representation is commonly computed for DLPs produced under the same loading conditions --- weekday DLPs for instance, or DLPs from the same season. Consumers represented by RLPs are then typically clustered using lock-step \cite{Wang2013b} distance measures such as ED, and clustered with a wide variety of algorithms. Similarly, the median and 95\textsuperscript{th} percentile load profiles were used in \cite{Alonso2022ClusteringGrids}, instead of the average load profile. 

Some studies represent consumers with features extracted from the RLPs themselves, often related to peak timing or other simple statistical descriptors \cite{Michalakopoulos2024APrograms}. Yilmaz et al. \cite{Yilmaz2019} and Kaur et al. \cite{Kaur2022BehaviorApproach} take the averaging process even further, suggesting to represent consumers with feature vectors composed of average consumption across 4 non-overlapping diurnal time periods, along with measures of daily variation or variation across different loading conditions. Sandels et al. \cite{Sandels2019} suggest a similar approach, though with 32 strata derived from segregating consumption according to 4 non-overlapping diurnal time periods, 4 temperature ranges and 2 day-types (weekdays and weekends). Relatedly, Meng et al. \cite{Meng2023} use a longer vector composed of concatenated RLPs, one for each month of data.

Despite some exceptions \cite{Ahir2022}, it has generally been acknowledged that clustering the raw or normalised long time series is ineffective for consumer segmentation. Aside from averaging and average-based feature extraction, long time series specific features such as autocorrelations, partial autocorrelations or wavelet coefficients have also been used to represent overall temporal dependencies exhibited by consumers \cite{Alonso2022ClusteringGrids,Alonso2020HierarchicalAutocovariances,Tureczek2018}. Chen et al. \cite{Chen2017} suggested using Principal Component Analysis (PCA) components that explain at least 90\% of the data variance, though this was only tested for segmenting consumers for single days of data. Another popular category of representation utilises parameters from appropriate time series models. Motlagh et al. \cite{Motlagh2019} propose a Delay Coordinate Embedding (DCE) representation, which treats each consumer's series as a dynamic system. The system's states are represented in the phase space and modelled via a mapping strategy, whose parameters are then used to represent consumers and assess their similarity. Wang et al. \cite{Wang2016b} propose modelling consumption across four regular diurnal periods using Markov sequences. They employ the Symbolic Aggregate Approximation to discretise consumption into states, then fit transition probability matrices for each of the four periods, using Kullback-Leibler divergence to assess consumer similarities.

Our CROCS framework is not the first to employ multiple stages of clustering for consumer segmentation --- or indeed, for time series clustering more generally. Sun et al. \cite{Sun2020} and Tsekouras et al. \cite{Tsekouras2007Two-stageCustomers} both proposed two-stage frameworks for clustering energy consumers. Sun et al. \cite{Sun2020} suggested an ensemble clustering approach, where the first stage produced many partitions of consumers by clustering their DLPs independently on each day of the relevant period using $k$-means with ED. These daily partitions were then amalgamated into a similarity matrix indicating how frequently households were grouped together, with a final spectral clustering performed on this matrix to produce the definitive consumer segmentation. On the other hand, the first stage of the method proposed by Tsekouras et al. \cite{Tsekouras2007Two-stageCustomers} clusters the DLPs of each consumer independently, as for our CROCS framework. They select the optimal partition for each consumer obtained from $k$-means, adaptive vector quantisation, fuzzy $c$-means and HAC by optimising a set of six Relative Validity Indices (RVIs) (such as the Davies-Bouldin Index). However, unlike CROCS, for the second stage of clustering they suggested that a single prototypical load profile be selected to represent each consumer. This profile was primarily taken as the prototype of the cluster with highest membership --- for this reason we refer to this form of consumer representation as the Dominant Consumption Profile (DCP). However, the authors do suggest that the cluster with some other characteristic, such as peak load demand, could be used instead.

\Cref{Tab:OtherTwoStageClusteringMethods} collects two-stage methods that have been proposed in the literature, categorising them according to their end-goal --- either consumer segmentation or load profile clustering. This includes methods that have been proposed in the broader time series clustering literature, with their end-goals interpreted in terms of smart meter clustering. Unlike consumer segmentation, load profile clustering is concerned with the segmentation of DLPs, regardless of which consumers produced them on which days. Many more two-stage methods have been proposed for load profile clustering than for consumer segmentation. Two-stage methods for load profile clustering are typically implemented in one of two ways. Refinement methods operate as for the first stage of CROCS, independently clustering the DLPs of each consumer individually. However, the second stage then proceeds to cluster all prototypes from all consumers, seeking to find common diurnal patterns exhibited across the board. Hierarchical methods on the other hand will cluster DLPs within the groups obtained from the first stage of clustering, often using a different method to capture different characteristics at each level.

For consumer clustering, we have categorised the daily ensemble method of Sun et al. \cite{Sun2020} as an ensemble-based two-stage method, and the DCP-based method of Tsekouras et al. \cite{Tsekouras2007Two-stageCustomers} as representation-based. The purpose of the first stage of clustering in a representation-based two-stage method is to learn an appropriate representation to use for subsequently clustering the consumers (i.e. the whole time series) in the second stage of clustering. Our proposed CROCS framework and the DCP approach suggest that this representation should be local, that is, learned independently for each consumer. However, Nakashima et al. \cite{Nakashima2016PerformanceData} and Li et al. \cite{Li2019AStudy} propose two-stage clustering methods for general application which instead utilise a global representation. Speaking in the context of smart meter time series, these methods first perform a load profile clustering, finding common consumption patterns exhibited in DLPs across all consumers on all days. By clustering all DLPs in this ``pooled'' manner, consumers can then be represented according to the frequency with which they display the discovered prototypical patterns. In particular, both of these studies utilise proportion or count vectors to represent consumers, which indicate how frequently each consumer's DLPs were found to belong to each of the global clusters. We refer to this form of consumer representation as the Global Pattern Frequency (GPF) representation.

\input{Table---OtherTwoStageClusteringMethods}

\subsection{Limitations of Existing Studies} \label{Sec:LiteratureReview-Limitations}

Our review of the literature identifies several limitations that variously affect existing single- and multi-stage consumer clustering approaches. We now discuss each of these limitations in turn.

\subsubsection*{(i) Inability to reflect intra-consumer variation} \label{Sec:LiteratureReview-Limitations(i)}
The success of consumer clustering depends critically on how each consumer's long load time series is represented for comparison \cite{Motlagh2019}. While computationally efficient and easy to interpret, it has been widely acknowledged that aggregation-averaging processes, such as is used to create RLPs, miss the substantial intra-consumer variability observed for real-world consumers \cite{Yerbury2024b,Rajabi2020,Motlagh2019,Yilmaz2019,Rajabi2017,Jin,Li2016Multi-ResolutionData,McLoughlin2015}. Individual consumers can exhibit diverse usage behaviours from day to day, such that if clustered, their DLPs would belong to different clusters on different days \cite{Yerbury2024b,Yilmaz2019,Li2016Multi-ResolutionData}. Furthermore, averaging across these distinct behaviours can result in RLPs that are not only unrepresentative of any constituent DLPs, but may also lie in low-density regions of the data space --- particularly if the DLP distribution is non-convex \cite{Li2016Multi-ResolutionData}. 

Alternative representations that process the entire long time-series --- such as those using autocorrelations, Markov model parameters or delay coordinate embedding map parameters --- may capture specific aspects of overall temporal variation (like periodicity, state transitions, or attractor dynamics) and avoid direct averaging, but they still fundamentally compress multi-modal behavioural landscapes into fixed-length feature vectors. This compression obscures the explicit structure of distinct behavioural modes, making their shapes, prevalence, and relationships not directly accessible from the representation. 

The aforementioned two-stage consumer clustering approaches are better equipped to avoid this limitation. For instance, the daily ensemble clustering method from Sun et al. \cite{Sun2020} and the GPF approaches from Nakashima et al. \cite{Nakashima2016PerformanceData} and Li et al. \cite{Li2019AStudy} allow consumers to belong to different clusters on different days. Meanwhile, the DCP method from Tsekouras et al. \cite{Tsekouras2007Two-stageCustomers} still only uses a single profile for consumer segmentation in the second stage. In doing so, the method fails to take full advantage of the rich representation discovered in the first stage of independent consumer DLP clustering. 

Because intra-consumer variability has been recognised as an indicator of suitability for DSM and DR strategies \cite{Yilmaz2019,Jin,Dent2014,Kwac2014}, clustering methods that fail to account for intra-consumer variability may miss important distinctions among consumers that are particularly relevant for such programs. An effective clustering approach for consumer segmentation must therefore capture within-consumer behavioural diversity and allow the clustering process to take this facet into account.

\subsubsection*{(ii) Inability to represent common intra-consumer variation in clusters} \label{Sec:LiteratureReview-Limitations(ii)}

In parallel with the tendency for existing methods to summarise individual consumer behaviour over long periods using a single RLP, by default the behaviour of entire groups of consumers tends to be summarised by a single prototypical DLP \cite{Michalakopoulos2024APrograms,Meng2023,Qiu2023PersonalizedMarket}. The dominant approach is to reduce the load data across $m$ consumers and $p$ days into $k$ clusters (with $k \ll m$), each represented by one characteristic DLP \cite{Motlagh2019}. 
If the intra-consumer variability missed by a singular RLP is substantial, then the behavioural diversity missed when summarising multiple consumers with a single profile is likely even more pronounced. This reduction is not necessarily an explicit design choice optimised for applications of consumer segmentation; rather, it is likely driven by the structure of conventional clustering pipelines which are built around centroid-based representations (e.g. $k$-means), and thus lack the natural capacity to summarise a cluster by more than one prototype. Though richer within-cluster representations are of course possible in principle, they are not standard practice in the literature.

While it is certainly impractical and unproductive to capture every nuance of a group’s behaviour, there is value in allowing more than one representative profile per cluster --- particularly in cases where a group of consumers exhibit the same set of distinct behavioural modes. A more expressive approach to group representation can reveal important behavioural differences relevant to DR program design. For example, one group of consumers may exhibit a dominant usage pattern on 90\% of days, while consumers from another group alternate between two primary behaviours with nearly equal frequency (e.g., 45\% and 40\%). Although both groups exhibit internal consistency, the second group’s multi-modal behaviour is poorly captured by a single prototype. While existing methods may suffice for the first group, program designers may achieve greater success by accounting for the shared behavioural diversity expressed in the second group.

\subsubsection*{(iii) Inability to identify similar consumers on asynchronous schedules} \label{Sec:LiteratureReview-Limitations(iii)}

Another shortcoming of many consumer clustering approaches lies in their reliance on \textit{temporally anchored comparisons} --- that is, consumer usage tends to be compared on strictly the same calendar days, often as a result of aligning long consumption series in time. This framing assumes that behavioural similarity must be expressed concurrently in order to be meaningful for segmentation applications. However, for many DSM and DR applications, it is not critical whether two consumers exhibit similar usage patterns on exactly the same day, but rather whether they exhibit similar patterns at all.

In practice, similar diurnal consumption behaviours may manifest on different days depending on consumers’ schedules and routines. This is particularly relevant in the residential sector, which accounts for roughly a quarter of final electricity consumption in both Australia and the EU \cite{ABS2023EnergyAccount,Eurostat2024Energy}, and represents a major untapped source of DSM and DR flexibility \cite{Parrish2020AResponse}. Unlike the regularity of commercial and industrial demand, residential usage is increasingly shaped by varied employment arrangements. In Australia, for instance, only about 60\% of workers follow a standard Monday–Friday schedule, with the remainder engaged in non-standard, casual, flexible, or shift work; additionally, over a third of employees now usually work from home, a proportion that has risen since the Covid-19 pandemic and is projected to continue growing \cite{abs_working_arrangements_2024,Ku2022,Ng2022}. In practice, this means consumers may be behaviourally equivalent yet out of sync in time --- a similarity that temporally anchored approaches are poorly equipped to capture.

Temporally anchored clustering approaches include long time series methods which rely on order-sensitive features such as frequency-domain features, or time series model coefficients. These long series approaches are typically implemented using synchronised input windows, and are sensitive to the order in which behaviours are expressed across diurnal cycles. Similarly, representing consumers by their raw long time-series and using elastic distance measures such as Dynamic Time Warping (DTW) \cite{Holder2023AClustering}, which introduce a degree of local warping, maintains temporal anchoring of consumer series globally. Meanwhile, the ensemble method from Sun et al. \cite{Sun2020} is the only two-stage method that enforces temporal anchoring. While not temporally anchored, the DCP representation could only identify asynchronous similarity between consumers if by chance they share the same dominant pattern, with multi-modal similarity likely to be missed.

By enforcing calendar alignment, these methods can fragment otherwise coherent groups of consumers, particularly in residential settings where routines are increasingly asynchronous. This limits the effectiveness of segmentation and undermines the value of DSM and DR initiatives for both retailers and consumers.

\subsubsection*{(iv) Sensitivity to behavioural anomalies} \label{Sec:LiteratureReview-Limitations(iv)}
Most existing consumer clustering approaches implicitly assume that all the constituent DLPs within a consumer’s time series are equally informative for grouping, assigning them identical weight during feature extraction or similarity calculations. However, real-world electricity consumption includes anomalous days \cite{Yerbury2024b,Motlagh2019}, which may arise due to special occasions, travel, illness, temporary changes in household occupancy, equipment malfunction or other irregular events. While some of these events constitute a legitimate part of a consumer's usage history, they are typically non-representative of routine behaviour and are usually less relevant for downstream applications. When clustering methods fail to differentiate between typical and atypical usage, even just a few of these behavioural anomalies can exert disproportionate influence over the representation of a consumer, allowing the clustering process to be biased undesirably \cite{Motlagh2019}.

This issue is particularly pronounced for crude dimensionality reduction techniques, such as RLPs or feature extraction methods like Principal Component Analysis \cite{Motlagh2019}, where anomalous behaviours become embedded in the consumer's representation and cannot be disentangled from typical patterns. While one could attempt to mitigate this sensitivity by removing days with extreme values or irregular patterns prior to clustering, a more robust approach would assign greater weight to the most prevalent and stable patterns of behaviour within each consumer’s DLP distribution, while appropriately downweighting atypical behaviour. By shifting the focus onto recurring patterns, and reducing the influence of outliers, clustering can produce more interpretable and actionable groupings, while still acknowledging the full spectrum of behavioural diversity.

\subsubsection*{(v) Requiring time-synchronised input data} \label{Sec:LiteratureReview-Limitations(v)}
Many consumer clustering approaches require time-synchronised input windows, meaning all series must start and end at the same timestamp with no temporal offsets \cite{Motlagh2019}. This constraint becomes problematic when consumers have different quantities of available historical data. Such a scenario is commonly encountered in incomplete deployment programs, where installation dates will vary between consumers. Yet, this scenario will persist indefinitely as households relocate, resetting the quantity of relevant historical data. 

Techniques relying on globally computed representations, such as matrix factorisation (PCA or Singular Value Decomposition) expressly require equivalent input windows to operate. Many local representations such as the Discrete Fourier Transform, Discrete Wavelet Transform, or Symbolic Aggregate Approximation still require uniform-length inputs to generate compatible representations, as input length directly determines the dimensionality of the resulting representations. Similarly, using the long time-series directly necessitates equal-length inputs. For these methods, this limitation forces either truncation of valuable historical data or exclusion of newer consumers with limited history, ultimately reducing the representativeness and inclusivity of the resulting segmentation. Two-stage methods (excepting Sun et al.'s daily ensemble approach \cite{Sun2020}) and model-based methods such as Motlagh et al.'s \cite{Motlagh2019} DCE where models have a fixed number of parameters can more naturally circumvent this limitation.

\subsubsection*{(vi) Sensitivity to regular discontinuities} \label{Sec:LiteratureReview-Limitations(vi)}
For DR programs such as time-of-use tariffs, many energy retailers will make a distinction between weekdays and weekends for residential consumers due to the predominance of commercial consumption during weekdays \cite{Kiguchi2021}. However, some methods are poorly suited to handling the discontinuities that arise when weekday and weekend series are analysed separately --- for instance, when all weekday daily load profiles are concatenated into a single time series with weekend days removed. The resulting discontinuities can introduce spurious temporal dependencies and distort the underlying structure of the data, potentially misleading the clustering process. For example, frequency-based representations such as Discrete Fourier Transform or Discrete Wavelet Transform, time series models, or autocovariance-based representations may become unreliable when applied to discontinuous time series. However, RLPs and two-stage methods are naturally more equipped to handle this scenario.

\subsubsection*{(vii) Unsatisfactory handling of missing data} \label{Sec:LiteratureReview-Limitations(vii)}
Periods of missing data can arise due to device malfunctions, communication interruptions, or other unknown reasons \cite{Hemanth2022,Kuppannagari2021,Ryu2020}. While shorter periods are reasonable candidates for imputation, it is also common to simply remove significantly affected records \cite{Davarzani2016}. In the context of consumer clustering, the extent of record removal that is required depends on the consumer representation method. The aforementioned methods that require time-synchronised or continuous input cannot abide by the minimal removal of individual affected days, requiring instead that the consumer be dropped from the analysis altogether. On the other hand, RLPs, and two-stage methods that first cluster consumer DLPs or pooled DLPs are more robust to the removal of affected days, and maintain greater generalisability and representativeness by including the maximum amount of reliable data.

\subsubsection*{(viii) Limited scalability to real-world populations} \label{Sec:LiteratureReview-Limitations(viii)}

As smart meter deployments continue to expand globally, data mining methods capable of efficiently operating at realistic scales are becoming increasingly important. However, a balance must be struck, as scalability should not come at the cost of significant compromises to efficacy. Many of the proposed consumer clustering approaches seemingly prioritise computational efficiency by applying coarse dimensionality reduction techniques --- like the aforementioned use of averaging to obtain RLPs or short feature vectors. As previously discussed, this sacrifices behavioural nuance and reduces clustering efficacy, ultimately diluting the potential impact of DSM and DR programs.

The scalability of consumer clustering approaches largely depends on whether they employ globally or locally fitted representations. Methods employing global representations, such as GPF or PCA face greater computational challenges than those using local representations fitted independently to each consumer. This is because global methods must jointly process data across all consumers, resulting in memory and computational demands that scale poorly with the number of consumers or the length of their time series. Moreover, when new consumers are added to a dataset, global representations need to be recomputed in full, further undermining their practicality in continuously expanding deployments. In contrast, local methods decouple the representation step across consumers and naturally support parallel implementations. They also adapt more readily to expanding datasets, since representations can be updated independently for each new consumer without requiring recomputation across all consumers.

Consider for instance two-stage methods utilising global representations. The GPF representation proposed by Li et al. \cite{Li2019AStudy} requires clustering all DLPs from all consumers simultaneously, while Nakashima et al.'s \cite{Nakashima2016PerformanceData} GPF approach requires clustering all subsequences generated by a fixed-size sliding window applied to all consumer time series. In contrast, RLPs, DCPs autocovariances, wavelet coefficients, time series models, and CROCS support per-consumer processing for their local representations. This makes them more robust to scaling in terms of both the number of consumers and the volume of data per consumer, and more serviceable in streaming scenarios than their global counterparts. 

Despite the growing need for scalable solutions that maintain representation quality, relatively few studies engage meaningfully with this trade-off \cite{Michalakopoulos2024APrograms,Bedingfield2018}. 
A robust consumer clustering methodology must therefore achieve computational efficiency without sacrificing segmentation accuracy, ultimately capitalising on the substantial investment made in smart metering infrastructure and maximising impact in downstream applications.

%% file: Table---OtherConsumerClusteringStudies.tex
\begin{table}[!htb]
    \scriptsize
    \rowcolors{1}{TableGray}{TableWhite}
    \begin{adjustbox}{center}
        \begin{tabular}{p{0.4cm} p{0.45cm} p{6.8cm} p{2.4cm} p{6cm}} \toprule \hiderowcolors 
        
            \multicolumn{1}{l}{} & \multicolumn{1}{l}{} & \multicolumn{3}{c}{\textbf{Clustering Approach Description\textsuperscript{a}}} \\ \cmidrule(l){3-5}
            
            \textbf{Ref} & \textbf{Year} & \textbf{Representation} & \textbf{Distance} & \textbf{Clustering Algorithm} \\ \midrule
            \showrowcolors

            \cite{Michalakopoulos2024APrograms} & 2024 & RLP and features extracted from RLPs & ED & $k$-means, $k$-medoids, HAC (Average, Complete, Single, and Ward's linkage) and DBSCAN \\[0.2cm]

            \cite{Meng2023} & 2023 & Concatenated monthly RLPs & ED & $k$-means \\[0.2cm]

            \cite{Qiu2023PersonalizedMarket} & 2023 & RLP & --- & Deep embedded clustering \\[0.2cm]

            \cite{Kaur2022BehaviorApproach} & 2022 & 7 attributes computed over one year of consumption & --- & Gaussian mixture models \\[0.2cm]

            \cite{Alonso2022ClusteringGrids} & 2022 & 19 autocorrelation features, median or 95\textsuperscript{th} percentile diurnal profiles. & ED & $k$-means \\[0.2cm]

            \cite{Ahir2022} & 2022 & Long time-series & Manhattan & HAC (Ward linkage) \\[0.2cm]

            \cite{Alonso2020HierarchicalAutocovariances} & 2020 & 96 autocorrelations, 96 partial autocorrelations, or 9 quantile autocovariances & ED & HAC (Complete linkage) \\[0.2cm]

            \cite{Motlagh2019} & 2019 & Delay coordinate embedding map parameters & ED & Not specified \\[0.2cm]

            \cite{Yilmaz2019} & 2019 & RLPs, or mean values alongside averaged standard deviation across 4 time periods in the day & ED & $k$-means\\[0.2cm]

            \cite{Sandels2019} & 2019 & Mean values for 32 strata (from 4 time periods in the day, 4 temperature ranges and 2 day types) & ED & $k$-means \\[0.2cm]

            \cite{Tureczek2018} & 2018 & Wavelet features or autocorrelations & ED & $k$-means \\[0.2cm]

            \cite{Chen2017} & 2017 & PCA features explaining at least $90\%$ of the variance & ED & $k$-means \\[0.2cm]

            \cite{Wang2016b} & 2016 & Markov state transition probability matrices computed on a Symbolic Aggregate Approximation of the load data ($3\times 3$ for each of 4 time periods in the day)  & Kullback-Leibler Divergence & Clustering by Fast Search and Find of Density Peaks \\[0.2cm]

            \cite{Chicco2012a} & 2012 & RLP & Minkowski Metric with $p=1,2,3,4,5$ & $k$-means, Fuzzy $c$-means, Follow the Leader, HAC (Average and Single linkage) \\[0.2cm]
                            
            \bottomrule \hiderowcolors
            \multicolumn{5}{p{\dimexpr\textwidth-2\tabcolsep-2\arrayrulewidth+0.2cm}}{\textsuperscript{a}\scriptsize Density-Based Spatial Clustering of Applications with Noise (DBSCAN), Euclidean Distance (ED), Hierarchical Agglomerative Clustering (HAC), Principal Component Analysis (PCA), Representative Load Profile (RLP)} \\
        \end{tabular}
    \end{adjustbox}
    \caption{Studies that have proposed consumer clustering approaches, and the representation methods, distance measures and clustering algorithms of which they are composed. These methods all utilise a single stage of clustering, while other multi-stage methods are collected in \Cref{Tab:OtherTwoStageClusteringMethods}.}
    \label{Tab:OtherCustomerClusteringStudies}
\end{table}

%% file: Table---OtherTwoStageClusteringMethods.tex
\begin{table}[!tb]
    \scriptsize
    \rowcolors{1}{TableGray}{TableWhite}
    \begin{adjustbox}{center}
        \begin{tabular}{p{2.3cm} | p{2.5cm} p{2cm} p{4cm} p{4.5cm}} \toprule \hiderowcolors 
            \textbf{Categorisation} & \textbf{Stage One} & \textbf{Stage Two} & \textbf{Energy Domain References} & \textbf{General Time Series References} \\ \midrule 
            \multicolumn{5}{c}{\textcolor{gray}{\textbf{Consumer Clustering}}} \\ \midrule 
            
            \textbf{Representation} & Within-consumers & Consumers & \cite{Tsekouras2007Two-stageCustomers}, CROCS & \n \\[0.2cm]
                                         & Pooled & Consumers & \n & \cite{Nakashima2016PerformanceData}, \cite{Li2019AStudy} \\[0.2cm] 
                                         
            \textbf{Ensemble} & Consumers & Consumers & \cite{Sun2020} & \n \\[0.2cm] \midrule
            \multicolumn{5}{c}{\textcolor{gray}{\textbf{Load Profile Clustering}}} \\ \midrule

            \textbf{Refinement} & Within-consumers & Pooled & \cite{Wang2023Long-TermFramework}, \cite{Banales2021}, \cite{Afzalan2021Two-StageRepresentation}, \cite{Mets2016Two-StageTransformation}, \cite{Li2016Multi-ResolutionData}, \cite{Kwac2014} & \n \\[0.2cm] 

            \textbf{Hierarchical} & Pooled & Pooled & \cite{Toussaint2020}, \cite{Toussaint2019} & \cite{Wang2021AData}, \cite{Zhang2020Two-phaseStations} \\[0.2cm]

                                       & Consumers & Pooled & \n & \cite{Lai2010AAnalysis} \\[0.2cm] 
                            
            \bottomrule \hiderowcolors
        \end{tabular}
    \end{adjustbox}
    \caption{Studies that have proposed two-stage clustering methodologies for general time series and smart meter data specifically. Note that we have categorised the general time series methods as they would be applied to the clustering of smart meter data. Note that for the different stages: ``Within-consumers'' refers to clustering the DLPs (or recurrent sub-periods) of each consumer (or long time-series) independently; ``Consumers'' refers to clustering the consumers (or long time-series); and, ``Pooled'' refers to clustering the DLPs (or recurrent sub-periods) altogether, without regard for which consumer produced the DLP and when.}
    \label{Tab:OtherTwoStageClusteringMethods}
\end{table}

%% file: TheFramework.tex
\section{CROCS: The Two-Stage Consumer Clustering Framework}
\label{Sec:TheFramework}

This section presents the proposed \textbf{Clustered Representations Optimising Consumer Segmentation} \textbf{(CROCS)} framework in detail. We begin with the first stage, in which each consumer's DLPs are clustered independently to produce their RLS. We then introduce the WSMD set-to-set dissimilarity measure that drives the second stage. Together, these two stages produce a consumer clustering, as depicted in \Cref{Fig:CROCS}. Finally, we describe the optional graph-based community detection procedure for computing RRLSs, which enhance the interpretability of the segmentation through multi-profile summaries of the shared diurnal behaviours defining consumer clusters.

\afterpage{
    \clearpage
    \begin{landscape}
        \begin{figure}
            \centering
            \includegraphics[width=\linewidth]{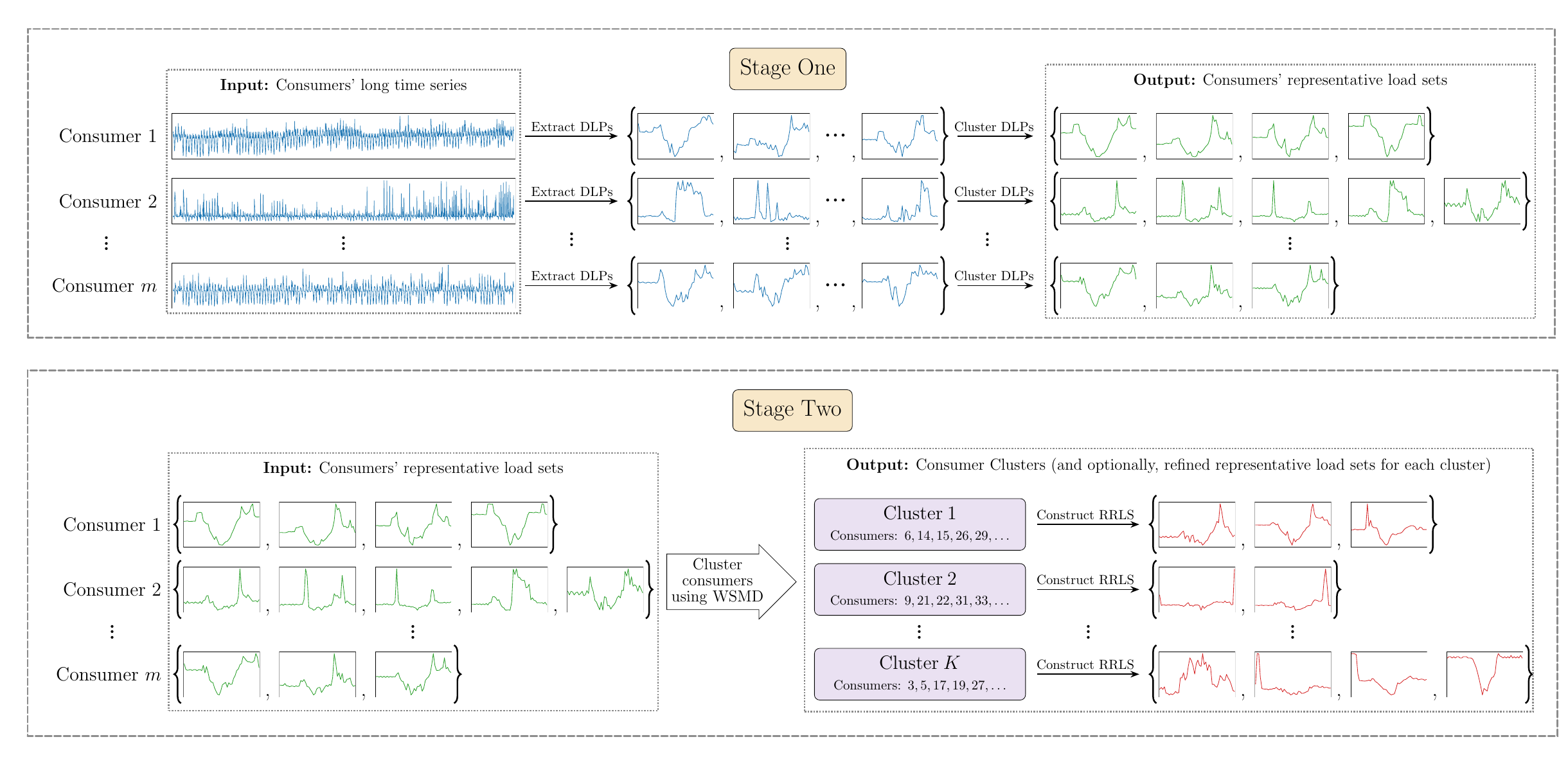}
            \caption{A visual depiction of the two-stage CROCS framework.}
            \label{Fig:CROCS}
        \end{figure}
    \end{landscape}
\clearpage
}

\subsection{Stage One: Clustering Consumers' Daily Load Profiles} \label{Sec:TheFramework_StageOne}

The first stage of CROCS addresses the fundamental challenge of representing each consumer's load behaviour for clustering in a way that is compact, yet faithful to their behavioural diversity. Rather than the popular approach of collapsing a consumer's entire history into a single aggregate profile, stage one independently clusters each consumer's DLPs. The resulting cluster prototypes capture an interpretable summary of the consumer's typical diurnal consumption patterns in a \textbf{Representative Load Set} \textbf{(RLS)}. As a flexible unsupervised representation, the RLS can be adapted to the inherent complexity of each consumer’s load behaviour, such that consumers with simple regular routines may require only a few representative patterns, while those with more diverse usage could be characterised by richer sets. While smart meter time series contain various natural periods (e.g. weeks or years), stage one clusters the 24-hour diurnal period as it is the fundamental periodic unit of energy consumption.


In a more general context, this first stage clusters periodic units of long time series to derive representations for those series. Depending on the time series domain, the periods could span minutes (e.g. machine operation cycles), weeks (e.g. retail demand), or years (e.g. agricultural yields). We therefore introduce notation for a general dataset composed of long time series produced by $m$ distinct entities. For the $i\textsuperscript{th}$ entity, suppose we have observed $p_i$ periods, each consisting of $\phi$ measurements taken at regular intervals.


Having thus acknowledged the general case, we constrain subsequent discussion to the context of smart meter time series data. As such, $m$ represents the number of distinct consumers, $p_i$ represents the number of days of data considered for the $i\textsuperscript{th}$ consumer, and $\phi$ relates to the sampling frequency of the observations (e.g. $\phi = 24$ or $48$ for hourly or half-hourly sampling rates respectively). We denote the $i\textsuperscript{th}$ consumer's long time series as $$\xx_i = 
    \begin{bmatrix} 
        x_{i,1} & x_{i,2} & \cdots & x_{i,\phi} & x_{i,\phi+1} & x_{i,\phi+2} & \cdots & x_{i,2\phi} & x_{i,2\phi+1} & x_{i,2\phi+2} & \cdots & x_{i,p_i \phi} \\
    \end{bmatrix},$$
where $x_{i,j}$ is the energy consumption of the $i\textsuperscript{th}$ consumer through time $j$. Equally, with the $i\textsuperscript{th}$ consumer's $\ell\textsuperscript{th}$ DLP represented as $\xx_{i}^{\ell} = 
    \begin{bmatrix} 
        x_{i, \ell \phi + 1} & x_{i, \ell \phi + 2} & \ldots & x_{i, \ell \phi + \phi},
    \end{bmatrix}
$ for $\ell \in \left\{0, 1,\ldots,p_i-1 \right\}$, we can instead consider the set of each consumer's DLPs, as given by $\mathcal{S}_i = \{ \xx_i^{0}, \xx_i^{1}, \ldots, \xx_i^{p_i-1} \}$. By convention, each consumer's long time series is segmented into DLPs by partitioning at fixed midnight-to-midnight boundaries, though this could in principle be shifted if required for a particular application.

In the first stage of CROCS, clustering is applied independently to each consumer to obtain a partition of $\mathcal{S}_i$ for each $i \in \{1,2,\ldots,m\}$ into $k_i$ clusters\footnote{Note that these are assumed to be \textit{hard} partitions. Consider a set $\mathcal{X} = \{\xx_1, \xx_2, \ldots,\xx_m\}$. A hard partition of $\mathcal{X}$ is a collection of $k$ subsets $\bm{\mathcal{C}} = \{\mathcal{C}_1, \mathcal{C}_2, \ldots, \mathcal{C}_k\}$ such that $\bigcup_{i=1}^k \mathcal{C}_i = \mathcal{X}$, all $\mathcal{C}_i \neq \varnothing$, and $\mathcal{C}_i \cap \mathcal{C}_j = \varnothing$ for $i \neq j$, where $\varnothing$ is the empty set.}. This independence makes stage one straightforward to parallelise across consumers, offering greater computational efficiency when scaling to larger populations. As depicted in \Cref{Fig:CROCS}, the number of clusters composing the RLS can vary from consumer to consumer, or it can be fixed for all consumers. If allowed to vary, the number of clusters can be selected autonomously using an appropriate RVI \cite{Yerbury2024}. This is the methodology applied in the similar first stage of the DCP-based approach in Tsekouras et al. \cite{Tsekouras2007Two-stageCustomers}. A simpler and more robust approach to the selection of $k_i$ that fixes the same value ($k$) for all consumers is discussed in \Cref{Sec:Results_CROCS_StageOne_Picking_k}. Subsequently, the resulting prototypes for each consumer are contained in their RLS, $\hat{\mathcal{S}}_i = \{ \bm{\pi}_i^1, \bm{\pi}_i^2,\ldots, \bm{\pi}_i^{k_i} \}$. Note that each prototype $\bm{\pi}_i^j$ in the RLS represents a cluster containing $n_i^j$ DLPs, such that $\sum_{j=1}^{k_i} n_i^j = p_i$.

CROCS should be regarded primarily as a framework for approaching consumer clustering. As such, we do not prescribe any particular clustering approach (i.e. normalisation procedure, representation method, distance measure, clustering algorithm and prototype definition) for constructing RLSs --- the framework is compatible with any of the plethora of existing clustering components. 
However, for a comparison of representation methods, distance measures and clustering algorithms applied specifically for clustering DLPs as here in stage one, we direct the reader to our previous comparative study \cite{Yerbury2024b}. Meanwhile, the aforementioned prototypes could be for instance average, medoid or median profiles as the practitioner deems optimal for the given application. Furthermore, aligning with an earlier point regarding the utility of normalisation to direct a clustering algorithm to find similarity based on shape rather than magnitude, we recommend some form of normalisation (e.g. min-max or $z$-normalisation \cite{Yerbury2024}) be applied to the DLPs prior to clustering.

\subsection{Stage Two: Clustering Consumers} \label{Sec:TheFramework_StageTwo}

The second stage of CROCS performs a clustering of the consumers, that is, the $m$ entities producing the long time series $\xx_i$. Unlike the first, this stage makes use of a very particular representation method --- the RLSs ($\hat{\mathcal{S}}_i$) obtained from the first stage of clustering. 

Comparing consumers for clustering via their RLSs requires a distance measure tailored to their structure, and for this purpose we introduce the \textbf{Weighted Sum of Minimum Distances} \textbf{(WSMD)}. WSMD extends the original Sum of Minimum Distances \cite{Eiter1997} by incorporating information about the size of the clusters represented by each prototype in $\hat{\mathcal{S}}_i$. In doing so, it ensures that all prototypes contribute to the pairwise consumer dissimilarities in proportion to the frequency of the behaviours they capture.

\begin{figure}[!hbtp]
    \centering
    \includegraphics[width=\linewidth]{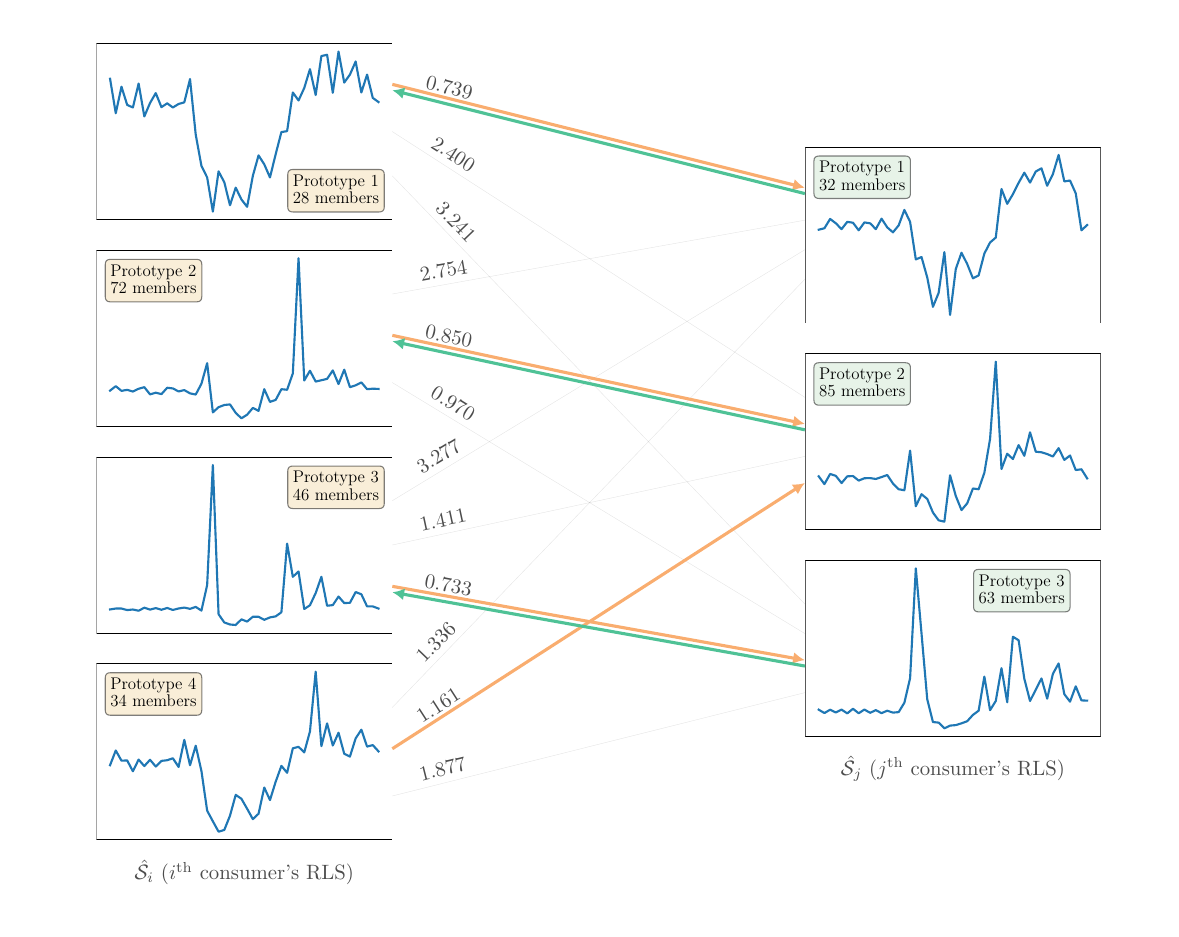}
    \caption[]{A visualisation of the WSMD set-to-set distance measure applied to a pair of consumer representative load sets, $\hat{\mathcal{S}}_i$ (left) and $\hat{\mathcal{S}}_j$ (right), with prototype frequencies and prototype dissimilarities annotated. The resulting WSMD value is computed as per \Cref{Eq:WSMD_Example}.}
    \label{Fig:WSMD}
\end{figure}

In particular, suppose that $d\left (\cdot, \cdot \right)$ is a time series dissimilarity measure that can be applied to DLPs, then the WSMD between $\hat{\mathcal{S}}_i$ and $\hat{\mathcal{S}}_j$ is given by
\begin{equation*}
    \Delta_{\text{WSMD}} \left( \hat{\mathcal{S}}_i, \hat{\mathcal{S}}_j \right) = \frac12 \left[\frac{1}{p_i}\sum \limits_{a=1}^{k_i} n_i^a \min \limits_{b} d\left( \bm{\pi}_i^a, \bm{\pi}_j^b \right) +  \frac{1}{p_j}\sum \limits_{b=1}^{k_j} n_j^b \min \limits_{a} d\left( \bm{\pi}_j^b, \bm{\pi}_i^a \right)\right].
\end{equation*}
For each prototype in a consumer's RLS, WSMD finds the closest prototype from the other consumer's RLS. The distance between consumers is then taken as the weighted average of these minimum distances. While any time series dissimilarity measure can be used within the WSMD, consistency can be achieved by using the same measure as was used to cluster consumer DLPs in the first stage. A WSMD computation between two consumers has been visualised in \Cref{Fig:WSMD}, where 
\begin{multline}
\label{Eq:WSMD_Example}
    \Delta_{\text{WSMD}} \left( \hat{\mathcal{S}}_i, \hat{\mathcal{S}}_j \right) = \frac12 \Biggl[ \frac{1}{180} \left( 28\cdot0.739 + 72\cdot0.850 + 46\cdot0.733
    + 34\cdot1.161 \right) \\ + \frac{1}{180} \left( 32\cdot0.739 + 85\cdot0.850 + 63\cdot0.733\right) \Biggr] = 0.825.
\end{multline}

The second stage of the CROCS framework then achieves a partition of consumers into clusters by pairing the WSMD distance measure and RLS representation method with any existing clustering algorithm.

\subsection{Computing Refined Representative Load Sets} \label{Sec:TheFramework_RRLS}

Consumer clustering methods often struggle to provide interpretable and meaningful explanations for why particular consumers have been grouped together, often leaving practitioners with little insight into what shared behavioural patterns define consumer groups. The CROCS framework addresses this interpretability challenge by offering practitioners the option of generating a \textbf{Refined Representative Load Set} \textbf{(RRLS)} for each consumer cluster. Each RRLS contains one or more prototypical DLPs that describe the various ways that consumers within a cluster typically use their energy. 
Unlike single-prototype cluster summaries, the RRLS summary more accurately captures distinct behavioural modes shared among consumers within a cluster. Furthermore, the frequency, or ``coverage'' of each hyperprototype is quantifiable, indicating its prevalence across both consumers and days. This optional summary step in the CROCS framework transforms opaque consumer cluster assignments into interpretable behavioural profiles, enabling practitioners to understand not just which consumers belong together, but precisely what diurnal consumption patterns they share.

\begin{figure}[!hbtp]
    \centering
    \includegraphics[width=1\linewidth]{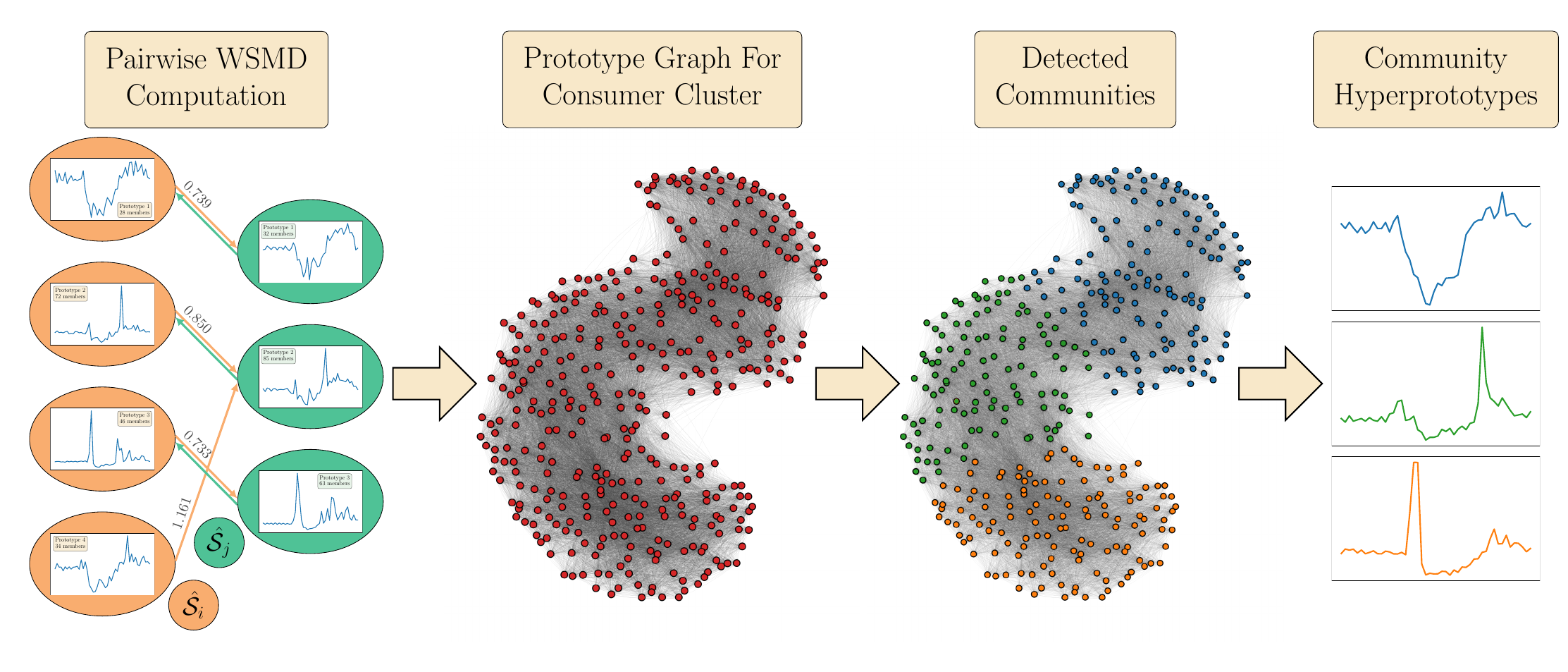}
    \caption[]{A visualisation demonstrating computation of the RRLS hyperprototypes for consumer clusters.}
    \label{Fig:RRLS}
\end{figure}

The process of computing an RRLS for an individual cluster of consumers is depicted in \Cref{Fig:RRLS}. To begin, the prototype graph or network induced by all included consumers' pairwise WSMD computations is constructed. In such a graph, the vertices represent the consumers' RLS prototypes, and the edges reflect the \textit{directional} connections discovered between those prototypes by WSMD, i.e. the coloured edges shown in \Cref{Fig:WSMD}. These directed edges are weighted according to the pairwise dissimilarity of the prototypes. However, because WSMD utilises a dissimilarity measure $d(\cdot,\cdot)$, and smaller dissimilarities should correspond to stronger connections, an appropriate transformation must be applied to convert the dissimilarities into similarities --- for example, by assigning weights as $1/d(\cdot,\cdot)$\footnote{While such transformations are required for a majority of community detection algorithms, some methods have been proposed that directly operate on dissimilarities (or length data) as edge weights \cite{Molnar2024CommunityPartitioning}, i.e. smaller edge weights are interpreted as stronger connections.}. Furthermore, the vertices are weighted according to the number of DLPs represented by each consumer prototype. For instance in \Cref{Fig:WSMD}, the vertex representing prototype 4 for the $i$\textsuperscript{th} consumer (bottom left) would have a weight of 34. Additionally, a directed edge would exist to consumer $j$'s 2\textsuperscript{nd} prototype (middle right) with a weight of $1/1.161 = 0.861$.

Densely, or strongly, connected regions of these prototype graphs represent shared modes of usage within the consumer cluster. Identifying these densely connected sub-graphs allows us to compute refined prototypical diurnal profiles that capture those shared modes of usage. Thus with the prototype graph constructed, any community detection algorithm compatible with weighted directed graphs can be applied to obtain a partition of the vertices \cite{Fortunato2010CommunityGraphs,Fortunato2016}, as depicted in the third panel of \Cref{Fig:RRLS}. For each detected community, a corresponding hyperprototype is obtained (depicted in the fourth panel of \Cref{Fig:RRLS}) as the prototype of the included vertices. 

The number of distinct consumers and days covered by the vertices in each community provides an indication of the coverage of the corresponding hyperprototypes. While practitioners can choose the number of communities (i.e. the number of hyperprototypes in the RRLS) to suit their application --- allowing for either concise or detailed behavioural summaries --- the quantification of coverage is most meaningful when seeking to identify the natural number of behavioural modes present in the data. However, using medoid prototypes offers an effective approach for representing core consumption patterns while remaining robust to occasional anomalous behaviours, thereby reducing sensitivity to the choice of number of communities.

Whilst an argument could be made for simply clustering the union of all constituent consumer RLS prototypes, the graph-based approach instead preserves the explicit similarity structure established during the WSMD computation, maintaining methodological consistency between the consumer clustering and hyperprototype extraction. Similarly, although an RRLS could in principle be computed for consumer partitions generated by any segmentation method, pairing it with CROCS ensures that the shared behaviours captured by the RRLS are precisely those that drove the segmentation --- not patterns identified post hoc that may be unrelated to the basis on which consumers were grouped.

For intuition, consider the ideal scenario where consumers in a cluster share exactly $n_c$ similar consumption patterns. Stage one clustering would ideally discover these patterns, resulting in each consumer being represented by $n_c$ prototypes in their RLS. Hence the resulting prototype graph would be composed of exactly $n_c$ connected components. With each of the connected components readily identifiable as a community, the consumer cluster would then be described by an RRLS with $n_c$ hyperprototypes --- one per common consumption behaviour. In reality, the prototype networks will be much more complex, but the underlying principle remains: densely connected regions of the prototype graph reveal the core behavioural modes that unite consumers within a cluster.

%% file: Methodology.tex
\section{Experimental Setup}
\label{Sec:Methodology}

This section presents the datasets, metrics, and implementation details used to evaluate CROCS in our subsequent experiments.

\subsection{Real Data}
Throughout the course of our experiments we make use of two publicly available Australian smart meter datasets: the Ausgrid Solar Home Electricity (AG) dataset \cite{Ausgrid2013,Ratnam2017} and the Smart-Grid Smart-City (SGSC) dataset \cite{Ausgrid2014}\footnote{Links to these datasets are provided on the associated \href{https://github.com/yerbles/CROCS}{GitHub} repository}.

The AG dataset comprises 300 households in New South Wales with rooftop photovoltaic (PV) systems, each with 1,096 half-hourly sampled DLPs recorded between 1st July 2010 and 30th June 2013. Meters separately recorded gross household consumption, PV generation, and (for some households) an off-peak controlled load. In line with typical interval metering --- which records combined imports and exports rather than separately metered consumption and generation --- we derived net household demand by combining general and controlled load and subtracting PV generation.

The SGSC dataset was used in the form prepared by \cite{Roberts2020}, where extensive pre-processing and filtering was applied. The resulting subset comprises 3,953 households in New South Wales, each with 365 half-hourly DLPs from 1 January to 31 December 2013, also representing net household demand.

\subsection{Synthetic Data}

We also present results from experiments using synthetic datasets, where a ground truth partition is available. For this purpose we employed the synthetic DLP generator introduced in \cite{Yerbury2024b}. This generator produces min-max normalised half-hourly DLP instances that conform to one of 20 fundamental cluster \textit{shapes}. Using the baseline parameter settings specified in \cite{Yerbury2024b}, the generator produces unique profiles through a stochastic process, while ensuring profiles derived from the same shape form recoverable clusters. The 20 shapes represent distinct residential consumption patterns, differing by the timing, number, shape and relative magnitude of peak consumption events. The generator is also capable of producing outlier DLPs that don't conform to any of the 20 cluster shapes with high probability, and don't form clusters of their own.

To construct a synthetic dataset, the specified number of consumers were distributed among $K^*$ true clusters, with cluster sizes drawn from a multinomial distribution with uniform probabilities, emulating a balanced consumer distribution across clusters. Each synthetic consumer cluster was defined by drawing a unique subset of $k^*$ DLP shapes from the available 20, ensuring that no two consumer clusters had identical sets of defining shapes. For each consumer, DLPs were then generated as new instances from random shapes within their cluster’s subset until the appropriate number of days was obtained. As required, a controlled number of those profiles ($n_O$) were instead designated to be outliers that didn't conform to any of the 20 shapes. Further details regarding the composition of the synthetic datasets will be provided as the experiments are introduced.

Various investigations in the present study --- such as evaluating CROCS design choices and benchmarking CROCS against competing methodologies --- are necessarily conducted through controlled experiments on synthetic datasets where the true underlying cluster structure is known, as equivalent analyses with real data are subject to well-recognised obstacles. Firstly, the authors are not aware of any real smart meter datasets with reliable ground-truth consumer labels. Even if such datasets were available, a large and diverse collection would be required to systematically vary the same key characteristics in a controlled manner, rendering a systematic investigation impractical. More fundamentally however, multiple structurally distinct partitions could provide equally valid representations of the same complex real-world behavioural data \cite{Farber2010,Gagolewski2021}. Consequently, differences observed between clusterings obtained under different parameter settings could not be unambiguously attributed to improved or degraded performance. In the absence of ground truth, one must instead rely on RVIs. While these are commonly employed, no consensus exists on which indices are most suitable for this domain, and many favour domain-agnostic cluster concepts that may not adequately reflect relevant consumer segmentation goals \cite{Yerbury2024b}. It is for these reasons that controlled synthetic experiments have been adopted throughout this study, complemented with real data wherever possible.

\subsection{Metrics}
The accuracy with which synthetic consumer clusters were recovered was measured using three External Validity Indices (EVIs): Adjusted Rand Index (ARI) \cite{Hubert1985}, Adjusted Mutual Information (AMI) \cite{Vinh2010}, and Pair Sets Index (PSI) \cite{Rezaei2016}. These indices represent three complementary methodological categories for comparing partitions --- pair-counting (ARI), information-theoretic (AMI), and set-matching (PSI) respectively \cite{Wagner2007}. ARI and AMI are widely recommended due to their adjustment for chance \cite{Milligan1996,Steinley2004PropertiesIndex,Javed2020}, while PSI is also chance-corrected and, uniquely, equally affected by errors in both small and large clusters. Given their different formulations, agreement among ARI, AMI, and PSI demonstrates that our conclusions are robust to the choice of evaluation index. For conciseness, we only reported ARI values where all three measures yielded consistent results.

\subsection{CROCS Configuration}
As previously discussed, CROCS could be configured to use any of the plethora of clustering approaches and components from among the clustering literature. We will now discuss the specific configuration that we elected to use throughout our experiments. 

The first stage of CROCS was implemented with min-max normalisation, applied independently to each DLP. This is a very common normalisation procedure in the smart meter time series clustering literature \cite{Michalakopoulos2024APrograms,Tureczek2018,Chen2017,Wang2016b,Tsekouras2007Two-stageCustomers}, and ensures that DLPs range within the interval $\left[0,1\right]$. Dynamic Time Warping (DTW) \cite{Sakoe1978} with a 1 hour warping window (denoted as DTW-2 for half-hourly data) demonstrated strong performance for clustering of half-hourly DLPs specifically in our previous comparative study \cite{Yerbury2024b} when combined with either Hierarchical Clustering with Ward's linkage (HAC-Wa) or $k$-medoids (KMd) \cite{Aggarwal2014}. Accordingly, DTW-2 is utilised with these algorithms by default. Finally, medoid prototypes have been used to populate RLSs in preference to pointwise means, which can lie in low-density regions of the data subspace, produce excessively smoothed profiles, or become ill-defined under elastic distance measures like DTW --- risks that are substantially mitigated by the medoid's guarantee of being an actual observation from the dataset.

Whilst not strictly necessary, we have elected to use the same clustering algorithm in the second stage of CROCS as was used in the first stage.  We used the Python package \texttt{aeon} for DTW, \texttt{sklearn\_extra} for KMd (using the partitioning around medoids method), \texttt{sklearn} for $k$-means, and both \texttt{scipy} and \texttt{fastcluster} for HAC-Wa. Both KMd and $k$-means were run with 30 initialisations and a maximum of 200 iterations, using the $k$-medoids$++$ and $k$-means$++$ initialisation schemes respectively \cite{ilprints778}. The code used to implement CROCS and reproduce the experiments presented in this paper has been made available on \href{https://github.com/yerbles/CROCS}{GitHub}. Additional details specific to particular experiments are provided in context.

%% file: Results.tex
\section{Results}
\label{Sec:Results}

\subsection{Stage One: How many clusters?} 
\label{Sec:Results_CROCS_StageOne_Picking_k}

A fundamental consideration when implementing the CROCS framework concerns the selection of $k_i$, or the number of clusters to extract from each consumer's set of DLPs in stage one. This parameter directly controls the size and representational complexity of each consumer's RLS. While it is certainly possible to individually optimise $k_i$ for each consumer using RVIs as in \cite{Tsekouras2007Two-stageCustomers}, investigation reveals a surprisingly robust alternative:
\begin{quote}
    Overestimate $k_i$, using the same value for all consumers --- simply $k$. 
\end{quote}

This approach increases the likelihood that the major behavioural modes of consumers are adequately represented in consumers' RLSs. Meanwhile, the weighting of the WSMD set distance ensures that if a single behavioural mode has been split across two or more clusters, that this won't have a significant effect on the pairwise consumer dissimilarities (as demonstrated in \Cref{Sec:Results_CROCS_SetDistanceComparison}). While this incurs minor additional computational complexity, we consider this acceptable given that RVI-based optimisation of each $k_i$ would require generating more partitions with different numbers of clusters. Moreover, initial experimentation suggested that automated RVI optimisation for DLP clustering can show a tendency towards the extremes of the considered range --- either oversimplified clusterings ($k_i=2$) or excessive partitioning ($k_i \simeq p_i$) --- rather than identifying the intermediate value that best captures the behavioural range, a limitation also noted for certain criteria in previous works \cite{Milligan1985,Batool2021a}.

To demonstrate the utility of overestimation, we applied CROCS to synthetic datasets containing $m=300$ consumers with $p_i =90$ days of data each, emulating a season's worth of consumption data. Consumers were distributed among $K^*=8$ clusters, with each consumer's 90 DLPs randomly sampled from their cluster's unique subset of DLP shapes. Two complementary synthetic data scenarios were considered: one without outlier DLPs and values of $k^* \in\left\{ 2,3,\ldots,12 \right\}$, and another where $k^*$ was fixed at 6 and $n_O \in \left\{0,5,10,\ldots,80 \right\}$ random DLPs were replaced with outliers. For each of these combinations, 50 independent datasets were generated.
In both cases, the data were clustered using the CROCS framework, with the number of stage one clusters systematically varied across $k\in\left\{ 2,3,\ldots,90\right\}$. The true number of consumer clusters, $K^*$, was assumed known for the second stage of clustering.

\begin{figure}[!t]
    \centering
    \begin{subfigure}{\textwidth}
        \centering
        \includegraphics[width=\linewidth]{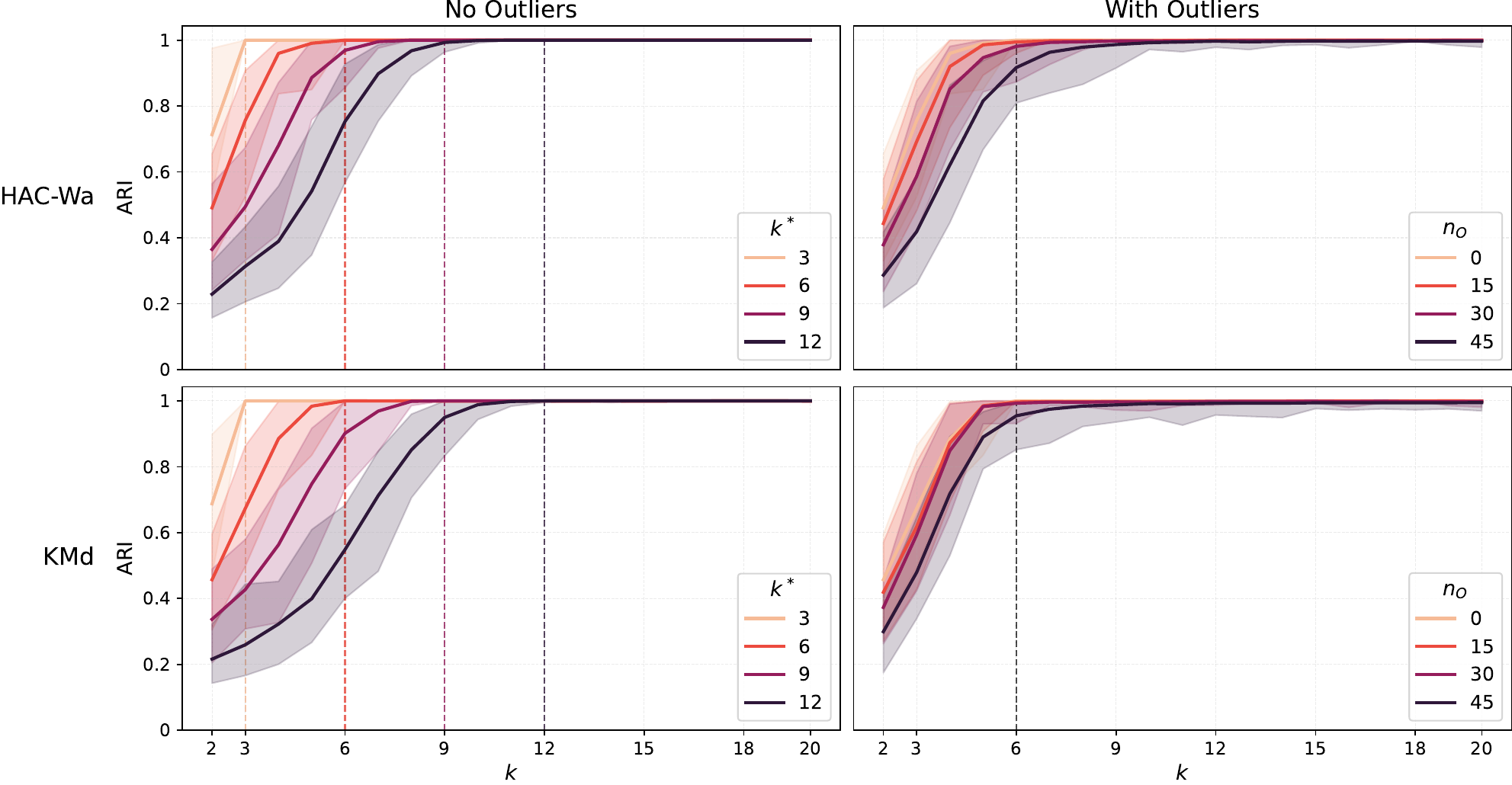}
        \caption{Recovery of Synthetic Consumer Clusters}
        \label{Fig:overestimating_k1_ari}
    \end{subfigure}
    
    \vspace{1em}
    
    \begin{subfigure}{\textwidth}
        \centering
        \includegraphics[width=0.8\linewidth]{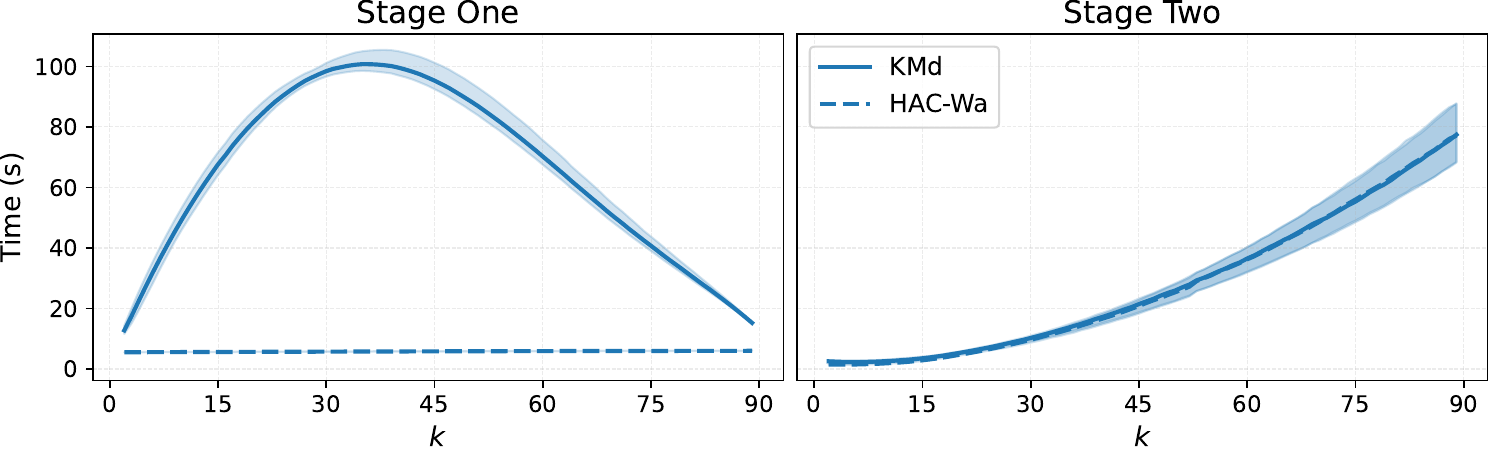}
        \caption{Computational Time}
        \label{Fig:overestimating_k1_timing}
    \end{subfigure}
    \caption{Mean performance and computational time (with pointwise 95 percentile intervals) of the CROCS framework over synthetic datasets with different numbers of true stage one clusters ($k^*$) and different numbers of outliers ($n_O$). (a) Recovery of consumer cluster labels according to ARI for HAC-Wa and KMd with different numbers of stage one clusters ($k$). (b) Computational time (in seconds) for both Stages of the CROCS framework.}
    \label{Fig:overestimating_k1}
\end{figure}

The results from these experiments are presented in \Cref{Fig:overestimating_k1}, with \Cref{Fig:overestimating_k1_ari} displaying consumer ground-truth label recovery according to ARI\footnote{Note that the following observations are consistent for both AMI and PSI as well.  Also, while only a subset of the described scenarios are shown for clarity in \Cref{Fig:overestimating_k1_ari}, these observations were consistent across all combinations of $k^*$ and $n_O$.} when CROCS was applied with either HAC-Wa (top) or KMd (bottom). These experimental results reveal a clear and consistent pattern: overestimating $k^*_i$ with the same value ($k$) for all consumers provides robust performance without compromising the accuracy of consumer cluster recovery. Once sufficient stage one clusters are extracted to capture each consumer's behavioural diversity, further increases in $k$ produce a performance plateau rather than degradation. This finding holds across both experimental scenarios, though the importance of overestimation becomes more pronounced in the presence of outliers. 

Without outliers, it was observed to be acceptable to even mildly underestimate $k^*$ for larger values, likely due to some sharing of the 20 synthetic cluster shapes between consumer clusters --- something which is not unlikely in the context of real data. With outliers, there is a greater need for overestimation to ensure accurate recovery of consumer clusters. According to our results with $k^*=6$, even if 50\% of DLPs were classified as outliers, using $2k^*$ would allow acceptable recovery of consumer clusters. For reference, a real consumer DLP dataset was clustered by domain experts in \cite{Yerbury2024b} which was composed of 365 DLPs, and it was determined that about $26\%$ of those DLPs should be considered as outliers.

The computational implications of overestimation follow predictable scaling patterns, as shown in \Cref{Fig:overestimating_k1_timing}. Stage two exhibits quadratic growth in execution time as a function of $k$, reflecting the $\bigO{k^2}$ pairwise distance computations required between consumer RLSs. Meanwhile, stage one scales with the complexity of the chosen clustering algorithm --- the complexity of KMd depends on the choice of $k$\footnote{For KMd, the algorithm's update step evaluates $k(p_i-k)$ possible medoid swaps, which is smallest when $k$ is near the extremes and largest when $k$ is intermediate, resulting in the observed peak in execution time.}, while the distance matrix and dendrogram computations in HAC-Wa are identical for any $k$. Since the stage one clustering operations can be performed independently for each consumer, execution time can be reduced dramatically through parallel implementation. In contrast, stage two performs a single clustering operation across all consumers, making the quadratic scaling in RLS size the dominant computational bottleneck for a given size dataset. 

In summary, while overestimating $k_i$ with a uniform choice of $k$ seemingly incurs additional computational time, particularly in stage two, these costs represent a reasonable trade-off in light of the simplicity and robustness afforded. The computational overhead remains practical, especially considering that it eliminates the need to select an appropriate set of RVIs, and perform extensive clustering to optimise those RVIs independently for each consumer --- a process that would itself be computationally expensive and potentially less reliable.

\subsection{Comparing Consumer Representations}
\label{Sec:Results_Comparing_Consumer_Representations}

As discussed in \Cref{Sec:LiteratureReview-Limitations(i)}, adequately capturing intra-consumer variation is important for effective consumer clustering, as this variability can serve as an indicator of suitability for DSM strategies. In the CROCS framework, intra-consumer variation is preserved within the RLS in the form of a diverse set of prototypical daily consumption profiles. While the RLS offers comprehensive behavioural representation, it is valuable to consider how it compares to other DLP-based methods from the literature.

As outlined in \Cref{Sec:LiteratureReview-RelatedWorks}, several consumer representation methods are also based on processed DLPs. RLPs are the simplest example, relying on simple point-wise averages across all DLPs. DCPs suggested by Tsekouras et al. \cite{Tsekouras2007Two-stageCustomers} represent each consumer by the prototype of their most populated cluster from a similar stage one clustering to that in CROCS. In contrast, the GPF approach from Li et al. \cite{Li2019AStudy} represents consumers through proportion vectors indicating how frequently their DLPs resemble globally-discovered prototypical DLPs. Unlike the rigid RLP, the representational quality and efficiency of the RLS, DCP, and GPF are particularly dependent upon the number of stage one clusters ($k$) used to generate them. 

To evaluate how effectively each representation captures intra-consumer variability, we computed the reconstruction error (using ED) between real consumer's DLPs and their closest representative profile. For the RLP and DCP representations, which provide only a single representative profile per consumer, all DLPs were compared against this single profile. For the RLS and GPF representations, which rely upon multiple profiles, each DLP was compared against its closest profile from the available set. Given that this concept of reconstruction error aligns with the objective function of $k$-means (within-cluster sum of squares), we have used $k$-means with ED to generate the RLS, DCP, and GPF representations for this comparison\footnote{Though similar results were found when using KMd and HAC-Wa with either ED or DTW-2, and either mean or medoid prototypes.}.

\begin{figure}[!t]
        \centering
        \includegraphics[width=0.8\linewidth]{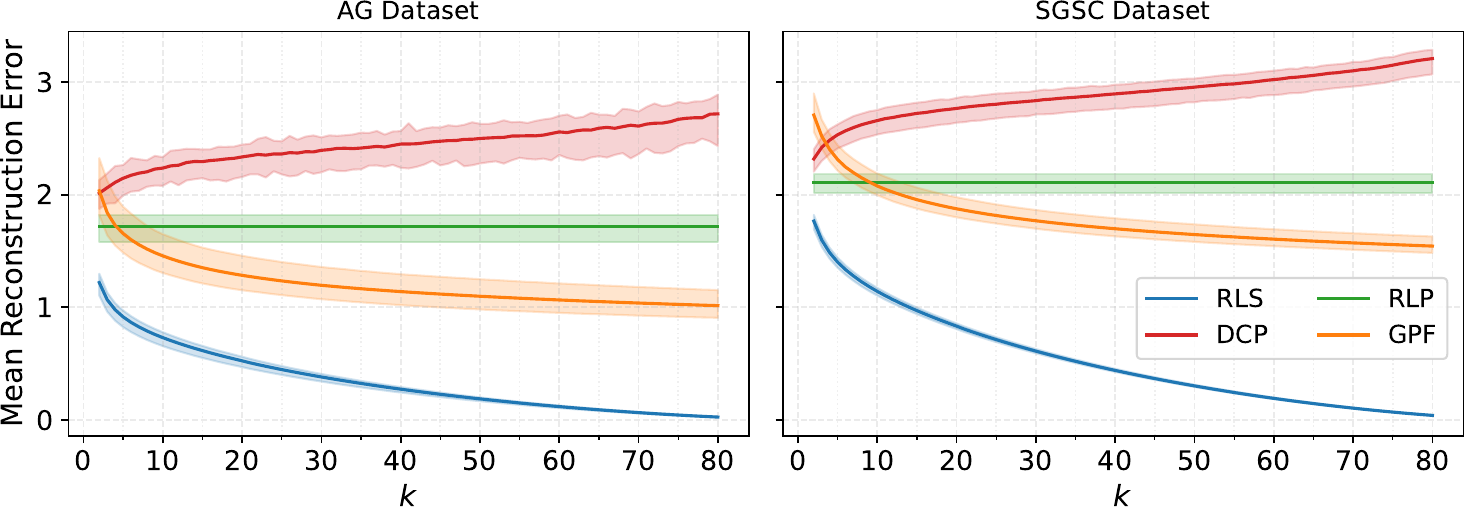}
        \caption{Mean reconstruction error (according to ED) for different load-profile based consumer representation methods across varying numbers of stage one clusters ($k$). Results are averaged across consumers and multiple 90-day periods: 24 periods for the AG dataset (left) and 10 periods for the SGSC dataset (right). Shaded regions show the full range of variation across the different time periods.}        
        \label{Fig:Reconstruction_Error}
\end{figure}

\Cref{Fig:Reconstruction_Error} displays the mean reconstruction error across all consumers for values of $k\in\left\{2,3,\ldots,80\right\}$. The left panel shows results for the AG dataset, averaged across 24 different 90-day periods (one starting on the first day of each month in 2011 and 2012), while the right panel shows results for the SGSC dataset averaged across 10 90-day periods (one starting on the first day of each of the first 10 months of 2013).

\begin{figure}[!tp]
        \centering
        \includegraphics[width=0.55\linewidth]{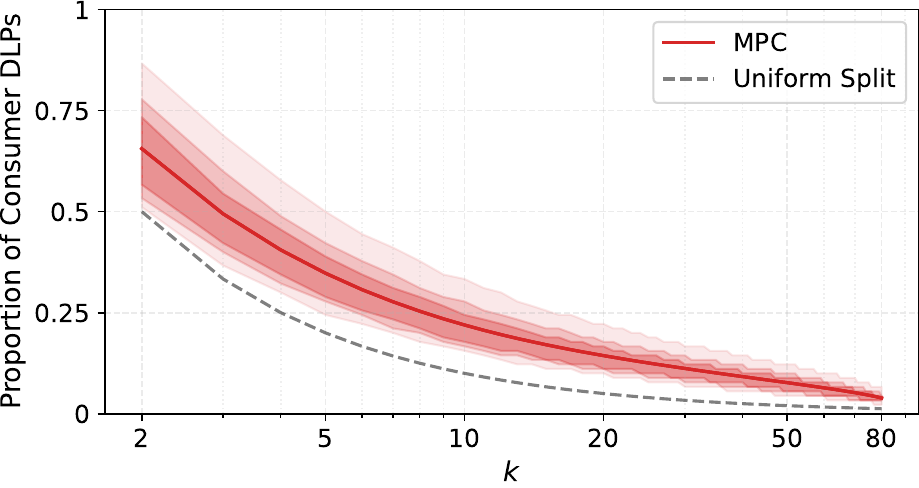}
        \caption{Proportion of DLPs appointed to the largest cluster across different values of $k$ (displayed on a log-scale). The red line shows the mean proportion, with shaded regions representing the central 50\%, 70\%, and 90\% intervals of the distribution. Results are based on all consumers from the AG and SGSC datasets over the same 90 day periods behind \Cref{Fig:Reconstruction_Error}, but clustered using DTW-2 with either HAC-Wa or KMd algorithms to achieve more representative partitions. The grey dashed line indicates the expected proportion under uniform cluster size distribution for comparison.}
        \label{Fig:Tsekouras_Proportions}
\end{figure}

Clearly, the multiple profiles in the RLS and GPF representations offer a significant advantage over the single profile representations. In particular, the single DCP exhibits a higher reconstruction error for all $k$ than even the simple RLP. The quality of the DCP representation in fact deteriorates as $k$ is increased. This deterioration occurs because as $k$ increases, each consumer's DLPs are distributed across more clusters, causing the most populated cluster to represent a progressively smaller fraction of the consumer's behaviour. \Cref{Fig:Tsekouras_Proportions} demonstrates this effect using the same datasets as \Cref{Fig:Reconstruction_Error}, showing the proportion of DLPs assigned to each consumer's largest cluster for different values of $k$. This particular analysis used DTW-2 distance with both HAC-Wa and KMd algorithms, with results aggregated across both clustering methods and all AG and SGSC datasets due to the consistency of the trend. The mean proportion only marginally exceeds what would be expected under uniform random partitioning into $k$ groups, following a similar declining trend as $k$ increases. By the time $k$ reaches 3, the largest cluster represents only about half of the DLPs --- not to mention that such a clustering is very unlikely to adequately segment a consumer's DLPs across their true set of behavioural modes.

Unlike DCP, the reconstruction error for RLS and GPF decreases steadily as $k$ increases, approaching zero at the limits of $p$ and $mp$, respectively. However, comparing them directly at the same $k$ is not straightforward as RLS uses $k$ local prototypes per consumer (totalling $mk$), whereas GPF uses $k$ global prototypes shared across all consumers. In principle, with sufficiently large $k \gg p$, GPF could achieve the same reconstruction error of RLS. In practice, however, GPF has typically been applied with small $k$ values. For instance, Li et al. \cite{Li2019AStudy} used $k \in \left\{2,\ldots,10\right\}$ for $m=1\,463$ vehicles over one year, while Nakashima et al. \cite{Nakashima2016PerformanceData} used $k \in \left\{30,50,100\right\}$ for datasets with $mp$ on the order of $10^4$. This preference for modest $k$ stems from both computational and statistical issues. As $k$ increases, the dimensionality of the frequency vectors grows, causing the distances between them to become increasingly concentrated, thereby reducing their discriminative power \cite{Aggarwal2001}. At the same time, meaningful behavioural modes risk being fragmented across multiple clusters, so that subtle similarities between consumers are obscured when their counts are split across distinct dimensions. These factors constrain GPF’s practical effectiveness, making comparisons with $k \in \left\{2,\dots,80\right\}$ more representative of practical usage. Within this range, the localised RLS approach consistently provides lower reconstruction error, highlighting its advantage over global representations for efficiently capturing intra-consumer variability.

\begin{figure}[!htbp]
    \centering
    \includegraphics[width=0.9\linewidth]{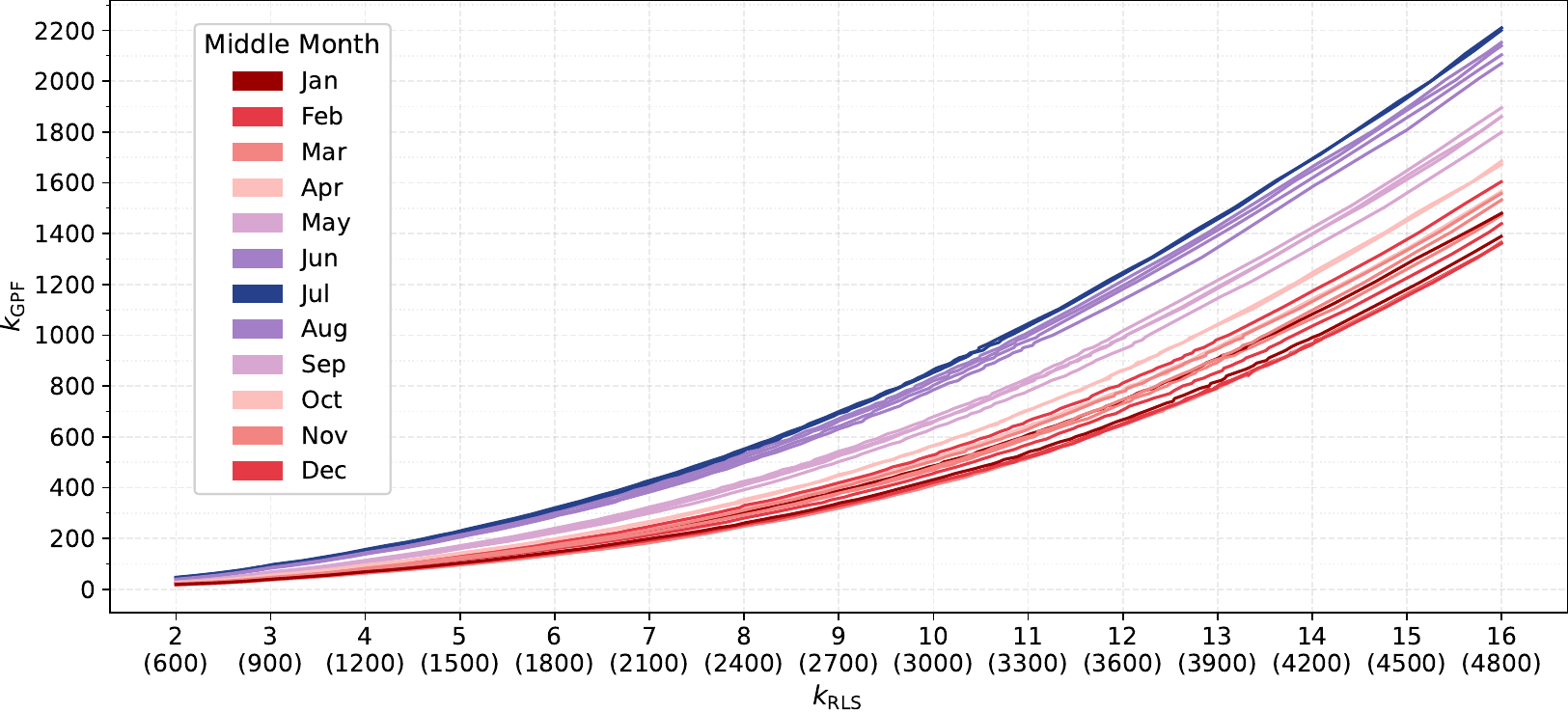}
    \caption{Values of $k_{\text{GPF}}$ ($y$-axis) and $k_{\text{RLS}}$ ($x$-axis) for the GPF and RLS representations respectively that achieve identical reconstruction errors across the 24 90-day periods from the AG dataset. Both representations have been computed using ED and $k$-means. Line shading reflects the climate of the midpoint month, with a palette mirrored around January and July to emphasise seasonal trends in the relationship between equivalent $k_{\text{GPF}}$ and $k_{\text{RLS}}$. Values of $k_{\text{RLS}}$ are provided on the $x$-axis with the effective number of representative profiles across all consumers in brackets underneath ($m\times k_{\text{RLS}}$).}
    \label{Fig:EquivalenceBetweenRLSandGPF}
\end{figure}


Despite the practical restrictions of applying GPF with large $k$ in practice, it is insightful to examine the relationship between the number of clusters required by GPF and RLS to achieve identical reconstruction errors. \Cref{Fig:EquivalenceBetweenRLSandGPF} shows the values of $k_{\text{GPF}}$ (y-axis) and $k_{\text{RLS}}$ (x-axis) that produce equivalent reconstruction errors for the AG datasets analysed in \Cref{Fig:Reconstruction_Error}. Linear spline interpolation was used to determine these equivalent cluster numbers across different starting months.

The results reveal significant redundancy in the RLS representation. For instance, when $k_{\text{RLS}}=5$, there are effectively 1500 local representative profiles for the AG consumers, yet the GPF representation suggests that only 100-200 global representative profiles would achieve the same reconstruction error. Interestingly, seasonal variation in this redundancy is also evident from the colour-coded lines in \Cref{Fig:EquivalenceBetweenRLSandGPF}. Again for $k_{\text{RLS}}=5$, approximately 100 global profiles provide equivalent reconstruction error during summer months, while over 200 are required during winter. This pattern suggests greater diversity in consumption behaviours during cooler months, potentially due to variability among households in terms of insulation quality, heating systems, temperature tolerance, and other demographic factors.

Rather than representing a limitation, this redundancy in the RLS representation supports the premise that consumers exhibit similar energy usage patterns, validating the potential effectiveness of consumer clustering. The CROCS framework leverages this redundancy to group consumers based on their shared behaviours. 

\subsubsection*{Computational Complexity}

While the RLS representation involves more individual representative profiles than the GPF representation, it offers considerable computational advantages. Consider RLS, where each of $m$ consumers' $p$ DLPs are independently clustered using HAC-Wa\footnote{Which has complexity $\bigO{n^2}$ for $n$ objects \cite{Mullner2013}.}, resulting in a complexity of $\bigO{mp^2}$\footnote{This assumes a distance measure with complexity $\bigO{\phi}$, as with ED or indeed DTW with a small fixed warping window. This factor is dropped in this context as $\phi$ is a fixed constant determined by the sampling rate.}. GPF meanwhile clusters all $mp$ DLPs together, at a complexity of $\bigO{m^2p^2}$, scaling poorly with the number of consumers in the dataset. When combined with the second stage of clustering, involving $k^2$ WSMD distance computations for each consumer pair, the complexity of the CROCS framework is $\bigO{mp^2+k^2m^2}$. Meanwhile, using the GPF for consumer clustering collectively achieves a complexity of $\bigO{m^2p^2+km^2}$. The squared dependence on $m$ in GPF's first stage is traded for a linear dependence on $m$ in CROCS' first stage, while the linear dependence on $k$ in GPF's second stage becomes quadratic in CROCS. Since $k\ll m$ in practice, this trade-off represents a substantial computational advantage for CROCS, making consumer clustering feasible at scales where GPF approaches would become computationally intractable\footnote{As modular frameworks, the precise complexity of both CROCS and GPF depends on the chosen clustering components. However, the asymptotic advantage of CROCS over GPF holds for shared components.}. Moreover, CROCS also benefits further from straightforward parallelisation across consumers in its first stage.

\subsection{Stage Two: Is WSMD the best set distance measure?} 
\label{Sec:Results_CROCS_SetDistanceComparison}

WSMD was designed specifically for comparing consumer RLSs. However, any set-to-set distance could in principle be employed in stage two of CROCS instead. To demonstrate the advantages of WSMD, we conducted a comparative evaluation against a broad set of alternatives using synthetic datasets.

In this case, the synthetic datasets contained 50 consumers with 90 days of data each, distributed among $K^*=2$ clusters, with cluster sizes again determined by a multinomial distribution with uniform probabilities. These two clusters were defined by distinct pairs of DLP shapes randomly sampled from the available 20. One hundred such datasets were generated for each $n_O \in \left\{0,5,10,\ldots,80\right\}$. 

\input{Table---SetDistances}

\begin{figure}[!p]
    \centering
    \begin{subfigure}{\textwidth}
        \centering
        \includegraphics[width=0.87\linewidth]{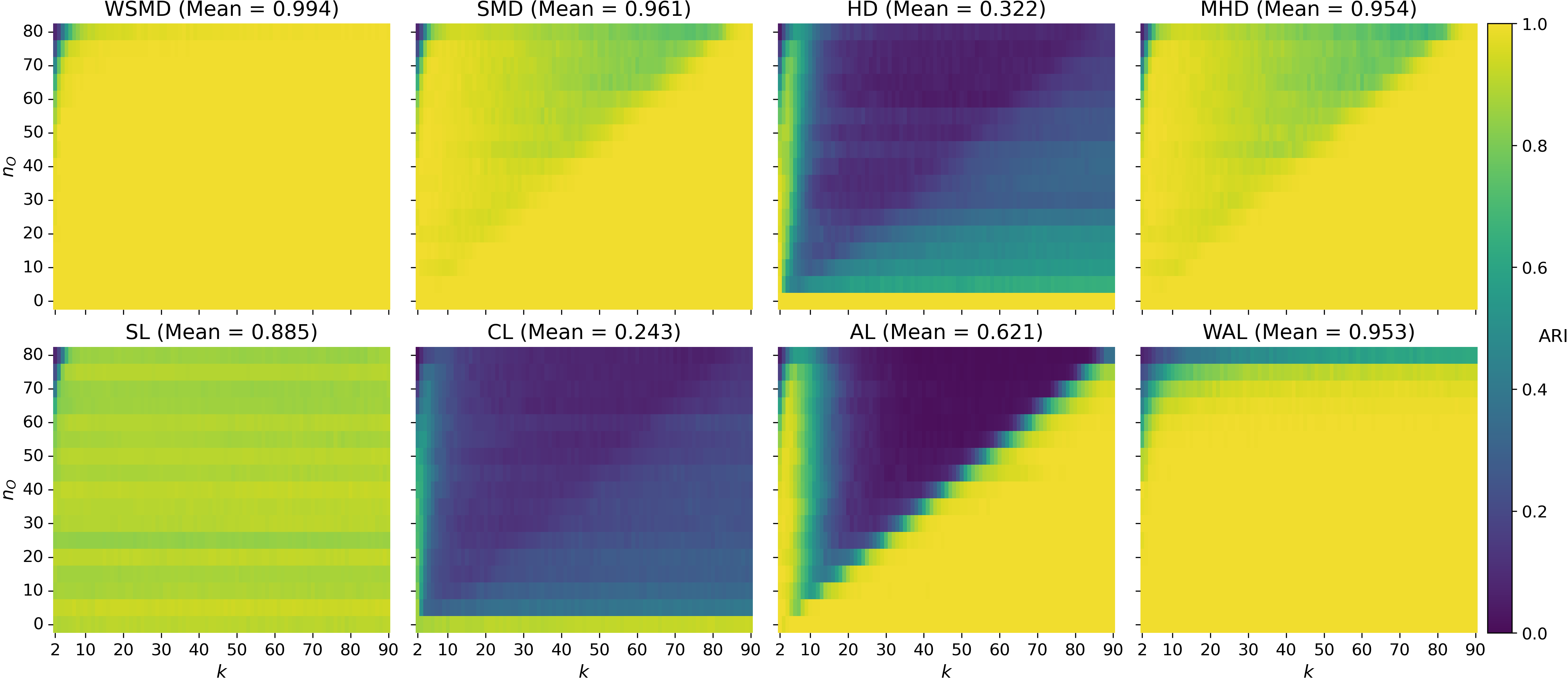}
        \caption{HAC-Wa}
        \label{Fig:ComparingSetDistances_HAC-Wa}
    \end{subfigure}
    
    \vspace{1em}
    
    \begin{subfigure}{\textwidth}
        \centering
        \includegraphics[width=0.87\linewidth]{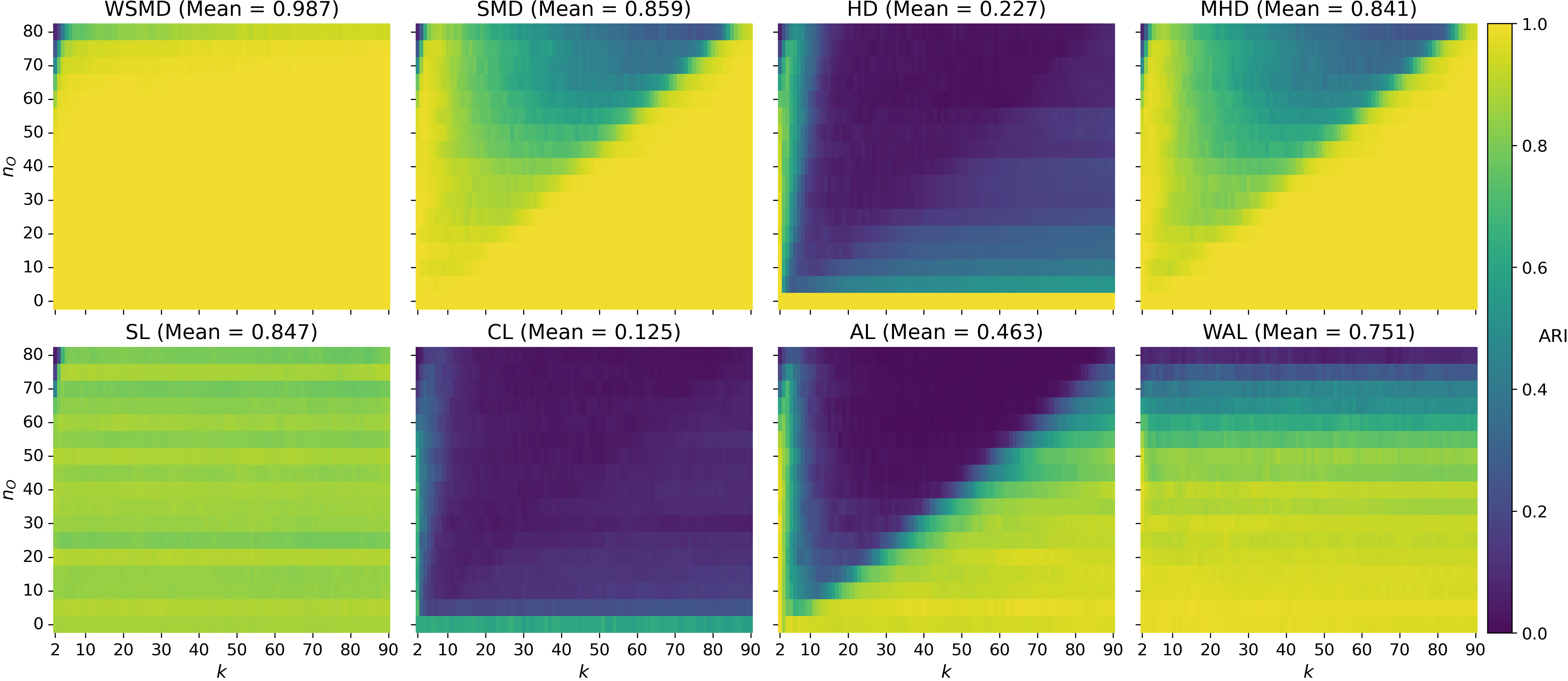}
        \caption{KMd}
        \label{Fig:ComparingSetDistances_KMd}
    \end{subfigure}
    \caption{Heatmaps of mean ARI values, indicating synthetic consumer cluster recovery when utilising different set distances to compute dissimilarity between consumer RLSs in stage two of the CROCS framework. In a) and b), the HAC-Wa and KMd clustering algorithms were used respectively. The average ARI across all numbers of outliers ($n_O$) and numbers of stage one clusters ($k$) is also provided.}
    \label{Fig:ComparingSetDistances}
\end{figure}

The consumers were clustered using CROCS with the number of stage one clusters systematically varied across $k\in\left\{2,3,\ldots,90\right\}$, using the default configuration. For the second stage of clustering, $K^*$ was assumed known, and 8 set distances (detailed in \Cref{Tab:SetDistances}) were employed to compare their suitability in recovering consumer clusters. Among these 8 are familiar set distances such as the Hausdorff distance \cite{Taha2015}, and others commonly used in agglomerative hierarchical clustering such as single, complete, and average linkage \cite{Aggarwal2014}. We also considered the original unweighted version of the WSMD (the sum of minimum distances), a version of the Hausdorff distance designed to emulate WSMD by replacing the first maximum operation with a mean, and a version of average linkage with pairwise distances weighted by the product of their respective cluster sizes.

The results from this experiment according to the ARI are presented in \Cref{Fig:ComparingSetDistances}, with HAC-Wa and KMd used as the clustering algorithm in \Cref{Fig:ComparingSetDistances_HAC-Wa,Fig:ComparingSetDistances_KMd} respectively. Note that these results were consistent with AMI and PSI. 

For both HAC-Wa and KMd, the proposed WSMD consistently obtained the largest ARI values out of all considered set distances, indicating better recovery of the consumer clusters. As recognised in \Cref{Sec:Results_CROCS_StageOne_Picking_k}, recovery of consumer cluster labels naturally becomes more difficult when more outliers are introduced. WSMD showed the greatest robustness to an increase in the number of outlier DLPs, only showing difficulties at extreme values, i.e. $n_O \geq 70$. Meanwhile, the closest competitor methods, SMD and MHD, were both affected by the increased proportion of outliers when overestimating $k$. As these set distances both place equal significance on all pairwise distances between RLS prototypes, when outliers are increasingly represented in the RLS, consumer similarity can be obscured. Furthermore, for SMD and MHD, perfect label recovery could be achieved for large values of $k$ that grow linearly with $n_O$, yet WSMD required the smallest overestimation to reliably discover the same consumer clusters. This robustness is desirable when the true $n_O$ is unknown in real datasets. This experiment provides assurance that WSMD is well-suited to computing pairwise dissimilarities between consumer RLSs.

\subsection{Identifying Similar Consumers on Asynchronous Schedules} \label{Sec:Results_Asynchronous}

As discussed in \Cref{Sec:LiteratureReview-Limitations(iii)}, many existing clustering approaches rely on temporally anchored comparisons, which can cause households with highly similar daily consumption patterns expressed on different days to be assigned to different clusters. This undermines the effectiveness of segmentation for downstream applications, where the similarity of underlying patterns often matters more than their alignment on the calendar. For example, two households with comparable routines but staggered work schedules may be equally well suited to the same tariff.

\begin{figure}[p]
    \centering
    \includegraphics[width=\linewidth]{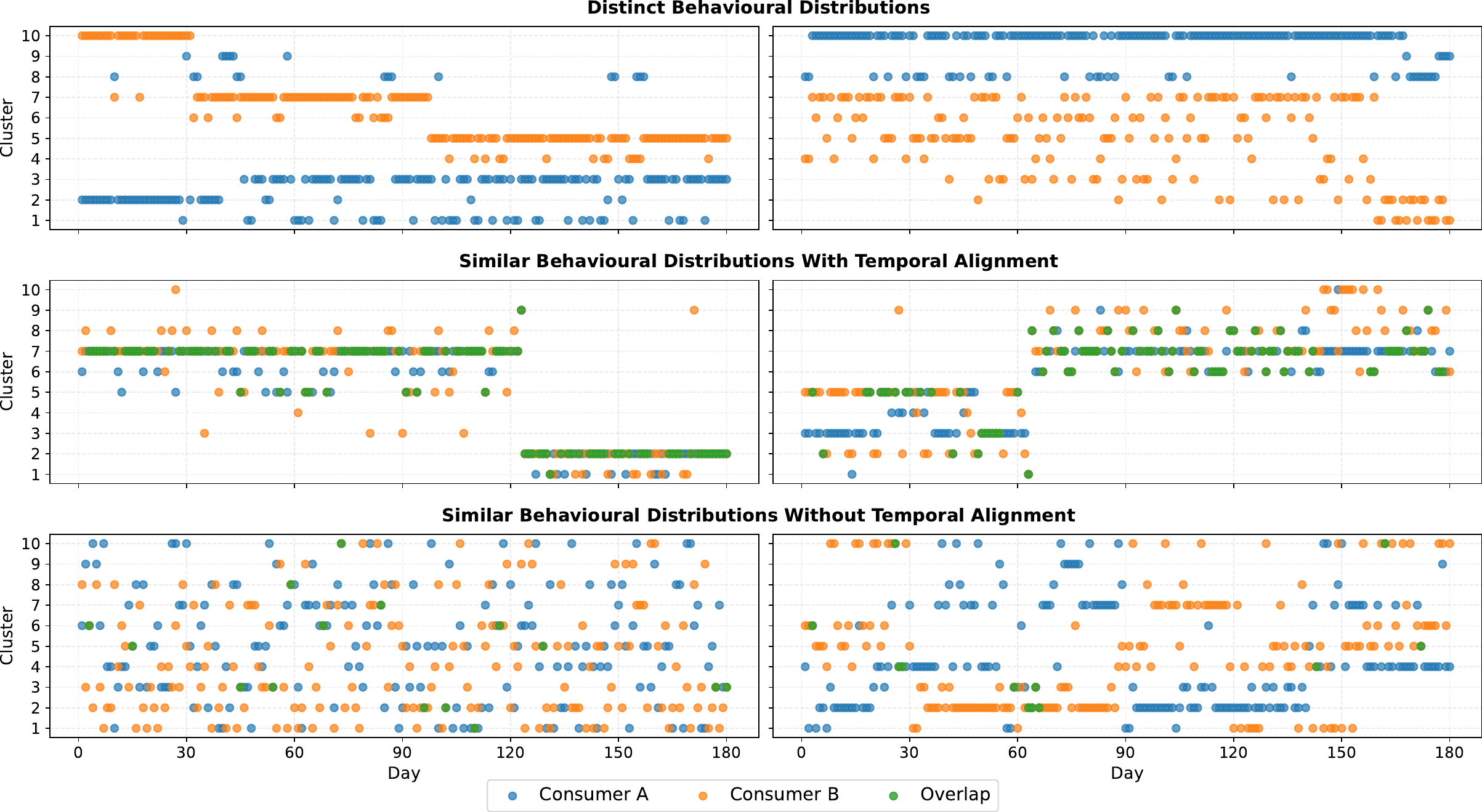}
    \caption{Temporal evolution of cluster membership for archetypal consumer pairs from the AG dataset. For each archetype, two example pairs (left and right) are shown to illustrate some within-category diversity. The 360 DLPs of these consumer pairs (from 6 months of consumption data) have been clustered into 10 clusters. Blue and orange markers indicate the respective cluster membership of the consumers' DLPs, with green indicating days with identical membership. The archetypal pairs include: (Top) Consumers with distinct behavioural distributions; (Middle) Consumers with similar behavioural distributions with temporal alignment; (Bottom) Consumers with similar behavioural distributions without temporal alignment.}
    \label{Fig:PairsExampleSchedules}
\end{figure}

To demonstrate different forms of consumer relationships present in real data, we examine archetypal pairs of consumers drawn from the AG dataset. In \Cref{Fig:PairsExampleSchedules}, each panel shows the daily cluster assignments of two consumers over a six-month period. Each dot corresponds to one consumer’s DLP, positioned by day and cluster label on the horizontal and vertical axes respectively. Blue and orange dots mark the cluster memberships of the individual consumers, and green dots highlight days when both consumers’ DLPs were assigned to the same cluster. The DLPs of the consumer pairs were clustered together into 10 clusters using min-max normalisation, DTW-2 and HAC-Wa. We provide \textit{two} examples for \textit{each} of three archetypal consumer relationships. In the top row are two consumer pairs whose DLPs did not share any clusters over the 6 month period, indicating fundamentally distinct consumption behaviours. In contrast, the middle and bottom rows show pairs of consumers whose DLPs were assigned to the same clusters, indicating shared consumption behaviours. In the middle row, these shared behaviours are temporally aligned, while in the bottom row, they occur asynchronously. Including two examples per archetype illustrates the diversity within each category, with differences in how similarity and alignment manifest. Ten clusters were sought to allow unique consumption behaviours to be adequately captured in their own clusters, while avoiding excessive fragmentation that could emphasise noise rather than meaningful behavioural differences.

\begin{figure}[p]
    \centering
    \includegraphics[width=0.66\linewidth]{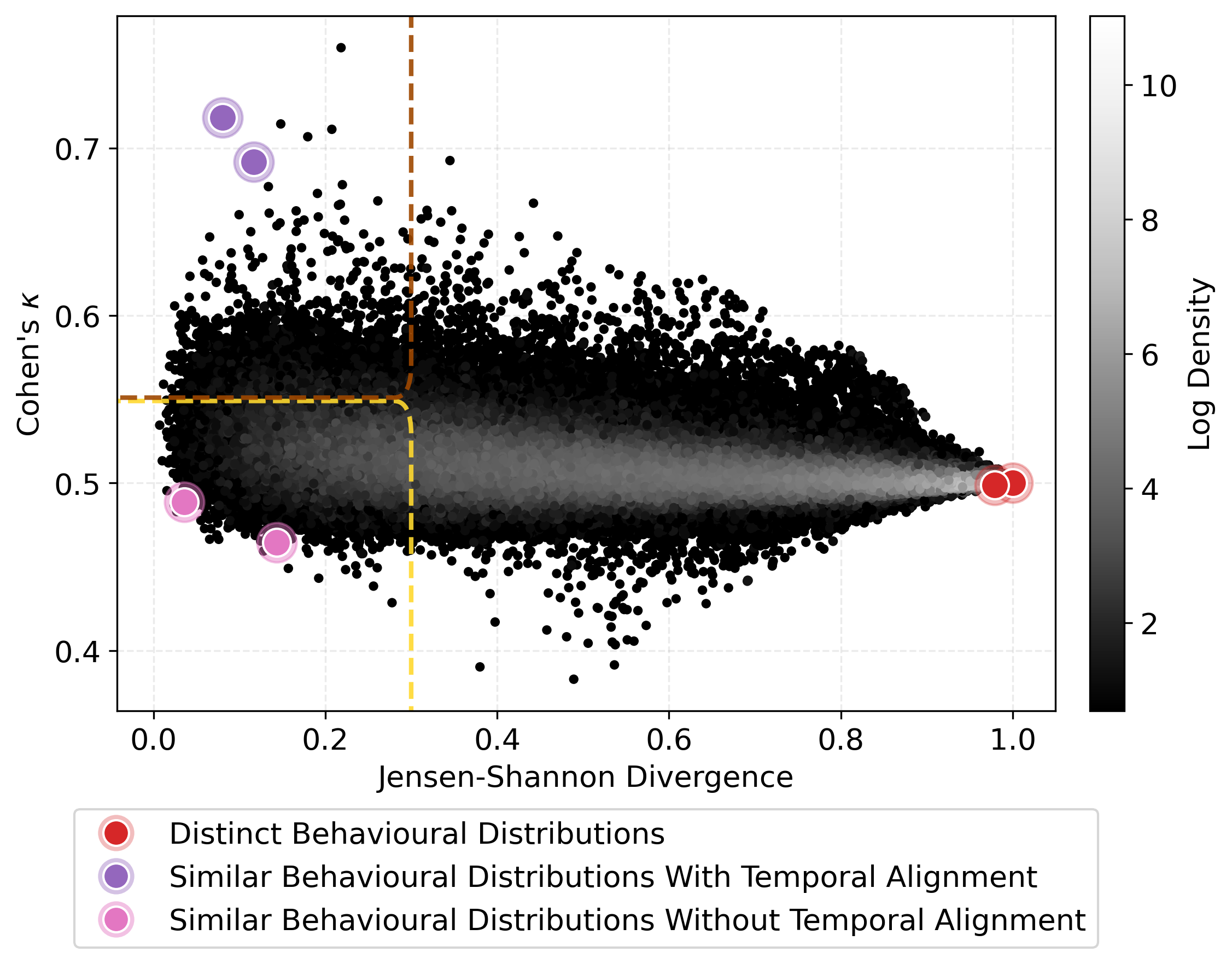}
    \caption{Log-density shaded scatterplot of JSD and Cohen's $\kappa$ computed for all pairs of AG consumers over 24 six-month periods starting in each month of 2011 and 2012. Each consumer pair's 360 DLPs were clustered together into 10 clusters. JSD and $\kappa$ were then computed between the resulting sequences. The bottom left region enclosed by the yellow dashed line ($\kappa < 0.55$, JSD $\leq 0.3$) indicates pairs of consumers that share similar behavioural distributions, with low temporal alignment. The upper left region enclosed by the brown dashed line ($\kappa > 0.55$, JSD $\leq 0.3$) indicates pairs that share similar distributions, with higher temporal alignment.}
    \label{Fig:JSD_vs_CK}
\end{figure}

The archetypal examples in \Cref{Fig:PairsExampleSchedules} define the extremes of consumer relationships, with most other consumers falling somewhere between these reference points. These relationships can be quantified using two complementary metrics. Firstly, similarity between behavioural distributions for a pair of consumers can be measured using the normalised Jensen-Shannon Divergence (JSD) \cite{61115} between cluster membership counts. The normalised JSD is symmetric, and bounded between 0 and 1 for discrete probability distributions, with \textit{lower values indicating greater overlap in consumer behaviours}. Secondly, the temporal alignment of those cluster sequences can be measured using normalised Cohen's $\kappa$ \cite{doi:10.1177/001316446002000104}, which accounts for the possibility of alignment due to chance. The normalised $\kappa$ ranges between 0 and 1, with values around 0.5 and lower indicating agreement by chance alone, and values approaching 1 indicating perfect agreement. 

Consumers with similar behavioural distributions that are not temporally aligned will have cluster sequences that produce low values of JSD, and $\kappa$ around 0.5. To understand how common these consumers are in a real-world dataset, we computed these two metrics for all consumer pairs in the AG dataset across multiple time periods. \Cref{Fig:JSD_vs_CK} provides a density-shaded scatterplot of JSD and $\kappa$ for all $\binom{300}{2}$ consumer pairs across 24 six-month periods starting in each month of 2011 and 2012. The six consumer pairs in \Cref{Fig:PairsExampleSchedules} are located within the plot with colour-coded markers for reference. 

The densest region of the scatterplot in \Cref{Fig:JSD_vs_CK} is at $(1.0,0.5)$, where consumer pairs have distinct behavioural distributions and, resultingly, no temporal alignment of their consumption behaviours. This is expected because each consumer can be paired with far more households that do not share its behaviours than those that do. However, a notable proportion of consumer pairs fall into the bottom left region of the scatterplot, bordered by the yellow dashed line, where they exhibit substantial overlap in their consumption behaviours but at different times. For indicative purposes, around $7.7\%$ of pairs fall into this region ($\kappa < 0.55$, JSD $\leq 0.3$), compared with only $0.6\%$ in the brown region ($\kappa > 0.55$, JSD $\leq 0.3$) where similar behaviours are more strongly aligned. The consumer pairs shown in the middle row of \Cref{Fig:PairsExampleSchedules} with temporal alignment (purple markers in the upper left of \Cref{Fig:JSD_vs_CK}) are at the extreme upper end of $\kappa$'s range, with $\kappa \approx 0.7$. The majority of consumers however have $\kappa$ values well below this (typically $<0.6$), indicating that strong behavioural \textit{and} temporal alignment is exceedingly rare. While the specific percentages of pairs in these regions depend on the chosen thresholds, the large disparity indicated here illustrates that asynchronous similarity is likely much more common than temporally aligned similarity. As a result, many existing approaches that rely on calendar synchronisation are likely to miss a substantial share of functionally similar consumers.

\subsubsection{Comparative Analysis} \label{Sec:Results_Asynchronous_ComparativeAnalysis}

Having demonstrated that consumers with asynchronous similarity are more abundant in real-world data than those with synchronised similar behaviours, we next evaluated how well existing clustering methods could identify such asynchronous similarity. To enable a controlled comparison, we conducted a synthetic data experiment where the ground truth consumer relationships were known. We begin by describing the process used to generate the synthetic data for this experiment.

As in previous synthetic experiments, clusters of consumers shared the same $k^*$ synthetic DLP shapes. However, to create more realistic consumption time series for methods that compute features from or operate directly on long time series, we determined each consumer's sequence of shapes by generating random sequences from first-order Markov models fitted to real-world consumer shape sequences. First-order Markov models have previously been used to characterise the daily evolution of consumer load profiles \cite{Teeraratkul2018}, and we validated their appropriateness for our data using Csiszar's $\phi$-divergences for testing Markov process order ($r$) \cite{Menendez2001}\footnote{This procedure estimates the order of dependence in a discrete-state sequence, but does not assess the validity of the Markov chain assumption itself, including properties such as the homogeneity of transition probabilities over time.}.

\Cref{Tab:Csiszars} shows the proportion of AG and SGSC consumers with cluster label sequences consistent with first-order dependence for $k \in \left \{2,3,\ldots,50\right\}$. DLP clustering for each consumer was performed using DTW-2 on min-max normalised profiles with either HAC-Wa or KMd. Following \cite{Menendez2001}, order testing was performed sequentially beginning at $r=5$, testing the null hypothesis of an $(r-1)\textsuperscript{th}$-order process against the alternative of an $r\textsuperscript{th}$-order process, and proceeding to successively lower orders until the null hypothesis was rejected. Sequences were classified according to the order at which rejection occurred, with $r=1$ corresponding to first-order dependence and failure to reject at $r=1$ indicating a random sequence of independent observations.

We analysed the 300 AG consumer sequences with the full 1096 days each, spanning 01/07/2010 to 30/06/2013\footnote{Using the dd/mm/yyyy format.}, and the 3953 SGSC sequences with the full 365 days each, covering all of 2013. The vast majority of consumers were well-fitted by a first-order Markov process. Even with highly conservative Bonferroni corrections accounting for all 49 values of $k$ and all consumers, the proportions remained high: 0.989 (HAC-Wa) and 0.986 (KMd) for AG consumers, and 0.737 (HAC-Wa) and 0.676 (KMd) for SGSC consumers.

Since the AG dataset provided more historical data for reliable estimation of Transition Probability Matrices (TPMs), we used it as the basis for synthetic data generation, restricted to those consumers whose cluster sequences exhibited first-order Markovian behaviour across all $k \in \left \{2,3,\ldots,50\right\}$ for both HAC-Wa and KMd. To generate synthetic data for a consumer cluster with $k^*$ distinct consumption patterns, we randomly selected one TPM estimated from the real-world AG clustering results with $k=k^*$ clusters. All synthetic consumers within the same cluster shared both the same $k^*$ synthetic consumption patterns and the same TPM, but each consumer's pattern sequence was independently sampled from this TPM. This approach encourages the synthetic time series to reflect a realistic evolution of daily patterns for the target number of clusters, while maintaining behavioural cohesion within consumer clusters.

\begin{table}[!h]
    \tiny
    \begin{adjustbox}{center}
        \begin{tabular}{p{3cm} p{1cm} p{1cm}} \toprule
            \multicolumn{1}{l}{} & \multicolumn{2}{c}{\textbf{Dataset}} \\ \cmidrule(l){2-3}
            \textbf{Clustering Algorithm} & \textbf{AG} & \textbf{SGSC}\\ \midrule
            
            HAC-Wa & 0.999 & 0.948 \\[0.1cm] 
            KMd & 0.996 & 0.926 \\[0.1cm] 
                                         
            \bottomrule 
        \end{tabular}
    \end{adjustbox}
    \caption[Csiszar's order test proportions for AG and SGSC datasets]{Proportion (to 3 significant figures) of consumer cluster membership sequences consistent with a first-order Markov process according to Csiszar's order test at the 0.05 significance level. Sequences were obtained for consumers individually with DTW-2 and either HAC-Wa or KMd clustering algorithms on the AG and SGSC datasets.}
    \label{Tab:Csiszars}
\end{table}

\begin{figure}[bhtp]
    \centering
    \includegraphics[width=\linewidth]{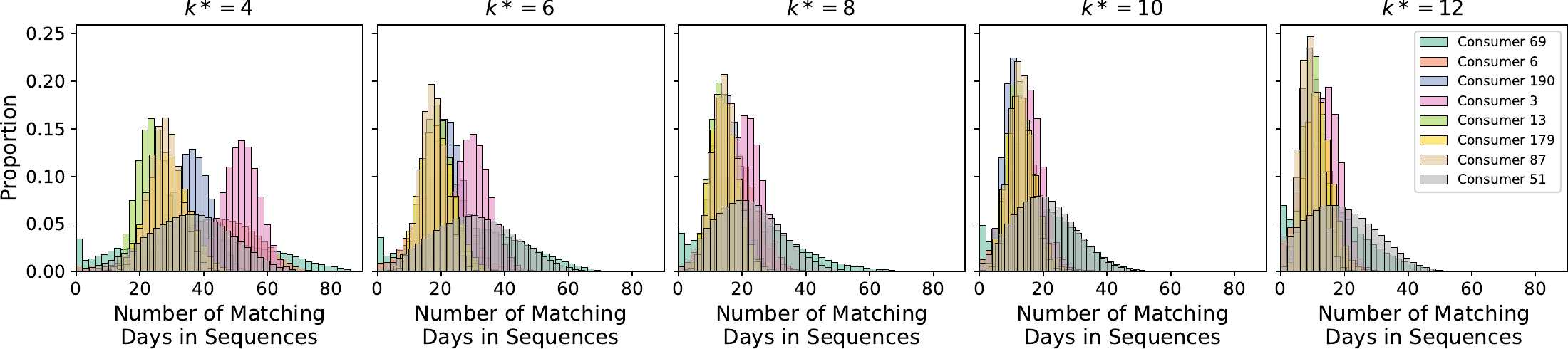}
    \caption{Histograms showing the proportion of days with identical cluster membership for consumer pairs within the same synthetic cluster. Sequences were generated for 8 random TPMs trained on real AG consumer cluster label sequences that were validated as first-order Markovian by Csiszar's test for all $k \in\left\{4,6,8,10,12\right\}$. Each histogram is based on 1000 synthetic sequences, spanning 90 days generated from each consumer's TPM.}
    \label{Fig:ProportionOfAlignedDays}
\end{figure}

The resulting temporal alignment of consumer pattern sequences from this generation process are shown in \Cref{Fig:ProportionOfAlignedDays}, where histograms record the proportion of days with identical cluster membership among 1000 90-day sequences sampled from 8 different TPMs trained on consumers from the AG dataset. The key takeaway is that for smaller values of $k^*$, synthetic consumers in the same cluster could have low or high temporal alignment, while for larger values, lower temporal alignment is the norm. 

The synthetic datasets used for this experiment were generated with 300 consumers, split into $K^*=8$ clusters, with $k^*\in\left\{4,6,8,10,12\right\}$ distinct synthetic DLP shapes. A total of $n_O \in \left\{0,20,40,60\right\}$ DLPs from the 90 day sequences were replaced with outlier DLPs. Furthermore, $N_O \in\left\{0,50,100\right\}$ consumers from the 300 were designated as outlier consumers. These outlier consumers had their own distinct set of $k^*$ DLP shapes, and their own distinct TPM for generating shape sequences. These outlying consumers were intended to obscure the otherwise well-separated consumer clusters, but recognition of their outlier status was not pertinent to our experiment. As such, when computing the EVIs, the labels of these consumers were ignored. Ten datasets were generated for each combination of settings.

\begin{figure}[!b]
    \centering
    \begin{subfigure}{\textwidth}
        \centering
        \includegraphics[width=\linewidth]{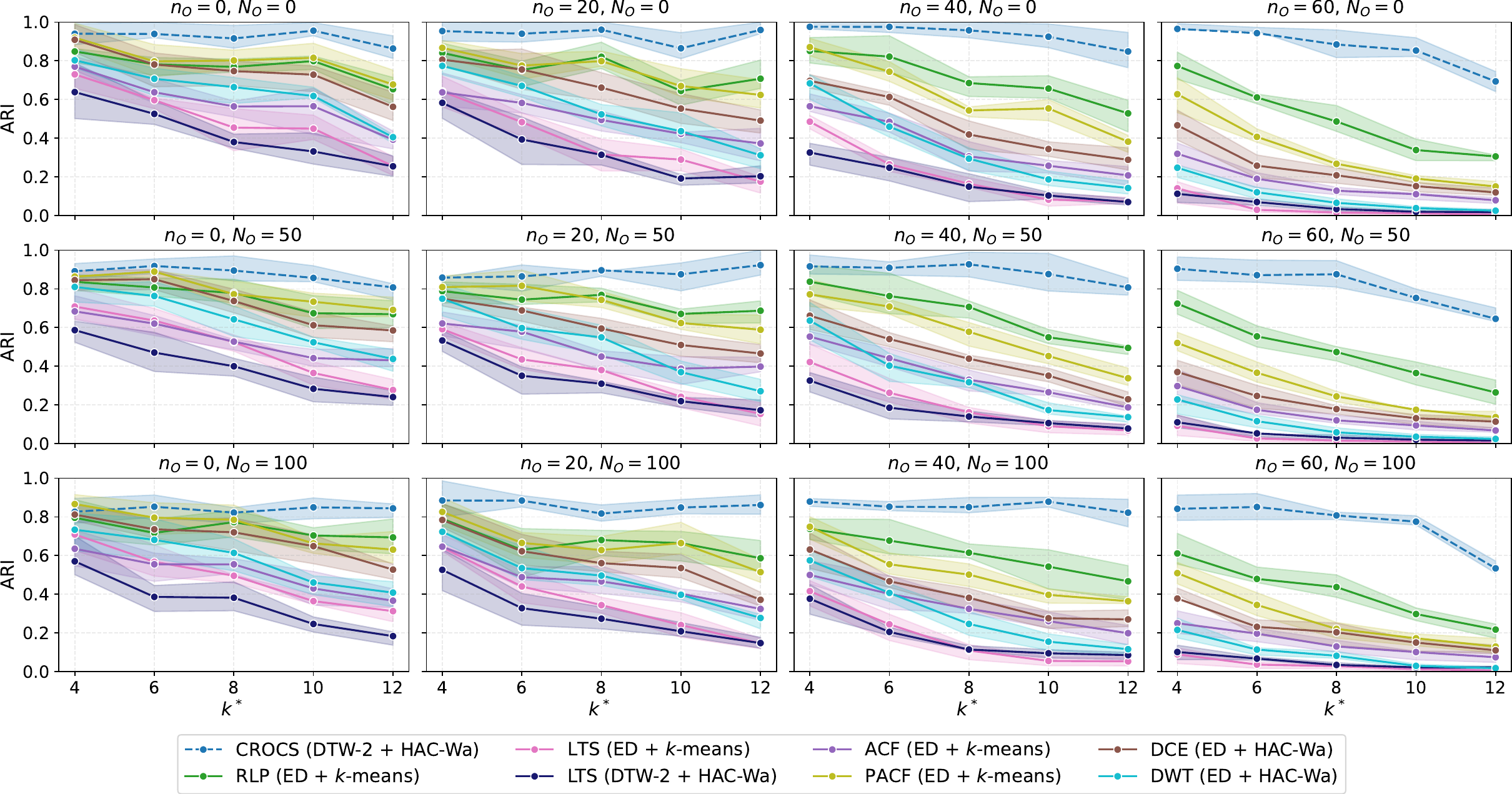}
        \caption{Single-Stage Methods vs CROCS}
        \label{Fig:SyntheticData_OneStage}
    \end{subfigure}
    
    \vspace{1em}
    
    \begin{subfigure}{\textwidth}
        \centering
        \includegraphics[width=\linewidth]{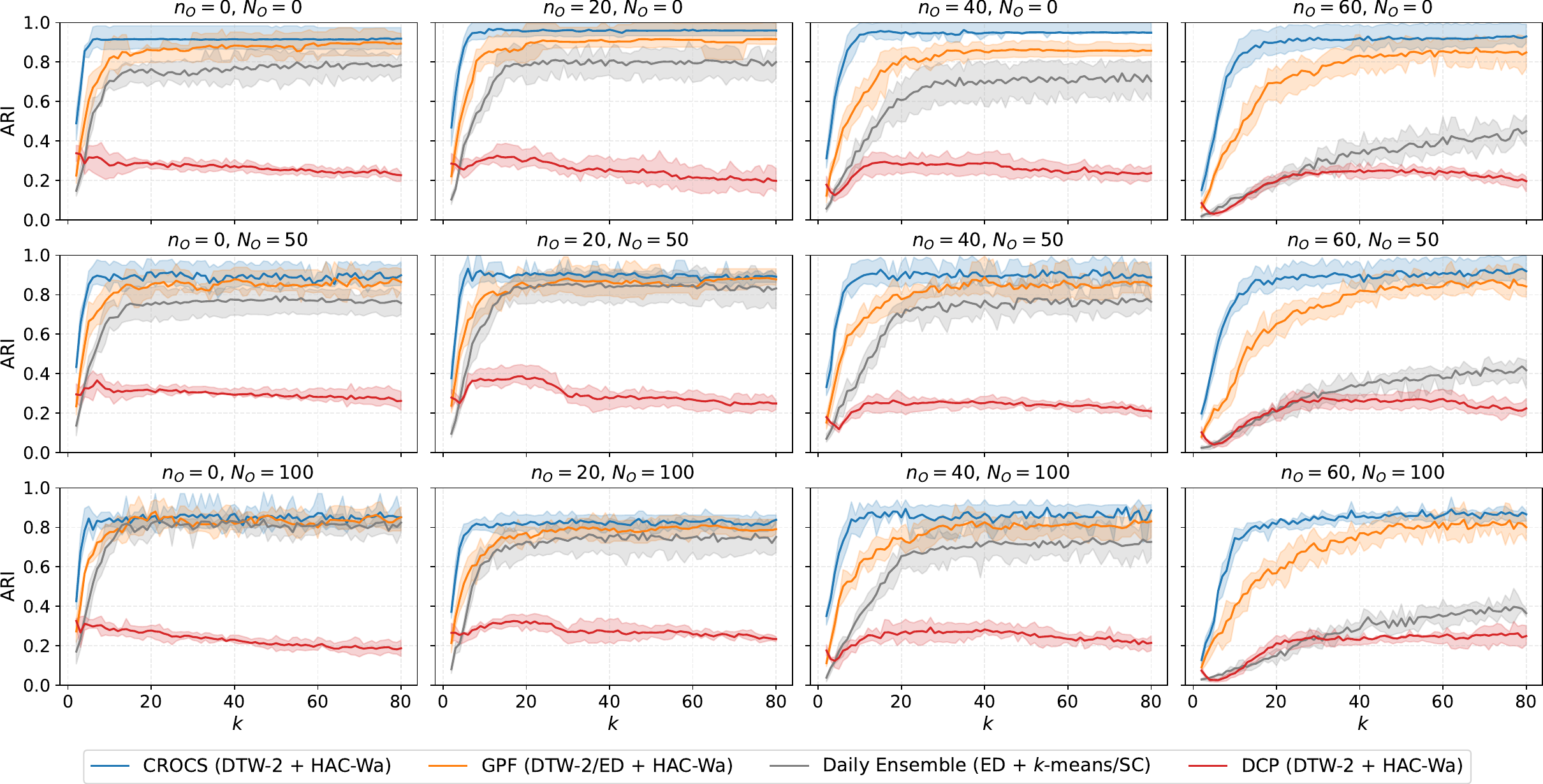}
        \caption{Two-Stage Methods vs CROCS}
        \label{Fig:SyntheticData_TwoStage}
    \end{subfigure}
    \caption{Mean ARI (with 50 percentile intervals) indicating recovery of synthetic consumer clusters by CROCS and other one and two-stage clustering methods proposed in the literature. Note that the results shown in (b) are for $k^* = 8$, though similar plots for the other values of $k^*$ provided consistent results.}
    \label{Fig:SyntheticData}
\end{figure}

To benchmark the capacity of CROCS to discover asynchronous consumer similarity, we selected a broad range of representative methods from the consumer clustering literature. These methods fall into two main categories: single-stage clustering approaches and two-stage clustering approaches. The single-stage methods include the following representations: RLPs, normalised Long Time Series (LTS), Autocorrelation (ACF) and Partial Autocorrelation features (PACF), Delay Coordinate Embedding (DCE) map parameters as in \cite{Motlagh2019}, and Discrete Wavelet Transform (DWT) coefficients. Each representation was applied to the data using multiple clustering algorithms, and with multiple parameter settings where appropriate. All representations were paired with $k$-means, HAC-Wa and KMd using ED. RLPs and LTS were additionally applied using DTW-2. As for \cite{Alonso2020HierarchicalAutocovariances}, the ACF and PACF were computed using 96 lags. DCE map parameters were found for embedding dimensions $m$, and delay $\tau \in \left\{5,10,15,20,25,30\right\}$. Furthermore, instead of applying a linear neural regression to obtain the map parameters, the analytical solution was used. Finally, for DWT we considered the Haar and Coiflet-8 mother wavelets (as in \cite{Tureczek2018}) at all possible levels of decomposition, using either all coefficients, just the approximation coefficients, or the last pair of detail and approximation coefficients. For a best-case comparison with CROCS, the combination of parameters, clustering algorithm and distance measure that produced the highest average ARI with each representation is the version for which results are presented.

Meanwhile, the two-stage methods include the following: GPFs and DCPs with DTW-2 and either HAC-Wa or KMd, and the Daily Ensemble method as implemented by Sun et al. \cite{Sun2020}, where $k$-means was applied independently on each day's DLPs in the first stage followed by spectral clustering of a similarity matrix indicating how frequently households were grouped together in the second stage. These methods are united with CROCS by the parameter $k$, that is, the number of clusters used in the first stage of clustering. For all methods, we clustered the datasets with $k \in \left \{2,3,\ldots,80\right\}$. \Cref{Fig:SyntheticData} summarises the results according to the ARI from these experiments, with single- and two-stage methods compared with CROCS in \Cref{Fig:SyntheticData_OneStage} and \Cref{Fig:SyntheticData_TwoStage} respectively.

In \Cref{Fig:SyntheticData_OneStage}, the mean ARI values are plotted with a 50 percentile interval for each $k^*$, with columns containing results for datasets with different numbers of DLP outliers ($n_O$), and rows containing results for datasets with different numbers of outlying consumers ($N_O$). The performance of CROCS when $k=15$ is shown with a dashed line for comparison. CROCS was found more capable of recognising asynchronous similarity than all of the other single-stage methods, regardless of $k^*$ or the number of each kind of outlier. CROCS is also the least affected by $k^*$, with most other methods showing a more precipitous drop in mean ARI with increasing $k^*$. This collective drop in performance is likely related to the previous observation from \Cref{Fig:ProportionOfAlignedDays} that as $k^*$ increases, our synthetic consumers show much lower temporal alignment of their common consumption patterns. Interestingly, the simple RLP appears to be the next best method after CROCS, likely due also to its independence from the ordering of consumer DLPs. Clustering approaches that use the normalised LTS performed the worst --- especially for the largest values of $n_O$--- likely due to the combination of asynchronous similarity and the curse of dimensionality.

Due to sharing the important and comparable parameter $k$, for the two-stage methods we have instead provided results for $k^*=8$ in \Cref{Fig:SyntheticData_TwoStage}, with the $x$-axis instead showing variation in performance for different values of $k$. However, plots for $k^* \in \left\{4,6,10,12\right\}$ provided consistent observations. The rows and columns still mirror those of \Cref{Fig:SyntheticData_OneStage}. This time we can see that there is less separation between CROCS and the next best performer in GPF, especially for lower $n_O$. However, as $n_O$ increases, the GPF representation requires much larger values of $k$, relative to CROCS, to account for the additional outlying shapes in the datasets. Meanwhile, the daily ensemble method also struggles considerably with increasing values of $n_O$ as consumers exhibit shapes on more days that are not aligned with the core shapes of their ground truth cluster. The DCP performs poorly in all scenarios, as consumers' most common consumption profiles will not always be shared with the other consumers in their cluster due to the random sampling of pattern sequences from their shared TPMs.

These results demonstrate that CROCS consistently outperforms both single-stage and two-stage clustering methods when identifying consumers with similar but asynchronously expressed consumption patterns, with the performance gap widening as real-world complexity increases.

\subsection{Scalability} \label{Sec:Results_Scalability}

The practical value of a consumer clustering method depends not only on segmentation quality, but also on its ability to process datasets at real-world scales. To evaluate the computational performance of CROCS, we applied it to synthetic datasets comprising tens of thousands of consumers over a three-month (90 day) period. CROCS was implemented using DTW-2 with HAC-Wa and min-max normalisation. As discussed in \Cref{Sec:Results_CROCS_StageOne_Picking_k}, stage one runtime with HAC-Wa is consistently faster than with KMd and independent of the choice of $k$, while stage two runtime is similarly independent of the number of consumer clusters ($K$).

Stage one clustering and the pairwise consumer distance matrix computation with WSMD were parallelised across 50 cores on a single node of a Linux HPC system equipped with AMD EPYC 7543 processors (2.8 GHz base clock). Jobs were executed in Python 3.12.3 with up to 256 GB memory (as required for the dataset size). Runtimes reported are wall-clock times for the first and second stages of clustering, inclusive of parallelisation overhead.

Recall that the computational complexity of CROCS is $\bigO{mp^2 + k^2m^2}$, where $m$ is the number of consumers, $p$ the number of days, and $k$ the number of stage one clusters. Despite quadratic dependence on both $m$ and $p$, in practice $m \gg p$ is by far the more relevant case, as historical data more than a few years old is often of limited current segmentation value, while consumer populations can easily reach massive scales. Accordingly, we generated datasets with $m\in \left\{5000, 10\,000, 20\,000, 40\,000, 80\,000 \right \}$. We also considered practical numbers of stage one clusters $k\in\left\{2,3,\ldots,30\right\}$ and numbers of consumer clusters $K\in \left\{2,3,\ldots,20\right\}$.

\begin{figure}[!b]
    \centering
    \includegraphics[width=0.52\linewidth]{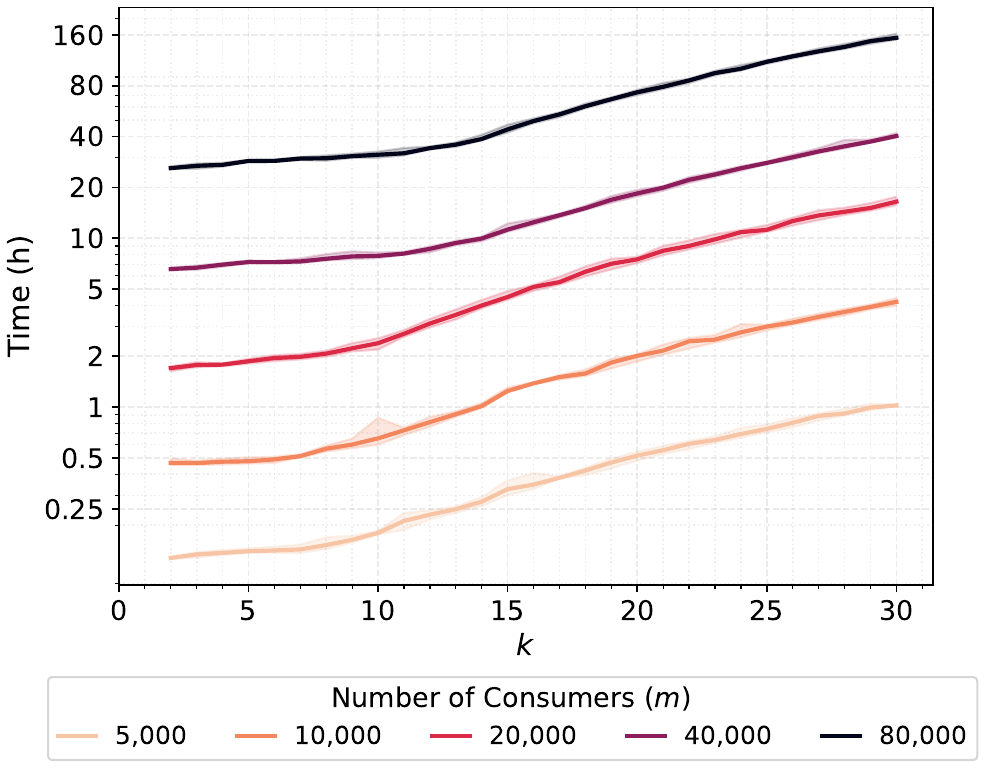}
    \caption{Execution time of CROCS (total wall-clock runtime in hours averaged over ten runs, inclusive of parallelisation overhead) when applied to datasets with varying numbers of consumers ($m$) with $p=90$ DLPs per consumer. The data were clustered using DTW-2 with HAC-Wa and min-max normalisation, and results are shown on a log-scale for different numbers of stage one clusters ($k$).}
    \label{Fig:ScalabilityPlot}
\end{figure}

Execution times for CROCS across the different parameter settings are reported in \Cref{Fig:ScalabilityPlot}. As expected, the quadratic dependence on $m$ is evident: doubling the number of consumers consistently leads to about a fourfold increase in runtime, with a comparable effect observed when increasing $k$ over the practical range considered. For reference, datasets of 10,000 consumers with $k=15$ completed in under two hours, while datasets of 80,000 consumers completed within two days. These datasets are far beyond the scales available in public datasets used in most prior smart meter clustering studies --- typically under 7,000 consumers \cite{Wang2019}.

The results here show that CROCS maintains practical runtimes while preserving the flexibility needed to achieve meaningful and operationally useful clustering outcomes. Although our implementation was parallelised across 50 HPC cores, there remains scope for further optimisation, though future improvements may involve trade-offs between speed and segmentation quality, particularly for applications at even larger scales or when faster turnaround is required. Notably, we did not employ any of the available faster variants of clustering or distance computation --- such as accelerated or approximate DTW variants \cite{10.1145/2783258.2783286}, and parallel or approximate hierarchical clustering \cite{10.1007/978-3-540-27866-5_47,Kull2008} --- suggesting that further efficiency gains are possible should such methods be incorporated. In addition, many practical applications could suitably be addressed by analysing representative samples rather than entire populations, making CROCS readily applicable at operationally relevant scales.

%% file: Table---SetDistances.tex
\begin{table}[!p]
    \scriptsize
    \rowcolors{1}{TableGray}{TableWhite}
    \begin{adjustbox}{center}
        \begin{tabular}{p{4.8cm} p{1.2cm} p{9.5cm}} \toprule \hiderowcolors 
            \textbf{Set Distance} & \textbf{Acronym} & \textbf{Equation}\\ \midrule \showrowcolors
            
            Weighted Sum of Minimum Distances & WSMD & $\Delta_{\text{WSMD}} \left( \hat{\mathcal{S}}_i, \hat{\mathcal{S}}_j \right) = \frac12 \left[\frac{1}{p_i}\sum \limits_{a=1}^{k_i} n_i^a \min \limits_{b} d\left( \bm{\pi}_i^a, \bm{\pi}_j^b \right) +  \frac{1}{p_j}\sum \limits_{b=1}^{k_j} n_j^b \min \limits_{a} d\left( \bm{\pi}_j^b, \bm{\pi}_i^a \right)\right]$ \\[0.25cm] 

            Sum of Minimum Distances & SMD & $\Delta_{\text{SMD}} \left( \hat{\mathcal{S}}_i, \hat{\mathcal{S}}_j \right) = \frac12 \left[\sum \limits_{a=1}^{k_i} \min \limits_{b} d\left( \bm{\pi}_i^a, \bm{\pi}_j^b \right) +  \sum \limits_{b=1}^{k_j} \min \limits_{a} d\left( \bm{\pi}_j^b, \bm{\pi}_i^a \right)\right]$ \\[0.25cm] 

            Hausdorff Distance & HD & $\Delta_{\text{HD}} \left( \hat{\mathcal{S}}_i, \hat{\mathcal{S}}_j \right) = \max \left\{ \max \limits_{a} \min \limits_{b} d\left( \bm{\pi}_i^a, \bm{\pi}_j^b \right), \max \limits_{b} \min \limits_{a} d\left( \bm{\pi}_j^b, \bm{\pi}_i^a \right) \right\}$ \\[0.25cm] 

            Modified Hausdorff Distance & MHD & $\Delta_{\text{MHD}} \left( \hat{\mathcal{S}}_i, \hat{\mathcal{S}}_j \right) = \max \left\{ \frac{1}{k_i} \sum \limits_{a=1}^{k_i} \min \limits_{b} d\left( \bm{\pi}_i^a, \bm{\pi}_j^b \right), \frac{1}{k_j} \sum \limits_{b=1}^{k_j} \min \limits_{a} d\left( \bm{\pi}_j^b, \bm{\pi}_i^a \right) \right\}$ \\[0.25cm] 

            Single Linkage & SL & $\Delta_{\text{SL}} \left( \hat{\mathcal{S}}_i, \hat{\mathcal{S}}_j \right) = \min \limits_{a} \min \limits_{b} d\left( \bm{\pi}_i^a, \bm{\pi}_j^b \right)$ \\[0.25cm] 

            Complete Linkage & CL & $\Delta_{\text{CL}} \left( \hat{\mathcal{S}}_i, \hat{\mathcal{S}}_j \right) = \max \limits_{a} \max \limits_{b} d\left( \bm{\pi}_i^a, \bm{\pi}_j^b \right)$ \\[0.25cm] 

            Average Linkage & AL & $\Delta_{\text{AL}} \left( \hat{\mathcal{S}}_i, \hat{\mathcal{S}}_j \right) = \frac{1}{k_i k_j} \sum \limits_{a=1}^{k_i} \sum \limits_{b=1}^{k_j} d\left( \bm{\pi}_i^a, \bm{\pi}_j^b \right)$ \\[0.25cm] 

            Weighted Average Linkage & WAL & $\Delta_{\text{WAL}} \left( \hat{\mathcal{S}}_i, \hat{\mathcal{S}}_j \right) = \frac{1}{p_ip_j}\sum \limits_{a=1}^{k_i} \sum \limits_{b=1}^{k_j} n_i^a n_j^b \, d\left( \bm{\pi}_i^a, \bm{\pi}_j^b \right)$ \\[0.25cm] 
                                         
            \bottomrule \hiderowcolors
        \end{tabular}
    \end{adjustbox}
    \caption{Set distances compared for stage two of the CROCS framework. Note that $\mathcal{S}_i$ is the $i\textsuperscript{th}$ consumer's RLS, which contains $k_i$ prototypical DLPs. The $a\textsuperscript{th}$ such prototype is $\bm{\pi}_i^a$, which represents $n_i^a$ of the consumers $p_i$ total DLPs. Furthermore, note that $d\left( \cdot, \cdot \right)$ is a time series dissimilarity measure.}
    \label{Tab:SetDistances}
\end{table}

%% file: Application.tex
\section{Real-World Application} \label{Sec:Application}

We now demonstrate an application of the CROCS framework to the real-world AG dataset, examining consumer clusters across both Australian winter and summer periods. The winter period spanned from 1\textsuperscript{st} June to 31\textsuperscript{st} August 2012 inclusive, and summer from 1\textsuperscript{st} December 2012 to 28\textsuperscript{th} February 2013 inclusive. As discussed in \Cref{Sec:LiteratureReview-Limitations(vi)}, many energy retailers make a distinction between workdays and non-working\footnote{Incorporating both weekends and public holidays.} days due to the significance of coincident peak usage with commercial consumers. For this reason, only workdays within these windows were included in the analysis --- leaving 65 and 60 days during the winter and summer periods respectively. By excluding non-working days, this application aligns with realistic use cases such as the design of weekday time-of-use tariffs or the targeting of other DR programs that aim to reduce significant weekday consumption peaks.

For this scenario, CROCS was implemented with min-max normalised DLPs, DTW-2, medoid prototypes and KMd for both clustering stages, though HAC-Wa was also considered. The number of stage one clusters ($k$) and consumer clusters ($K$) were varied, with selections guided by consideration of RVIs and the judgement of domain experts. In both cases, $k=15$ was found to sufficiently capture the range of consumers' DLP shapes, and $K=5$ provided a partition of the consumers into a practical number of distinct clusters for the hypothetical use cases. 

Having introduced the concept of Refined RLSs in \Cref{Sec:TheFramework_RRLS}, we now compute them for the consumer clusters obtained in this application. The RRLSs provide hyperprototypes that summarise the common diurnal load patterns shared by consumers within each cluster, enhancing interpretability by making explicit the behavioural modes that define cluster membership.

As noted in \Cref{Sec:TheFramework_RRLS}, we do not prescribe any particular community detection method for identifying densely connected regions of the prototype graph. In this application, we employed the Leiden algorithm for its scalability and guarantee of well-connected communities \cite{Traag2019a}. In particular, the \texttt{leidenalg} Python package implementation was used with Reichardt--Bornholdt Erd\H{o}s--R\'enyi (RBER) quality function \cite{PhysRevE.74.016110}, which incorporates vertex weights, and edge weights (obtained as the inverse of DTW-2 dissimilarities). Various quality functions are available in \texttt{leidenalg} which exploit different graph attributes --- for instance, RBER and the Constant Potts Model (CPM) \cite{Traag2011} make use of vertex weights and edge weights, while traditional modularity considers edge weights and directions. The use of vertex weights is particularly relevant in this application, as each node (prototype) already represents multiple DLPs, and the optimiser accounts for this when defining the size of communities in the quality function.

The RBER quality function was selected over both CPM and modularity for a few reasons. Like CPM, RBER includes a resolution parameter $\gamma$ that enables the identification of both small and large communities, in contrast to modularity which struggles to resolve small but meaningful communities \cite{Traag2011}. However, CPM imposes an absolute density cut-off, requiring all communities exceed the fixed threshold $\gamma$ irrespective of the graph’s overall structure. RBER instead defines community density relative to the global edge density $\rho$, requiring communities be denser than $\gamma\rho$. This relative formulation means that a given value of $\gamma$ carries a more consistent interpretation across prototype graphs of varying size and density --- in contrast to CPM, where an absolute threshold tuned for one graph may be orders of magnitude too large or too small for another, requiring re-tuning from scratch. In any case, larger values of $\gamma$ impose stricter density requirements, typically producing a greater number of smaller communities. Importantly, recognising minor communities of outlying prototypes separates them from the major communities, resulting in more meaningful coverage statistics.

Similar to the tuning of $\gamma$ in \cite{Watson2022}, any desired number of communities was obtained by initialising $\gamma$ at 0.02 and incrementing it in steps of 0.01 until the target number of communities was achieved. The hyperprototypes from a range of numbers of communities were evaluated by domain experts to ensure representativeness, striking a balance between concisely summarising typical daily load patterns and preserving operationally meaningful behavioural diversity present in each community.

The resulting summer and winter consumer partitions are presented in \Cref{Fig:Application-SUMMER,Fig:Application-WINTER} respectively, with each cluster's corresponding medoid and mean hyperprototypes shown in the two right hand panels. The left-hand panel shows the communities discovered in the respective prototype graphs, while the second panel from the left reports coverage statistics in terms of consumer coverage ($x$-axis) and day coverage ($y$-axis), both expressed as percentages. A relatively large number of hyperprototypes (10-40) were selected for each consumer cluster, many of which represent small communities with outlying consumption patterns not widely shared across the cluster. The major hyperprototypes that captured consumption patterns common to a large proportion of consumers are prominently represented with distinct colours in \Cref{Fig:Application-SUMMER,Fig:Application-WINTER}, while outlying patterns are shown in a uniform navy blue tone.

The following sections examine the summer and winter partitions in detail, focusing on clusters' major hyperprototypes that capture the shared behavioural modes underlying weekday consumption.

\subsection*{Summer Partition}
\input{SUMMER_Hyperprototypes}

All consumers in the AG Solar Home Electricity dataset possess some form of generation capacity, and variation in the extent of generation emerged as a key factor distinguishing clusters and patterns shared within clusters. Taken together, consumers in Clusters 2, 3, and 4 showed relatively low levels of generation compared to their consumption, while those in Clusters 1 and 5 often exhibited generation comparable to or exceeding their demand.

Consumption patterns in cluster 1, the largest consumer group, were strongly differentiated by relative levels of generation. On approximately one third of days, households exhibited profiles dominated by significant generation, with only mild morning (06:00) and evening (20:00) peaks. Nonetheless, nearly all consumers displayed more substantial evening consumption around 18:00–19:00 on 60\% of days with relatively moderate or mild levels of generation.

Cluster 2 was characterised by dominant post-midnight consumption peaks on about 75\% of days, with minor morning (07:00) and evening (19:00–20:00) peaks. The outlying consumption prototypes also showed a tendency for peaks post midnight, with sporadic peak consumption elsewhere in the day.

Cluster 3 was characterised by minimal generation compared to consumption on 40\% of days, and moderate generation on 35\% of days. In either case, consumption peaked late in the evening between 20:00-21:00. For a small subset of consumers, consumption peaked later again on 10\% of days, just prior to midnight.

Cluster 4 was the smallest group and showed highly distinctive behaviour. For 90\% of days, consumption was concentrated at or just after midnight. This could indicate occupants with late-night routines such as shift workers, or could indicate the presence of appliances such as electric hot water systems, pool pumps. Minor peaks around 07:00 and 19:00 were also consistently observed for these consumers, as evidenced by the green mean hyperprototype.

Cluster 5's green hyperprototype, accounting for all consumers on 45\% of days, had a very similar shape to cluster 1's green hyperprototype, with an evening consumption peak around 19:00 and relatively moderate generation. However on another 35\% of days, a less prominent and more sustained peak in consumption could be observed later in the evening, around 20:00. Additionally, around 45\% of consumers used minimal electricity throughout the day after an early morning peak on 15\% of days. 

\input{WINTER_Hyperprototypes}

\subsection*{Winter Partition}

The consumer partition was more balanced for the winter period compared to summer, suggesting that consumers demonstrate greater uniformity through warmer seasons. This supports our earlier findings in relation to \Cref{Fig:EquivalenceBetweenRLSandGPF} which showed that more diverse consumption patterns tend to be seen during cooler seasons. 

Similar to its summer counterpart, Cluster 1's hyperprototypes were strongly differentiated by relative levels of generation. All three major hyperprototypes (accounting for about 85\% of days) from cluster 1 exhibit moderate morning consumption peaks around 07:00 with varying degrees of evening consumption typically centred between 17:30–18:30. Generation was significant compared to consumption on just over half of the days from this period (pink and orange hyperprototypes). 

Cluster 2 consumers, by contrast, exhibited limited generation relative to consumption on around two thirds of days, and significant consumption peaks throughout the evening between 17:00–21:00. Around 20\% of days displayed higher generation for around 85\% of consumers, coinciding with a slight shift in evening usage later and morning usage earlier — suggesting a degree of responsiveness to weather signals.

Consumption peaked in the morning around 07:00-08:00 for the consumers in cluster 3 on 35\% of days, and in the evening around 18:00-19:00 on another 40\% of days. For around 50\% of consumers in this cluster, a ``standby'' profile --- with significant generation and almost negligible
consumption --- was exhibited on 15\% of days. 

Cluster 4 resembled the summer counterpart, but with increased membership. Just over half of days were dominated by midnight peaks, with a tendency for sightly pre-midnight peaks for around 80\% of consumers on another one third of days. Secondary morning (05:00–07:00) and early evening (17:00–18:00) peaks were also present, though less pronounced.

Cluster 5, the smallest winter group, was similar to the summer Cluster 2 but reduced in size. For 90\% of days, households showed strong peaks in the early morning, supplemented by minor morning (06:00-07:00) and evening (17:00-19:00) peaks.

\subsection*{Implications}
The coverage plots in \Cref{Fig:Application-SUMMER,Fig:Application-WINTER} reveal that, within clusters, some consumers are not represented as well as the majority by the main hyperprototypes. For example, in winter cluster 5 one consumer had about 40\% of their days not captured by the dominant hyperprototype, whereas most others were covered to a much greater extent. Similar patterns can be observed across the other summer and winter clusters. In practice, program designers may wish to acknowledge this variability and prioritise targeting consumers whose behaviours are most closely aligned with the dominant hyperprototypes, or with their cluster’s RRLS more broadly. A natural way to assess this is by computing WSMD distances between each consumer’s RLS and their cluster’s RRLS, weighting hyperprototypes by prevalence. Such an approach identifies consumers who are most representative of the cluster’s shared behavioural modes, and hence most likely to respond in a predictable way to interventions. Alternatively, filtering could emphasise high-consumption households to maximise aggregate impact, while excluding those below a minimum consumption threshold reduces the risk of limited returns.

Meanwhile, many clusters did display clear and recurrent consumption patterns that cover the majority of consumers, and these provide a strong foundation for the design and targeting of DSM and DR strategies. Beyond these initial phases, CROCS also provides a practical means to evaluate program outcomes. Rather than re-clustering the entire dataset, utilities could revisit the consumers within a given cluster after some period of time, recomputing their stage one behaviours and pairwise similarities to derive updated hyperprototypes. Shifts in the prevalence or shape of these hyperprototypes would then indicate whether an intervention, such as a time-of-use tariff, appears to have had the intended effect. In this way, CROCS supports not only the design but also the ongoing assessment and refinement of demand-side strategies.

%% file: SUMMER_Hyperprototypes.tex
\begin{figure}[!t]

    \centering
    \begin{subfigure}[b]{\textwidth}
        \centering
        \includegraphics[width=\textwidth]{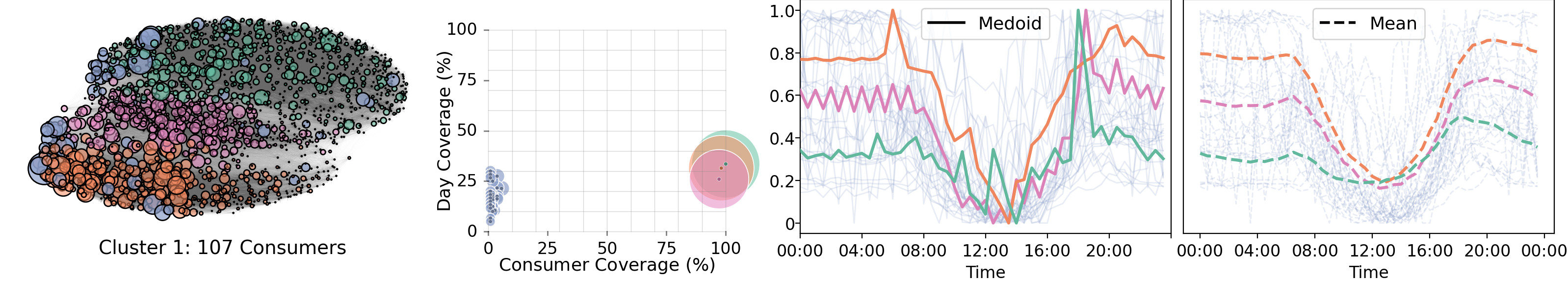}
    \end{subfigure}

    \begin{subfigure}[b]{\textwidth}
        \centering
        \includegraphics[width=\textwidth]{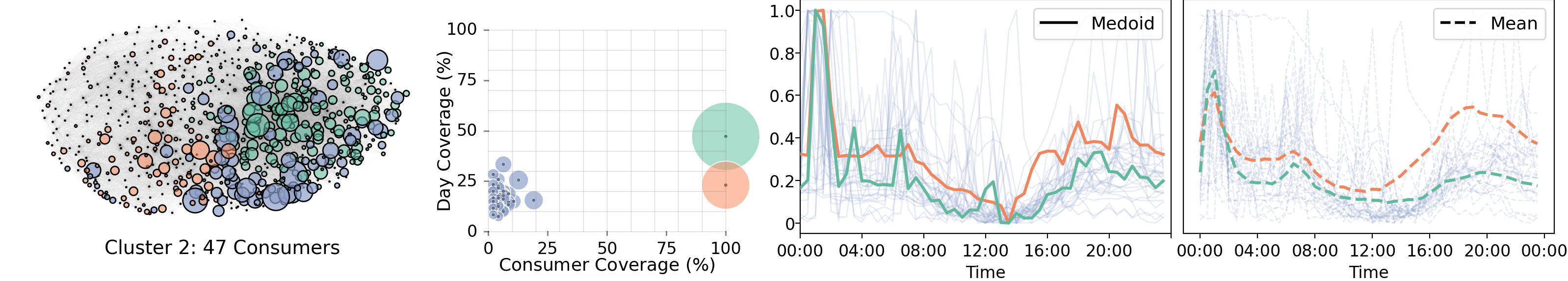}
    \end{subfigure}

    \begin{subfigure}[b]{\textwidth}
        \centering
        \includegraphics[width=\textwidth]{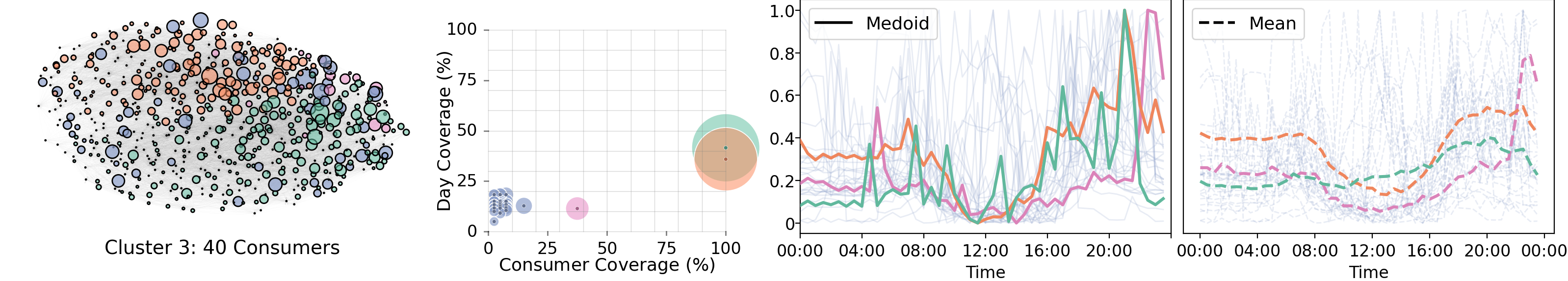}
    \end{subfigure}

    \begin{subfigure}[b]{\textwidth}
        \centering
        \includegraphics[width=\textwidth]{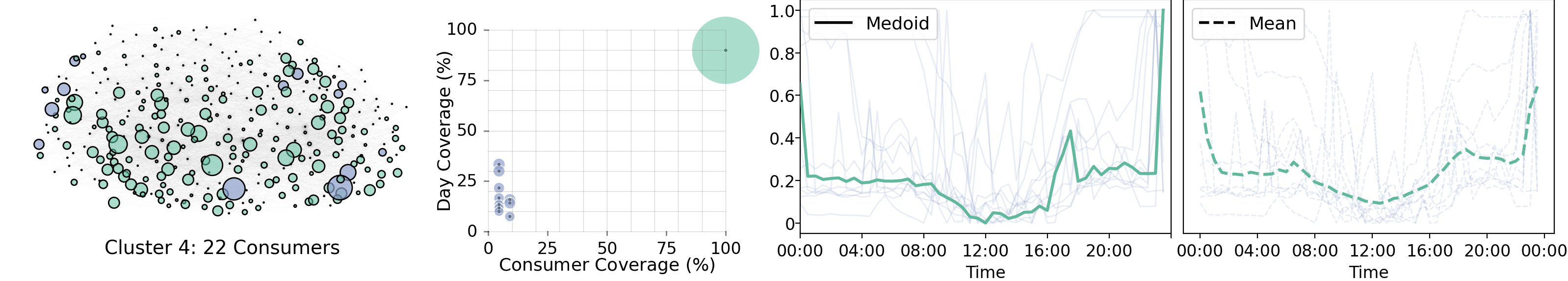}
    \end{subfigure}

    \begin{subfigure}[b]{\textwidth}
        \centering
        \includegraphics[width=\textwidth]{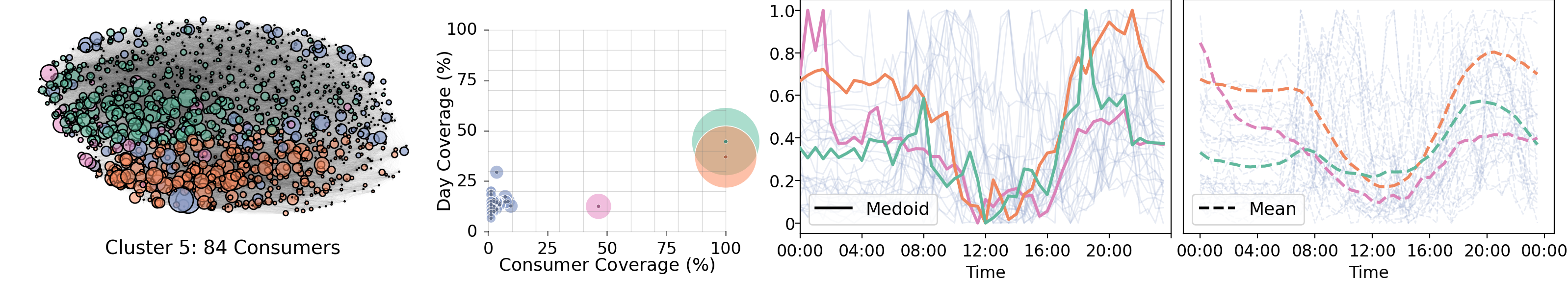}
    \end{subfigure}

\caption{Five consumer clusters for a \textbf{summer} period from the Ausgrid dataset, showing the communities discovered in the prototype graphs on the left, coverage statistics plots, and hyperprototypes for each major community discovered (medoids and means) on the right. Workdays from 1st December 2012 to 28th February 2013 inclusive (60 days) were clustered using CROCS with min-max normalised DLPs, DTW-2, $k=15$, medoid prototypes, and KMd for both stages. Major hyperprototypes representing common patterns shared by a significant proportion of consumers are highlighted with brightly coloured lines, while minor hyperprototypes capturing outlying consumption behaviours are shown in navy. Coverage statistics indicate the percentage of consumers ($x$-axis) and days ($y$-axis) represented by each hyperprototype, with circle sizes proportional to consumer coverage}
\label{Fig:Application-SUMMER}

\end{figure}

%% file: WINTER_Hyperprototypes.tex
\begin{figure}[!t]

    \centering
    \begin{subfigure}[b]{\textwidth}
        \centering
        \includegraphics[width=\textwidth]{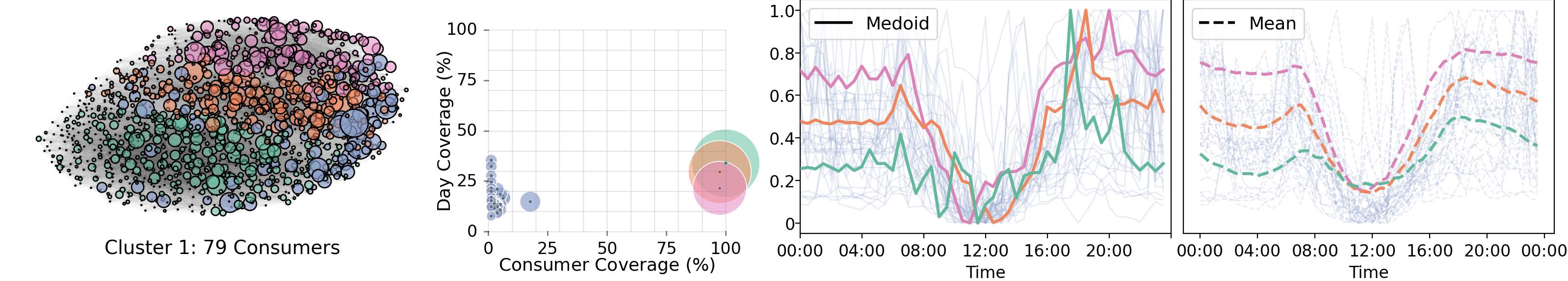}
    \end{subfigure}

    \begin{subfigure}[b]{\textwidth}
        \centering
        \includegraphics[width=\textwidth]{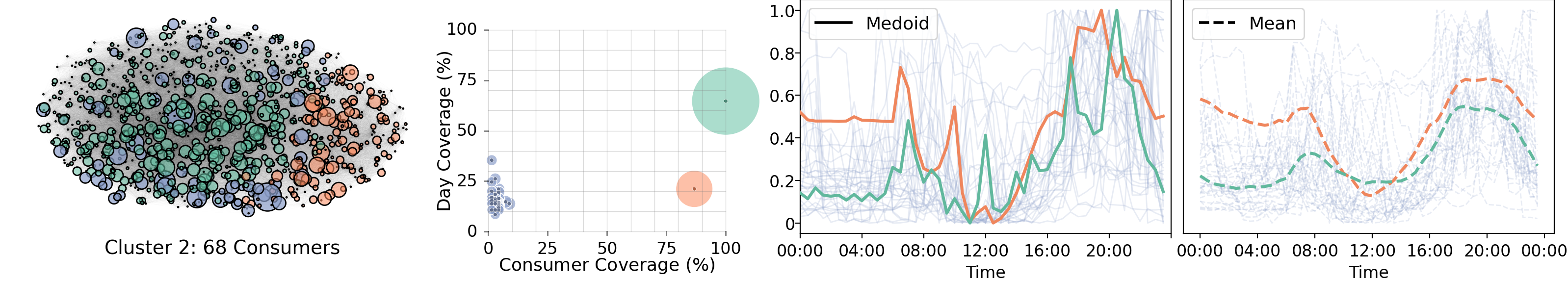}
    \end{subfigure}

    \begin{subfigure}[b]{\textwidth}
        \centering
        \includegraphics[width=\textwidth]{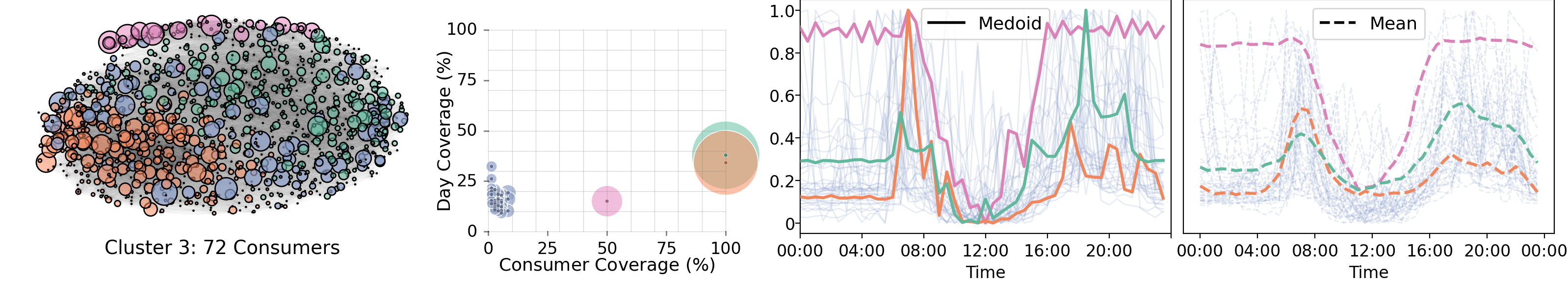}
    \end{subfigure}

    \begin{subfigure}[b]{\textwidth}
        \centering
        \includegraphics[width=\textwidth]{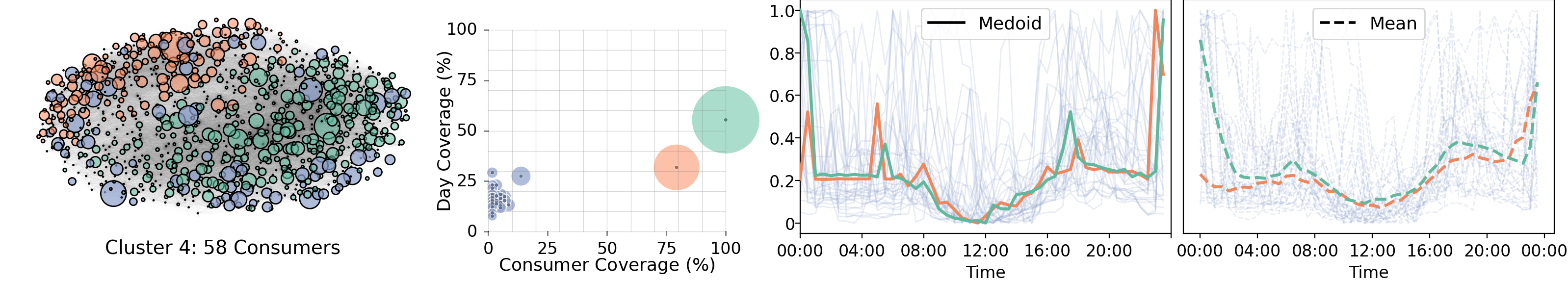}
    \end{subfigure}

    \begin{subfigure}[b]{\textwidth}
        \centering
        \includegraphics[width=\textwidth]{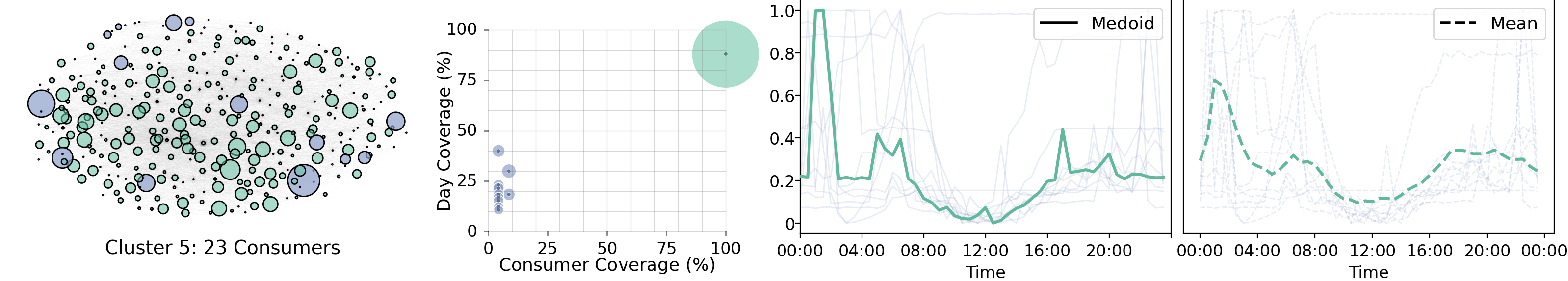}
    \end{subfigure}

\caption{Five consumer clusters for a \textbf{winter} period from the Ausgrid dataset, showing the communities discovered in the prototype graphs on the left, coverage statistics plots, and hyperprototypes for each major community discovered (medoids and means) on the right. Workdays from 1st June to 31st August 2012 inclusive (65 days) were clustered using CROCS with min-max normalised DLPs, DTW-2, $k=15$, medoid prototypes, and KMd for both stages. Major hyperprototypes representing common patterns shared by a significant proportion of consumers are highlighted with brightly coloured lines, while minor hyperprototypes capturing outlying consumption behaviours are shown in navy. Coverage statistics indicate the percentage of consumers ($x$-axis) and days ($y$-axis) represented by each hyperprototype, with circle sizes proportional to consumer coverage}
\label{Fig:Application-WINTER}
\end{figure}

%% file: Discussion.tex
\section{Discussion}
\label{Sec:Discussion}

In \Cref{Sec:LiteratureReview-Limitations} we identified a series of eight methodological challenges that constrain the effectiveness of existing consumer clustering methodologies. \Cref{Sec:Discussion-Limitations} examines how CROCS addresses each of these in turn, drawing on both the framework's design and experimental evidence. \Cref{Sec:Discussion-Guidance} then consolidates implementation considerations, before \Cref{Sec:Discussion-FutureDirections} considers future research directions.

\subsection{Addressing Limitations of Existing Methods}
\label{Sec:Discussion-Limitations}

\subsubsection*{(i) Capturing Intra-Consumer Variation} \label{Sec:Discussion-Limitations(i)}
Central to the CROCS framework, the RLS preserves behavioural diversity by providing a flexible, data-driven summary of a consumer's typical diurnal consumption behaviours, capturing both their shape and prevalence. As demonstrated in \Cref{Sec:Results_Comparing_Consumer_Representations}, this locally fitted consumer representation efficiently achieves lower reconstruction error than other load-profile based representations that either use single profiles (e.g., RLP, DCP), or multiple profiles discovered globally (e.g., GPF). Since intra-consumer variation has been recognised as an indicator of suitability for DSM and DR programs, a representation that faithfully captures this diversity is particularly valuable for consumer segmentation.

\subsubsection*{(ii) Preserving Common Intra-Consumer Variation in Clusters}
Given the literature emphasis on intra-consumer variation, it would be contradictory to then obscure shared variation when it defines consumer clusters --- as is done when clusters are summarised by single prototypes. As demonstrated in \Cref{Sec:Application}, many real-world consumer clusters were characterised by a small number of distinct behavioural modes that were shared across the majority of members. Representing each of these modes with their own profile in the CROCS RRLS makes explicit the concrete behaviours that connect consumers within a group, enhancing interpretability and providing a firm basis for downstream applications.

\subsubsection*{(iii) Recognition of Asynchronous Consumer Similarity}
Because the RLS representation is not temporally anchored, CROCS can cluster together consumers that exhibit similar diurnal patterns whether they occur on the same, or on different days. As illustrated between \Cref{Fig:PairsExampleSchedules,Fig:JSD_vs_CK}, such asynchronous pairs are likely far more numerous than their synchronised counterparts. While some other methods are in principle capable of recognising similar yet asynchronous pairs of consumers, it was demonstrated in \Cref{Sec:Results_Asynchronous} that CROCS consistently does so more effectively across a range of dataset conditions. 

It bears emphasising that CROCS extends recognition to asynchronous similarity while fully retaining its capacity to detect synchronous similarity. This capability is significant for downstream applications where concurrency isn't strictly essential --- for instance, in selecting households for distributed energy resource adoption or curtailable service incentives, or designing and deploying TOU tariffs. It also supports equity in program design, by ensuring that consumers with similar behavioural potential are not excluded from DSM or DR opportunities simply because their schedules are atypical.

\subsubsection*{(iv) Robustness to Behavioural Anomalies} \label{Sec:Discussion-Limitations(iv)}
The WSMD distance measure used in the CROCS framework weights consumption patterns according to their frequency when assessing consumer similarity. In this way, unusual behaviours are acknowledged within consumer RLSs but prevented from distorting the clustering outcome. Experiments in \Cref{Sec:Results_CROCS_SetDistanceComparison} demonstrate that WSMD consistently outperforms alternative set-to-set distances --- even in the presence of significant numbers of outlying diurnal patterns. Unlike other distances such as SMD or MHD, it also required the smallest overestimation of $k$ to achieve reliable results, making it well suited to real-world data where the prevalence of outliers is unknown. Furthermore, in the synthetic data experiments of \Cref{Sec:Results_Asynchronous}, CROCS demonstrated the greatest robustness to increasing numbers of outlying DLPs across all tested methods, with the closest competitor requiring substantially larger values of $k$ to compensate for their presence.

\subsubsection*{(v)-(vii) Flexible Handling of Non-Synchronised Data, Regular Discontinuities, and Missing Data}
CROCS naturally handles irregularities and discontinuities in input data that commonly hinder other consumer clustering approaches by treating each consumer's available DLPs as an unordered set when deriving their RLS.

Consequently, the framework does not require time-synchronised or equal-length series. Consumers with staggered installation dates or different quantities of relevant historical data can therefore be included without truncating longer records or discarding more recent installations. Regular discontinuities, such as those created when analysing workdays separately from weekends (as in \Cref{Sec:Application}), are likewise accommodated. Unlike approaches that require concatenating subsequences of long time series, gaps between selected subsets of days do not introduce spurious temporal dependencies into the RLS representation. Finally, missing data can be handled simply by omitting significantly affected days from the RLS computation, avoiding the need for large-scale imputation or consumer exclusion. This ensures that segmentation makes use of the maximum amount of reliable information while maintaining representativeness across the population. 

By addressing these limitations, CROCS minimises the amount of reliable data discarded, and provides more equitable foundations for DSM and DR program design.

\subsubsection*{(viii) Scalability and Practical Deployment}

While scalability is essential for practical deployment, computational efficiency should not come at the cost of segmentation quality. CROCS maintains this balance by decomposing the problem into two stages: the first stage of clustering for local consumer representations is fully independent and parallelisable, while computation of WSMD distances between compact consumer RLSs can also be parallelised in the second stage. This contrasts with effective global approaches such as GPF, which scale poorly with the number of consumers and whose representations must be entirely recomputed whenever new consumers are added. As shown in \Cref{Sec:Results_Scalability}, CROCS scales quadratically with the number of consumers while still achieving runtimes that render large-scale consumer segmentation practically feasible, in contrast to the much smaller populations typically addressed in prior studies. In addition, robustness to overestimation of the stage one parameter $k$ reduces the need for costly parameter tuning, providing a practical configuration strategy at scale that further supports deployment across large consumer populations.

\subsection{Configuration and Deployment Guidance}
\label{Sec:Discussion-Guidance}

This subsection consolidates practical guidance on the key design decisions involved in configuring and applying CROCS, bringing together findings from within and beyond this study.

\subsubsection*{Stage One}
CROCS is an inherently modular framework that does not prescribe any particular clustering approach or components for constructing consumer RLSs in stage one, allowing it to be tailored to the problem at hand or to accommodate methodological advances as they emerge. A comprehensive comparison of clustering components in the context of half-hourly sampled DLP clustering --- evaluating 31 distance measures, 8 representations, and 11 clustering algorithms from the smart meter time series and broader clustering literature --- is provided in \cite{Yerbury2024b}, whose findings inform the stage one configuration adopted throughout this paper. 

In particular, that study found that distance measures (applied to normalised DLPs) accommodating local temporal shifts while maintaining sensitivity to amplitude differences --- particularly DTW and $k$-Sliding Distance (KSD) --- provided the most consistent strong performance. Precise parameter tuning of their single intuitive warping window parameter was not found to be critical, with values between $0.5$ and $3$ hours for DTW consistently outperforming both the unconstrained default and alternative parameterless distances. Given the greater computational cost of KSD, this makes DTW a markedly practical choice, especially compared to other more complex or highly parametrised methods. Moreover, when combined with either KMd or HAC-Wa, DTW demonstrated remarkable robustness across a range of varied dataset characteristics, with KMd proving particularly effective in the presence of outlier DLPs. 

Regarding the remaining stage one components, some form of normalisation applied independently to each DLP is recommended as this directs the algorithm toward shape rather than amplitude differences. Min-max normalisation was adopted throughout this study as a simple and widely used choice, though alternatives may prove preferable depending on the application --- a question discussed further in \Cref{Sec:Discussion-FutureDirections}. Medoid prototypes are recommended over pointwise mean profiles, which can lie in low-density regions of the data subspace, produce excessively smoothed profiles, or become ill-defined under elastic distance measures. Moreover, medoid prototypes offer greater robustness to outlying DLPs in real-world data where anomalous days are common.

This leaves $k_i$, which is the number of clusters used to construct the $i\textsuperscript{th}$ consumer's RLS, as the only remaining parameter to configure for stage one. We recommend adopting the common value of $k$ across all consumers, rather than optimising $k_i$ for each consumer. In principle, $k$ should be at least $\max_ik_i^*$, where $k_i^*$ is the true number of distinct recurring behavioural modes for the $i\textsuperscript{th}$ consumer. However, since $\max_ik_i^*$ is unknown in practice, and outlier DLPs (which are common in real data \cite{Yerbury2024b}) require additional clusters to be adequately represented, overestimation is both necessary and desirable. As demonstrated in \Cref{Sec:Results_CROCS_StageOne_Picking_k}, performance plateaus once $k$ is sufficient to capture each consumer's behavioural diversity, with further increases incurring only modest additional computational cost without degrading segmentation quality. This robustness to overestimation is underpinned by frequency-weighting within WSMD, which ensures that behavioural modes split across multiple clusters due to overestimation do not disproportionately influence pairwise consumer dissimilarities. 

In practice, determining an appropriate value for $k$ need not be exact. The appropriate value will be influenced by basic dataset characteristics, such as the number of seasons spanned or whether weekends are included, and a modest overestimate relative to the expected behavioural complexity of the consumer population is advisable. As a rough reference, $k=15$ was found appropriate for the two datasets analysed in \Cref{Sec:Application}, where each covered a three-month period with weekends excluded. For practitioners seeking a more dataset-specific reference, careful clustering of the DLPs of a few randomly sampled consumers --- using relevant RVIs and visual inspection --- can provide an improved estimate of the typical behavioural complexity and outlier prevalence in their dataset, from which an appropriate overestimate of $k$ could then be derived. 

However, as stage two runtime scales quadratically with $k$, unnecessarily large values should still be avoided in large-scale deployments. Moreover, larger values of $k$ would effectively make WSMD computations higher dimensional, reintroducing the curse of dimensionality and thereby diluting the meaningfulness of consumer comparisons. The computational cost of such modest overestimation thus remains practical, particularly when weighed against the alternative of per-consumer RVI optimisation, which would itself require generating and evaluating many additional clusterings at considerable expense and with potentially less reliable outcomes.

\subsubsection*{Stage Two}
For the second stage of CROCS, we recommend implementing WSMD with the same distance measure used in stage one, enforcing conceptual consistency when computing distances between RLS prototypes. Meanwhile, the choice of clustering algorithm is not prescribed. The same algorithm applied in stage one was used for stage two throughout this study for convenience, though alternatives could be compared using RVIs or by inspecting the resulting RRLSs. The number of consumer clusters $K$ is not a novel consideration, and should be guided by application requirements, domain knowledge, and standard clustering practice.

\subsubsection*{RRLS Computation}
The RRLS can be computed using any community detection algorithm compatible with weighted directed graphs. In \Cref{Sec:Application}, the Leiden algorithm was employed with the RBER quality function, which accommodates vertex and edge weights and includes a resolution parameter $\gamma$. The number of hyperprototypes, or equivalently, the number of communities, is a user-controlled parameter governing the level of detail in the resulting cluster summaries, and for the Leiden algorithm can be varied by adjusting $\gamma$. Visualisation alongside the associated coverage statistics provides a natural basis for identifying the appropriate number of hyperprototypes to retain. Medoid hyperprototypes are likewise recommended to reduce sensitivity to this choice by remaining robust to more anomalous prototypes within each community.

With this configuration, determining an appropriate number of hyperprototypes requires some iteration, guided by the desired balance between conciseness and behavioural detail. Outlying patterns tend to fracture out as the number of communities increases, with more dominant communities emerging more gradually. This means more hyperprototypes may be needed than one might initially expect --- for instance, in \Cref{Sec:Application}, 10-40 hyperprototypes were selected per consumer cluster, many representing small communities of outlying patterns, alongside a few with significant coverage. If instead a quick and approximate summary is needed, rough values suffice, or as discussed in \Cref{Sec:Discussion-FutureDirections}, lighter-weight alternatives to the full graph construction represent a promising trade-off for improving the scalability of this optional step.

\subsubsection*{Scalability Considerations}
Finally, the overall computational complexity of the framework largely depends on the chosen clustering components, but the configuration adopted in this study is $\bigO{mp^2 + k^2m^2}$ (see \Cref{Sec:Results_Comparing_Consumer_Representations}). Both stages of CROCS offer opportunities for parallelisation that can significantly reduce runtime in practice. Stage one is fully parallelisable across consumers, while pairwise WSMD distance computations for clustering in stage two can likewise be distributed. Beyond parallelisation, faster variants of the recommended components, such as accelerated or approximate DTW \cite{10.1145/2783258.2783286} and parallel or approximate hierarchical clustering \cite{10.1007/978-3-540-27866-5_47,Kull2008}, represent further avenues for efficiency gains not exploited in this study. For deployments involving very large consumer populations, a practical strategy would be to apply CROCS to a representative sample rather than the full dataset.

\subsection{Future Directions}
\label{Sec:Discussion-FutureDirections}
CROCS addresses key shortcomings of existing methods, yet there are opportunities for further refinement and extension to maximise its utility in practical contexts.

Firstly, the community detection algorithm used to determine the RRLS summaries for consumer clusters, while effective for identifying shared behaviours, currently scales less efficiently than the two primary stages of CROCS. Future work could assess the trade-off between representation quality and scalability in more lightweight alternative algorithms that utilise simpler graph constructions obtained by omitting edge directions, node weights, or edge weights (or combinations thereof). Moreover, it could also be productive to consider alternative transformations of the WSMD dissimilarities into connection strengths (e.g. by subtracting dissimilarities from an appropriate maximum) or the use of community detection algorithms that operate directly on dissimilarities as edge weights instead. A more radical option would be to replace the graph conception entirely with a direct time series clustering of the prototypical DLPs.

Secondly, while the popular min–max normalisation was adopted in this study to ensure that clustering emphasised shape rather than magnitude, a range of alternatives have been used throughout the literature --- yet there is little systematic understanding of when and why one should be preferred. Future research could therefore evaluate the suitability of different procedures within the CROCS framework, and develop recommendations for which procedures are most appropriate for particular applications. For instance, alternative schemes such as $z$-normalisation or unit-norm scaling \cite{Yerbury2024} may provide distinct advantages. Some form of selective normalisation could also prove beneficial --- for instance, clustering unnormalised DLPs (or DLPs extracted from normalised long time series) in the first stage, which could differentiate shapes associated with higher and lower consumption levels for individual consumers --- with normalisation applied only to prototypes in the second stage to enable comparisons between consumers. Relatedly, pre-binning consumers by overall consumption level and then performing clustering separately on consumers within each bin, as suggested by \cite{Toussaint2019,Toussaint2020}, could provide an effective pathway for scaling segmentation to many hundreds of thousands or even millions of consumers.

Thirdly, the computational effort invested in stage one of CROCS yields consumer-level outputs that could support a range of analyses complementary to the system-level segmentation. The individualised RLS summaries provide a basis for targeted consumer-level analysis, such as detecting anomalous behavioural shifts arising from occupancy changes or the uptake of electric vehicles, PV or household batteries; identifying rebound effects in response to TOU tariff structures \cite{Yan2018}; or monitoring the success of demand-side interventions over time. Realising this potential represents a promising direction for future CROCS research. 

In a similar vein, CROCS could be extended beyond segmentation by hybridising it with forecasting tasks. While stage two clusters could be utilised for aggregate-level forecasting \cite{Kim2023Time-seriesData,Auder2018,Li2016Short-TermBehavior,Quilumba2015b}, stage one prototypes also provide natural primitives for forecasting, with sequence-based approaches such as Pattern Sequence Forecasting \cite{Martinez-Alvarez2011a,Martinez-Alvarez2008} or Markov-based forecasting \cite{Teeraratkul2018} applicable to consumers' stage one cluster label sequences. Future research could compare the
accuracy of such CROCS-informed forecasting with existing methods. In some applications, segmentation and forecasting may even serve as complementary goals, with this line of inquiry offering a pathway to integrate both within a single framework for operational use.

Finally, while CROCS has been developed and explored in this paper in the context of smart-meter data, the framework is certainly applicable to a broader class of problems where synchronised \textit{and} asynchronous similarities across sub-periods are of interest. For instance, household water consumption is highly analogous \cite{Steffelbauer2021,Malinowski2022}, where daily profiles of water use could be clustered in stage one, with stage two then grouping households by the distribution of these patterns. Electric vehicle charging time series provide another example, where individual charging sessions provide an analogue of the stage one ``sub-periods'' to be characterised with typical charging curves, while stage two would group drivers or charging sites according to the mix of sessions they exhibit, likely supporting infrastructure planning and grid integration. More generally, CROCS could be applied wherever long time series can be segmented into recurring periods --- for example retail demand cycles, machine operation cycles, or climate patterns. Beyond time series, it could also be extended to domains with repeated feature-described events across multiple entities, such as sports analytics or retail transaction data. In these cases, stage one would cluster event-level feature vectors and stage two would group entities (players or teams, customers or users) according to their prototypical features discovered in stage one. These examples are far from exhaustive, but they illustrate the broader potential of CROCS to uncover interpretable behavioural structures across a wide variety of dataset formats and domains.

%% file: Conclusion.tex
\section{Conclusion}
\label{Sec:Conclusion}

This study has introduced CROCS --- Clustered Representations Optimising Consumer Segmentation --- a novel two-stage clustering framework designed to address the eight methodological challenges identified in our review of existing approaches. In the first stage, each consumer’s daily load profiles are clustered independently to produce a high-quality, local consumer representation via cluster prototypes. The resulting Representative Load Sets (RLS) provide rich yet compact summaries of each consumer's typical consumption patterns, preserving intra-consumer variation in an interpretable form, while naturally accommodating non-synchronised data, regular discontinuities, and missing data. In the second stage, consumers are clustered by comparing their RLSs using the Weighted Sum of Minimum Distances (WSMD). This set-to-set distance incorporates all prototypical behaviours while weighting them by prevalence, enabling the discovery of both synchronous and asynchronous similarity between consumers and reducing the influence of anomalous patterns. Finally, we developed an optional cluster-level representation, the Refined RLS (RRLS). Constructed directly from the similarity structure revealed by CROCS, RRLSs summarise consumer clusters with multiple hyperprototypes that represent the distinct behavioural modes shared across members. This avoids the loss of nuance that occurs when clusters are compressed into a single profile, thereby providing a more interpretable and faithful characterisation of consumer groups discovered by CROCS.

Our experimental analyses revealed an important and under-explored property of electricity consumption data: consumers often express similar diurnal behaviours asynchronously --- exhibiting highly similar patterns but on different calendar days. We showed that pairs of such consumers appear more common than those where similar behaviours are temporally aligned, highlighting the limitations of clustering methods that rely on calendar synchronisation. CROCS also proved the most effective at detecting such asynchronous similarity, outperforming a broad range of existing methodologies --- including those that in principle should also recognise asynchronous similarity. This finding has important implications for DSM and DR program design, as it demonstrates that many consumers with similar potential or suitability may be misclassified when relying on temporally anchored approaches.

Beyond this key result, our experiments validated the framework’s design choices while establishing clear guidance for its application. Most notably, simple overestimation of the stage one parameter $k$ --- the number of prototypes in each consumer’s RLS --- proved to be a robust and practical configuration strategy for the framework’s sole parameter. The RLS also consistently outperformed alternative single- and multi-\textit{profile} representations, achieving lower reconstruction error across a wide range of practical $k$. WSMD emerged as the most effective set-to-set distance, demonstrating robustness to outliers and working particularly well in combination with the overestimation strategy. Furthermore, CROCS demonstrated practical feasibility through a structure that is readily parallelisable and open to further optimisation. In our application to Australian smart meter data, hyperprototypes and their coverage statistics yielded interpretable seasonal summaries, clarifying the concrete daily consumption patterns that brought consumers together within the same cluster.

Beyond consumer segmentation, CROCS can also provide insights at the level of individual consumers, with stage one clustering supporting tasks such as anomaly detection, monitoring for behavioural change, and identifying the uptake of new technologies. The framework is also inherently modular, so while it has been implemented here using min-max normalisation with constrained Dynamic Time Warping and either $k$-medoids or Ward's hierarchical clustering for consistency with comparative studies \cite{Yerbury2024b}, CROCS can readily adopt alternative methodologies as required. This modularity, combined with its domain-agnostic foundations, makes the framework readily extensible beyond smart meter data to any setting where long time series can be meaningfully decomposed into recurring periods.

By unifying methodological rigour with practical scalability in the energy domain, CROCS marks a step forward in behaviour-centric consumer segmentation --- offering a robust and extensible framework that enhances the analytical value of smart meter data while strengthening the operational foundations of DSM and DR strategies in increasingly complex electricity systems.